\documentclass[final,1p,times,authoryear]{elsarticle}

\usepackage{booktabs, tabularx}
\usepackage{multirow}
\usepackage{threeparttable}
\usepackage{longtable}
\usepackage{threeparttablex}
\usepackage{caption}

\usepackage{enumitem}
\setlist[itemize]{noitemsep}

\usepackage{amsmath,amsfonts,amssymb,accents}

\usepackage{algorithmic}
\usepackage{algorithm}
\usepackage{array}
\usepackage{graphicx}
\graphicspath{{img/}}
\usepackage{natbib}

\usepackage{rotating}
\usepackage[table]{xcolor}
\usepackage{tablefootnote}
\usepackage{hyperref}
\usepackage{adjustbox}
\usepackage{multicol}
\usepackage{color,soul}
\usepackage{booktabs}
\usepackage{multirow}
\usepackage{caption}
\usepackage{siunitx}
\usepackage{subcaption}
\usepackage{geometry}
\usepackage{pdflscape}
\makeatletter
\def\widebreve{\mathpalette\wide@breve}
\def\wide@breve#1#2{\sbox\z@{$#1#2$}%
     \mathop{\vbox{\m@th\ialign{##\crcr
\kern0.08em\brevefill#1{0.8\wd\z@}\crcr\noalign{\nointerlineskip}%
                    $\hss#1#2\hss$\crcr}}}\limits}
\def\brevefill#1#2{$\m@th\sbox\tw@{$#1($}%
  \hss\resizebox{#2}{\wd\tw@}{\rotatebox[origin=c]{90}{\upshape(}}\hss$}
\makeatletter

    \newcommand{\midsepremove}{\aboverulesep = 0mm \belowrulesep = 0mm}
    \midsepremove
    \newcommand{\midsepdefault}{\aboverulesep = 0.1mm \belowrulesep = 0.1mm}
    \midsepdefault
 \newcommand{\notesize}{\fontsize{8.5}{10.5}\selectfont}
 
\newtheorem{definition}{Definition}

\begin{document}
\begin{frontmatter}



\title{Time-Aware and Transition-Semantic Graph Neural Networks for Interpretable Predictive Business Process Monitoring \tnoteref{code} \tnoteref{funding} } 
\tnotetext[code]{The full codebase is available at \url{https://github.com/skyocean/TemporalAwareGNNs-NextEvent}}
\tnotetext[funding]{This research has been supported by the project MUSA - Multilayered Urban Sustainability Action, funded by the European Union - NextGenerationEU, under the National Recovery and Resilience Plan (NRRP) M4C2 I1.5: Strengthening of research structures and creation of R\&D innovation ecosystems, set up of territorial leaders in R\&D (CUP G43C22001370007, Code ECS00000037)}


\author[label1]{Fang Wang\corref{cor1}}
\ead{florence.wong@ku.ac.ae}
\cortext[cor1]{Corresponding Author} 
\affiliation[label1]{organization={College of Computing and Mathematical Sciences, Khalifa University},
           city={Abu Dhabi},
           country={UAE}}
\author[label2]{Ernesto Damiani}
\ead{ernesto.damiani@unimi.it}
\affiliation[label2]{organization={Department of Computer Science, Università degli Studi di Milano}, city={Milan},country={Italy}}
\begin{abstract}
Predictive Business Process Monitoring (PBPM) aims to forecast future events in ongoing cases based on historical event logs. While Graph Neural Networks (GNNs) are well suited to capture structural dependencies in process data, existing GNN-based PBPM models remain underdeveloped. Most rely either on short prefix subgraphs or global architectures that overlook temporal relevance and transition semantics. We propose a unified, interpretable GNN framework that advances the state of the art along three key axes. First, we compare prefix-based Graph Convolutional Networks(GCNs) and full trace Graph Attention Networks(GATs) to quantify the performance gap between localized and global modeling. Second, we introduce a novel time decay attention mechanism that constructs dynamic, prediction-centered windows, emphasizing temporally relevant history and suppressing noise. Third, we embed transition type semantics into edge features to enable fine grained reasoning over structurally ambiguous traces. Our architecture includes multilevel interpretability modules, offering diverse visualizations of attention behavior. Evaluated on six benchmarks with different decay factors, the proposed models achieve competitive Top-$k$ accuracy and Damerau–Levenshtein scores under a consistent global setup. A separate analysis of the decay coefficient $\lambda$ confirms its role in adapting attention, with effective values calibrated by temporal structure and sequence depth. By addressing architectural, temporal, and semantic gaps, this work presents a robust, generalizable, and explainable solution for next event prediction in PBPM.
\end{abstract}


\begin{keyword}
PBPM, GNN, GATconv, GCNconv, Time-Decay Attention, Semantic Edge Embeddings

\end{keyword}

\end{frontmatter}



\section{Introduction}
\label{sec:Intro}
Predicting the next event in an ongoing business process is a cornerstone of operational intelligence across domains such as customer support, healthcare, finance, and manufacturing. Predictive Business Process Monitoring (PBPM) leverages historical event logs to anticipate future steps, enabling proactive decision making and real time optimization \citep{pasquadibisceglie2021multi, zou2024prediction}. With the advancement of deep learning, graph neural networks have emerged as powerful tools for modeling complex event sequences \citep{harl2020explainable}. Despite increasing interest, existing graph neural network approaches for PBPM exhibit critical limitations in both their architectural design and their treatment of event and transition dynamics. 

First, most existing models rely either on short prefix based subgraphs \citep{dissegna2025graph} or on global attention applied across entire traces. Prefix based approaches are computationally efficient but narrow in scope, often failing to capture long range dependencies. In contrast, global models provide richer context but tend to treat all events uniformly, without discounting distant and low signal history. Few studies compare these two paradigms or offer practical guidance on when one outperforms the other, leaving a methodological gap. Second, the integration of temporal information remains underexplored, particularly in graph neural network based models, which are well suited to encode both node level and edge level temporal signals. Temporal interactions can occur both locally—between consecutive events—and globally—between individual events and the overall sequence timeline. While local time differences have occasionally been used as static edge features in global models, their role is often limited and inconsistently applied. In contrast, global time alignment mechanisms, such as dynamic attention windows centered on the prediction point, are conceptually well motivated but remain almost entirely unexplored. To date, no study has systematically examined the comparative benefits of local versus global temporal integration within GNN architectures, despite their natural capacity to support both perspectives. Third, transitions between events, such as from \textit{complete request} to \textit{close case}, or from \textit{complete request} to \textit{append close case}, often carry meaningful semantic differences but are rarely modeled explicitly. This omission limits the model’s ability to differentiate between sequences that are structurally similar but behaviorally distinct. Fourth, many traditional machine learning benchmarks simplify event sequences by collapsing repeated events and ignoring substatus labels (e.g., \textit{lifecycle: transition}). This reduction flattens the prediction space and overlooks common real world behaviors such as retries, escalations, and partial completions \citep{tama2019empirical}. While recent deep learning models have begun to jointly predict both event type and substatus \citep{rama2023embedding}, this practice is still not widespread and remains absent from many comparative evaluations. Finally, GNNs are still infrequently applied to PBPM \citep{wang2025hgcn} compared to recurrent or transformer based sequence models \citep{dissegna2025graph}. This underuse misses the opportunity to jointly model temporal progression and relational dependencies in event log data.

To address these limitations, we propose a unified and interpretable graph neural network framework for PBPM that advances the state of the art along three methodological axes:

\begin{itemize}[leftmargin=*]
\item \textbf{From Local Prefixes to Global Graphs:} We bridge the gap between prefix based Graph Convolutional Network (GCN) models and full sequence Graph Attention Network (GAT) models by systematically comparing their strengths and limitations. Our findings show that local models often fail to capture long range dependencies, while properly structured global models can recover these patterns and achieve improved predictive accuracy.

\item \textbf{From Static Edge Weights to Dynamic Temporal Attention:} To mitigate the influence of temporally distant and low signal events, we introduce a time decay mechanism that shapes attention based on proximity to the prediction point. This results in dynamic attention windows that adapt to sequence length and temporal dispersion, prioritizing recency while preserving informative historical signals.

\item \textbf{From Raw Transitions to Semantic Edge Embeddings:} We encode transition type semantics into edge representations, enabling the model to distinguish between structurally similar but behaviorally distinct paths. This supports more nuanced reasoning over process dynamics and enhances both prediction performance and interpretability.
\end{itemize}

Beyond architectural innovations, our model introduces a key practical advancement: the integration of \textbf{multilevel interpretability} through diverse visualizations designed for different analytical purposes. These visual tools allow domain experts to trace model behavior, understand attention dynamics, and assess prediction rationale in a transparent manner.

We validate our approach across six public benchmark datasets, where our models outperform existing baselines in both Top-k accuracy and sequence level alignment. Importantly, the interpretability analysis not only highlights predictive performance but also reveals the underlying decision logic, positioning the framework as both a forecasting solution and a diagnostic tool for process evaluation.

The remainder of this paper is structured as follows. Section \ref{sec:RW} reviews related work. Section \ref{sec:PD} introduces key preliminaries and definitions. Section \ref{sec:method} details our modeling pipeline and presents the proposed architectures. Sections \ref{sec:Exp} and \ref{sec:RS} describe the experimental setup and report the results. Finally, Section \ref{sec:conclusion} summarizes the contributions, discusses practical implications for business process optimization, and outlines limitations and directions for future research.

\section{Related Work}
\label{sec:RW}
Predictive Business Process Monitoring (PBPM) has progressed through successive methodological phases, each addressing prior limitations while introducing new challenges. The field has evolved from statistical and classical machine learning models \citep{di2022predictive} (e.g., decision trees \citep{de2016general}, random forests \citep{di2016clustering}, and regressions \citep{van2012process}) to neural architectures \citep{metzger2014comparing, park2020predicting} to better handle complex, long running processes. More recent studies highlight the rapid adoption of deep learning and graph based approaches \citep{deng2024enhancing}, reflecting a shift toward models that can capture temporal irregularities \citep{hennig2025leveraging}, structural dependencies, and contextual attributes at scale \citep{fioretto2025comparative}. This progression contextualizes the emergence of graph neural network approaches and clarifies where significant gaps remain, particularly in temporal modeling and transition semantics.

\subsection{From Prefix to Global Models}
Early approaches in PBPM modeled each case as a sequence prefix and applied deep learning models such as Long Short Term Memory networks (LSTM), Generative Adversarial Networks (GANs), Memory Augmented Neural Networks (MANN), and Convolutional Neural Networks (CNNs) \citep{tax2017predictive, camargo2019learning, di2017eye, taymouri2020predictive, khan2018memory, pasquadibisceglie2019using, di2019activity, sun2024next} to predict upcoming events. While effective on short sequences, these models struggle with long traces containing loops or concurrency \citep{wang2025comprehensive}, due to limited memory capacity and reliance on fixed prefix windows. To improve structural expressiveness, prefix based graph models such as Gated Graph Neural Networks (GGNNs) were proposed \citep{weinzierl2021exploring}, which encode direct event relationships within localized process segments. However, these models remain restricted to local context and are unable to dynamically reweight distant but informative events outside the prefix.

Beyond accuracy, prefix-based models also faced issues of prefix bias, where early predictions benefited from dense supervision but later stages suffered from data sparsity \citep{pauwels2021incremental}. Several works attempted to mitigate this with prefix augmentation \citep{pfeiffer2025learning} or sampling strategies \citep{rama2024towards}, yet performance remained sensitive to prefix length \citep{klinkmuller2018towards} and trace variability \citep{stevens2025generating}. Moreover, while CNNs and GAN \citep{hoffmann2021progan} based models captured local sequential patterns effectively, they often failed to generalize under long-range dependencies or concurrent branches. These challenges highlighted the need for architectures that process the entire trace as a structured object.

Building on this insight, researchers have developed global sequence architectures that process entire traces simultaneously. Transformer based models \citep{bukhsh2021processtransformer, wuyts2024sutran, amiri2024pgtnet} leverage self attention mechanisms to capture long range dependencies, demonstrating improved performance over prefix based methods. Hierarchical variants \citep{ni2023predictive} segment event sequences to capture multilevel semantics, addressing the complexity of nested or multigranular process structures. In parallel, global graph based approaches provide structured alternatives. For example, \citet{pasquadibisceglie2024prophet} apply heterogeneous Graph Attention Networks (GATs) to model interactions across multiple edge types, while \citet{rama2023embedding} combine Recurrent Neural Networks (RNNs) with GCNs to jointly capture temporal and structural dependencies. Although these models improve performance, they still lack clearly defined and explainable mechanisms for the dynamics among node positions, temporal relevance and semantic transitions.

\subsection{Temporal Dynamics in Process Prediction}
While many models treat transitions between events uniformly, recent graph based approaches have begun to emphasize their semantic diversity and predictive importance. \citet{pasquadibisceglie2024prophet} incorporate multi type edges to distinguish between control flow and attribute based connections, demonstrating that transition semantics contribute significantly to prediction accuracy. \citet{ayaz2024semi} apply graph autoencoders with edge conditioned convolutions to encode transition semantics, showing that different transition types require distinct representations. \citet{lischka2025directly} preserve transition multiplicity using multi edge graphs and propose edge centric GNNs that treat transitions as first class entities. Despite these advances, the integration of semantic transition information into attention mechanisms remains rare and underexplored in current literature.

Recent work has also highlighted the role of temporal stability \citep{teinemaa2018temporal} in shaping prediction accuracy, with studies showing that elapsed time between events carries complementary information beyond structural order \citep{van2011time}. Models that incorporate time aware embeddings \citep{nguyen2020time} demonstrate clear benefits in handling heterogeneous case duration and non-uniform inter-event gaps \citep{verenich2019survey}. Yet, most existing approaches either treat time and transition semantics in isolation or apply them heuristically, leaving open the question of how to jointly integrate temporal relevance and semantic diversity into a unified attention framework.

\subsection{Interpretability in PBPM}
As predictive models in PBPM become more complex, interpretability has become essential for stakeholder trust and actionable insight. Early efforts relied on post hoc explainability methods such as surrogate models and feature attribution techniques \citep{mehdiyev2021explainable, el2022xai}, but these typically offered limited visibility into the actual decision making processes of deep neural models. Recent research has shifted toward exploiting intrinsic interpretability within graph neural networks and attention based architectures. The PROPHET framework \citep{pasquadibisceglie2024prophet} integrates GNNExplainer to identify influential nodes and edges, while \citet{kappel2025explaining} analyze attention weights in Transformer models to visualize control flow relevance across traces. Despite these developments, interpretability in GNN models for PBPM remains underdeveloped. Most existing work offers limited transparency into how temporal or structural signals shape predictions, and few approaches provide multilevel explanations that support both technical model validation and process level decision support.

To address the above limitations, we propose a unified and interpretable graph neural network based framework that integrates temporal decay mechanisms and transition semantics. Our approach includes both prefix based and global models, enabling a systematic comparison between local subgraph and full trace input strategies.

\section{Preliminaries and Definitions}
\label{sec:PD}
To support our graph based modeling framework, we represent each business process trace as a \textbf{directed attributed graph} that encodes both event level and sequence level information for downstream prediction tasks.

\begin{definition}[Event-Sequence Graph Construction]
\label{def:GrapStr}
Given a process log, an event trace (sequence) $G_j$ is represented as a directed attributed graph object:
\[
\mathcal{G}_{j} = (\mathbb{I}_{j},\mathbb{V}_{j}, \mathbb{W}_{j}, \mathbb{E}_{j},  \boldsymbol{\delta}_{j}),
\]
where:
\begin{itemize}[label=\raisebox{0.25ex}{\tiny$\bullet$}]
    \item $\mathbb{I}_{j}$: Event (node) labels,
    \item $\mathbb{V}_{j}$: Node level attributes,
    \item $\mathbb{W}_{j}$: Edge attributes,
    \item $\mathbb{E}_{j}$: Directed edge indices,
    \item $\boldsymbol{\delta}_{j}$: Global temporal distances.
\end{itemize}
\end{definition}

We also define a set of \textbf{sequence level attributes} $\mathbb{F}_{j}$ (e.g., process instance metadata or aggregated statistics) as a separate input to preserve global context.

Unlike prior PBPM graph formulations that focus only on event labels and direct-flow connectivity, our construction augments edges with both local and global temporal attributes as well as semantic embeddings. Local temporal gaps are represented as edge weights, while global temporal distances are encoded as a separate vector aligned with all edges. In addition, categorical transition types are embedded into higher-dimensional vectors, making edge attributes multi-dimensional rather than scalar. Together with the inclusion of sequence-level features $\mathbb{F}_j$, this design preserves global context while enriching the expressive power of the event-sequence graph.

\begin{definition}[Event Label Vector]\label{def:EventLabel}
Each node $X_i$ in $\mathcal{G}_j$ is assigned a discrete event label drawn from a finite vocabulary of activity types $\mathcal{A}$ and optional substatus types (\textit{lifecycle: transition}) $\varsigma$. These two categorical variables are combined into a joint label: $\mathcal{A}_i' = (\mathcal{A}_i, \varsigma_i)$
which is mapped to a unique integer identifier via an encoding function $\mathcal{I}(\cdot)$. The resulting event label vector for graph $\mathcal{G}_j$ is $\mathbb{I}_{j} = [\mathcal{I}(\mathcal{A}_1'), \mathcal{I}(\mathcal{A}_2'), \dots, \mathcal{I}(\mathcal{A}_n')] \in \mathbb{N}^n$, where $n$ is the number of events in the trace.
\end{definition}

The activity label set $\mathcal{A}$ contains $m$ distinct activity types, and the substatus set $\varsigma$ contains $k$ unique transition states. The joint label space is defined as $\mathcal{A}' = \mathcal{A} \times \varsigma$, yielding up to $m \cdot k$ possible combinations. To support next event prediction, we append a special token $\texttt{eos}$ representing the end of sequence. This extends each label vector to length $n + 1$ and increases the total label space to $m \cdot k + 1$. The resulting shifted label sequence is used in both decoder style and full sequence prediction settings.

\begin{definition}[Node Attribute Vector]\label{def:NodeAttrs}
Each node $X_i$ in graph $\mathcal{G}_j$ is associated with an attribute vector:
$$
\mathbf{v}_{X_i} = [U_i]
$$
where $U_i$ includes categorical and numerical event-level features, excluding the activity and substatus labels.
\end{definition}

Categorical attributes are encoded using one-hot vectors, while numerical attributes are normalized to $[0,1]$ via min-max scaling. Missing values are imputed using the median for numerical attributes and a reserved token (e.g., $-1$) for categorical types.

The resulting node attribute matrix for graph $\mathcal{G}_j$ is:
$$
\mathbb{V}_{\mathcal{G}_j} =
\begin{bmatrix}
\mathbf{v}_{X_1} \\
\vdots \\
\mathbf{v}_{X_n}
\end{bmatrix}
\in \mathbb{R}^{n \times d_U}
$$
where $d_U$ is the dimensionality of $U_i$, and $n$ is the number of events in the trace.

\begin{definition}[Edge Attributes Vector]
\label{def:EdgeAttrs}
Each edge $(X_i \rightarrow X_{i+1})$ in a graph $\mathcal{G}_{j}$ is associated with an edge attribute vector $\mathbf{w}_{(i \rightarrow i+1)}$, which may include one or both of the following components, depending on the model configuration.
\end{definition} We define two types of edge level features used throughout our models:
\begin{itemize}[leftmargin=*]
\item \textbf{Temporal Difference:}\label{def:EdgeAttrs-time} 
Each edge includes a scalar feature $\Delta t_{i \rightarrow i+1}$ representing the time elapsed between two consecutive events:
$\Delta t_{i \rightarrow i+1} = T_{i+1} - T_i$,
where $T_i$ and $T_{i+1}$ denote the timestamps of events $X_i$ and $X_{i+1}$, respectively. All temporal values are min-max normalized within each trace.
\item \textbf{Transition type:}\label{def:EdgeAttrs-edge} An optional semantic transition label $\tau_{i \rightarrow i+1}$ encodes the categorical change in activity (e.g., ``Submit → Approve''). This is embedded as a learnable vector:
$\tau_{i \rightarrow i+1} \in \mathbb{R}^{k}$,
where $k$ is the dimensionality of the transition embedding space.
\end{itemize} 

For each graph $\mathcal{G}_{j}$ with $n$ nodes, the edge attribute tensor is defined as:
$\mathbb{W}_{j} \in \mathbb{R}^{(n-1) \times d_E}$
where $d_E \in \{1, k, 1+k\}$ depending on whether the model uses only temporal differences, only transition types, or both features. Each row $\mathbb{W}_{\mathcal{G}j}^i = \mathbf{w}{(i \rightarrow i+1)}$ corresponds to the edge attribute vector for the $i$-th transition.

\begin{definition}[Edge Index] \label{def:EdID}
The edge index $\mathbb{E}_{j} \in \mathbb{N}^{2 \times (n-1)}$ encodes the directed connections between consecutive nodes in the event sequence. \end{definition}
Each column defines a single edge as a source-target pair:
$$
e_{(i \rightarrow i+1)} = \mathbb{E}_{j}^{(:,i)} = 
\begin{bmatrix}
i \\
i+1
\end{bmatrix}
\quad \text{for } i = 1, \ldots, n-1
$$

\begin{definition}[Global Temporal Distance Vector]\label{def:time}
For each event $X_i$, we define its relative global temporal distance from the end of the sequence as: $\delta_i = T_{\text{final}} - T_i$
\end{definition}

This vector $\boldsymbol{\delta}_{j} \in \mathbb{R}^{n-1}$ is min-max normalized within each sequence and used in time decay aware attention models to emphasize events closer to the prediction point.

\begin{definition}[Graph Attribute Vector]\label{def:GraphVector}
Each graph $\mathcal{G}_j$ is associated with a global attribute vector $\mathbb{F}_j \in \mathbb{R}^{d_G}$, representing sequence level characteristics such as case category, total duration, or resource utilization.
\end{definition}

\begin{definition}[Prefix Subgraph]\label{def:PrefixSubgraph}
Given a graph $\mathcal{G}_j$ with $n$ events, a prefix of length $l < n$ is defined as:
$$
\mathcal{G}_j^{(p)} = (\mathbb{I}_j^{(p)}, \mathbb{V}_j^{(p)}, \mathbb{W}_j^{(p)}, \mathbb{E}_j^{(p)}, \boldsymbol{\delta}_j^{(p)}),
$$
where each component corresponds to the truncated structure of the original graph up to the $p$-th event.
\end{definition}

Each prefix subgraph inherits the structural, temporal, and semantic encoding of the original trace. It is labeled with the next event and its corresponding substatus for supervised learning. We evaluate predictive performance across multiple prefix lengths: $p \in \{2, 3, 5, 7, 10, 15, 20\}$, reflecting different stages of process execution.

\section{Research Pipeline and Methodology} \label{sec:method}
\subsection{Modeling Pipeline Overview}\label{sec:pipeline}
\begin{figure}[htbp!]
    \centering
     \caption{Overview of the proposed framework. Event logs are transformed into attributed event-sequence graphs capturing local time gaps ($\Delta t$), global time decay ($\boldsymbol{\delta}$), and transition semantics ($\tau$). Prefix-based GCNs model local temporal dependencies across multiple prefix sizes, while full-trace GATs progressively incorporate temporal and semantic cues through layered attention mechanisms. The outputs correspond to next-event predictions for GCNs and sequence of next-event generation for GATs, with interpretability provided in the dynamic temporal (decay) based variants.}
  \label{fig:pipe}
  \includegraphics[width=\textwidth]{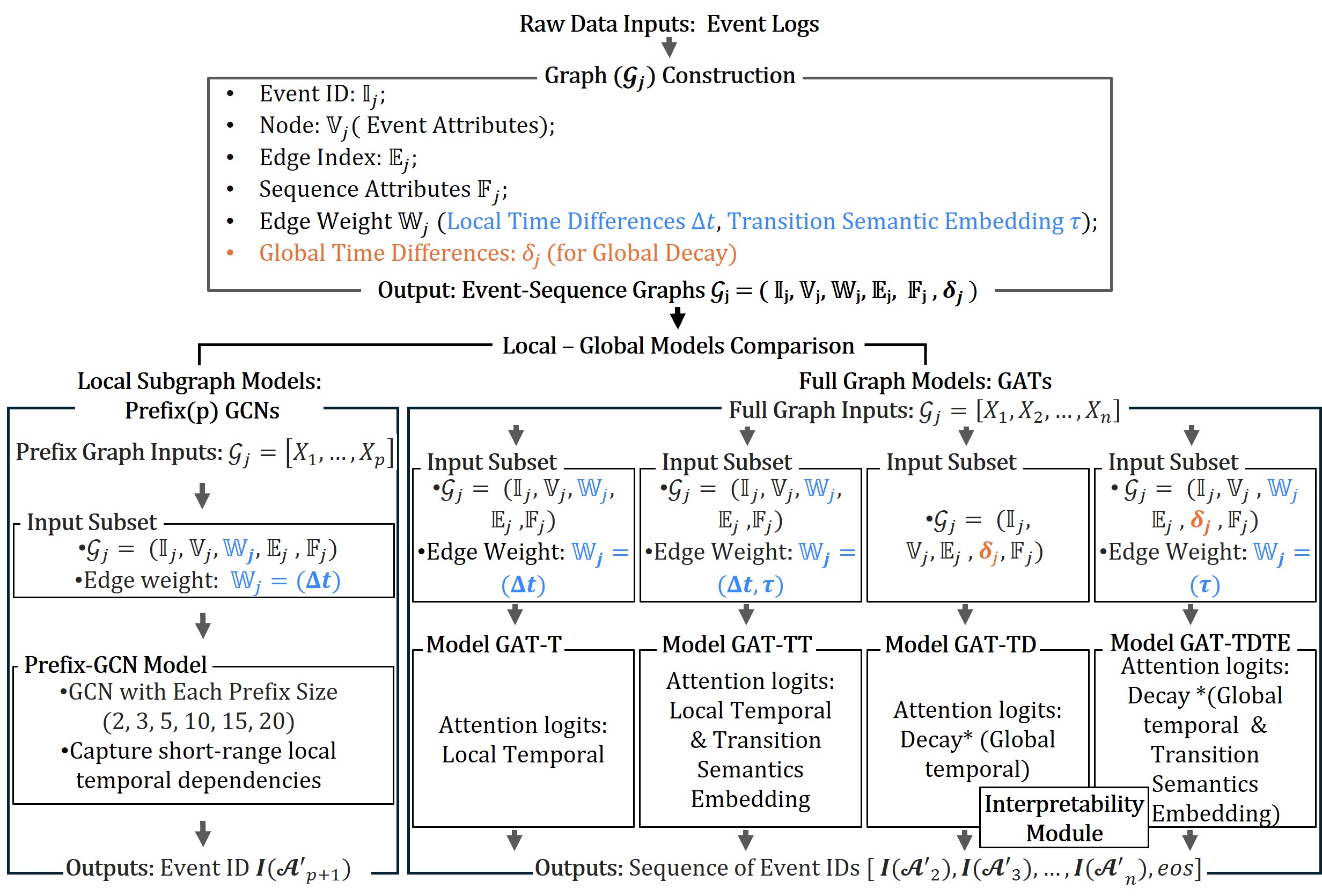}
\end{figure}

Our modeling framework evolves through a four stage pipeline, each stage progressively enhancing the models' capacity to capture temporal and semantic dependencies in event sequences. Figure~\ref{fig:pipe} provides a global overview of the unified architecture, highlighting how local–global comparison, temporal decay, and transition semantics are integrated across stages.
\begin{itemize}[leftmargin=*]
\item \textbf{Stage 1: GCN on Prefix Subgraphs.} We begin with localized learning on truncated trace graphs (Defination \autoref{def:PrefixSubgraph}) using GCNs. Edges encode local time differences $\Delta t_{i \rightarrow i+1}$, enabling next event prediction based on short prefixes. While serving as baselines for further modeling, this setup is suited for early stage monitoring with partial information and captures local dependencies effectively.

\item \textbf{Stage 2: GAT with Local Temporal and Transition Semantics.} To move beyond the constraints of prefix modeling, we employ GATs over full trace graphs. While the model has access to the entire event sequence, attention is shaped solely by local edge time gaps $\Delta t_{i \rightarrow i+1}$ and optional transition type edge embeddings $\mathbb{W}_{\mathcal{G}_j}^i$. This setup supports step wise next event prediction from the complete trace context, capturing long range dependencies and contextual structure.

\item \textbf{Stage 3: GAT with Dynamic Temporal Attention.} To model the recency effect in temporal processes, we apply an exponential decay function over each node's global temporal distance $\delta_i$, generating dynamic, case specific attention that prioritizes events based on actual elapsed time rather than fixed positions. When combined with semantic edge embeddings $\mathbb{W}_{\mathcal{G}_j}^i$, the attention mechanism becomes sensitive to both temporal proximity and transition semantics, enhancing robustness in traces with noise, loops, or repeated behaviors.

\item \textbf{Stage 4: Interpretability Module.} We extract attention related outputs and procedures to produce heatmaps, node influence rankings, critical window plots, and summary metrics of temporal attention dynamics. Additionally, we present a Temporal Attention Decomposition case study to illustrate how attention shifts over time. These visualizations reveal which events and transitions influence predictions, providing transparent insights for stakeholders and supporting model diagnosis and process improvement.
\end{itemize}

\subsection{Graph Convolutional Networks with Prefix Subgraphs}
\label{GCN}

\subsubsection{Graph Convolutional Networks}
Graph Convolutional Networks (GCNs) operate on graph structured data by iteratively aggregating information from a node’s neighbors to update its representation. In this work, we apply GCNs to \textit{prefix subgraphs} (Definition~\ref{def:PrefixSubgraph}) to support next event prediction based on partial trace observations.

We adopt the GCNConv operator proposed by \citet{kipf2016semi}, which uses a normalized adjacency matrix to stabilize message passing. The propagation rule at layer $l$ is adapted to our edge weighted, attributed graphs as follows:

\begin{equation}
\mathbb{V}^{(l+1)} = \sigma \left( \tilde{D}^{-\frac{1}{2}} \tilde{A}_w \tilde{D}^{-\frac{1}{2}} \mathbb{V}^{(l)} W^{(l)} \right)
\end{equation}

where:
\begin{itemize}[label=\raisebox{0.25ex}{\tiny$\bullet$}]
    \item $\mathbb{V}^{(l)} \in \mathbb{R}^{n \times d^{(l)}}$ is the node feature matrix at layer $l$,
    \item $W^{(l)}$ is the learnable weight matrix,
    \item $\sigma$ is a non-linear activation function,
    \item $\tilde{A}_w = A_w + I$ is the edge weighted adjacency matrix with self loops,
    \item $A_w \in \mathbb{R}^{n \times n}$ is the directed adjacency matrix where $A_{w_{i,i+1}} = w_{(i \rightarrow i+1)}$,
    \item $\tilde{D}$ is the diagonal degree matrix of $\tilde{A}_w$.
\end{itemize}

\subsubsection{Baseline Prefix GCN Model}
\label{sec:baseline_prefix}

\paragraph{Input and Output Structure}
For each graph-based sequence $\mathcal{G}_j$ with $n$ events, we generate supervised training samples by slicing the trace into prefix subgraphs of length $p < n$ (Definition~\ref{def:PrefixSubgraph}). Each training pair is defined as:
$$
[X_1, X_2, \dots, X_p] \rightarrow \mathcal{I}(\mathcal{A}'_{p+1}),
$$
where each node $X_i = [\mathcal{I}(\mathcal{A}'_i), \mathbf{v}_{X_i}]$ includes the encoded event label and its associated attribute vector. If $p = n$, a special \texttt{eos} token is appended to represent end-of-sequence prediction. This formulation enables training across varying prefix lengths and supports early prediction use cases.

The baseline model operates on each prefix graph $\mathcal{G}_j^{(p)}$ with edge attributes $\mathbb{W}_{j}^{(p)}$ containing only the local temporal difference $\Delta t_{i \rightarrow i+1}$.

\paragraph{Model Architecture} 
The model follows a three stream GCN architecture. First, the embedded node label vector $\mathbb{I}_{j}^{(p)}$ and the node attribute matrix $\mathbb{V}_{j}^{(p)}$ are independently processed by two parallel GCN layers, yielding representations $d \in \mathbb{R}^{n \times d_{\mathbb{I}}}$ and $f \in \mathbb{R}^{n \times d_{\mathbb{V}}}$, respectively. These are concatenated into $[d \Vert f]$ and passed through a third GCN layer to produce updated node embeddings. A mean pooling layer aggregates the node embeddings into a graph level vector: $\mathbf{z}_{j} = \texttt{MeanPool}(d \Vert f)$. In parallel, the sequence level attribute vector $\mathbb{F}_{j} \in \mathbb{R}^{d_G}$ is projected via a linear layer: $
\mathbb{F}_{j}^{d} = \texttt{Linear}(\mathbb{F}_{j}) \in \mathbb{R}^{d_F}$. The projected sequence level vector is concatenated with the pooled graph representation to form the final vector: $\mathcal{Z}_{j} = [\mathbf{z}_{j} \Vert \mathbb{F}_{j}^{d}] \in \mathbb{R}^{d_z + d_F}$. The combined vector is passed through a two layer feedforward network for prediction: $
\hat{y}_{j} = \texttt{MLP}(\mathcal{Z}_{j}) = \texttt{Linear}(\texttt{ReLU}(\mathcal{Z}_{j}))$.

\subsubsection{Summary}
The prefix based GCN baseline captures immediate sequential dependencies by learning from short, growing prefixes of event traces. It uses only local temporal differences on edges and omits full trace context, making it suitable for early prediction scenarios. Despite its simplicity, it leverages both event level attributes and sequence level features, providing a strong baseline for comparison with global attention based models.

\subsection{Graph Attention Networks with Full Sequence Graphs}
GATs enable flexible message passing by dynamically learning attention weights between connected nodes~\cite{velickovic2017graph}. Unlike GCNs, which aggregate neighbor features via fixed edge weights, GATs learn attention scores that reflect the relative importance of each neighbor during aggregation.

\subsubsection{Graph Attention Networks with Edge Aware Attention}\label{GAT-EW}
We develop two variants of Graph Attention Networks (GATs) with edge aware attention that operate on full process graphs $\mathcal{G}_j$ (Definition~\ref{def:GrapStr}), without prefix slicing.

\paragraph{Shared Structure}
In both models, each GATConv layer uses edge attributes $\mathbb{W}j$ (Definition~\ref{def:EdgeAttrs}) to compute attention coefficients $\alpha_{i,i+1}^{(l)}$ along edges $(i \to i+1)$. The unnormalized attention score at layer $l$ is computed by applying a learnable function to the concatenated transformed node features and edge attributes:
\[
\alpha_{i,i+1}^{(l)} = \mathrm{LeakyReLU} \left( \mathbf{a}^{(l) \top} [W^{(l)} \mathbf{v}_i^{(l)} \Vert W^{(l)} \mathbf{v}_{i+1}^{(l)} \Vert \mathbf{w}_{i,i+1}] \right).
\]
where:
\begin{itemize}[label=\raisebox{0.25ex}{\tiny$\bullet$}]
    \item $\mathbf{v}_i^{(l)}, \mathbf{v}_{i+1}^{(l)} \in \mathbb{R}^{d^{(l)}}$ are the node feature vectors at layer $l$,
    \item $W^{(l)}$ is the learnable linear transformation at layer $l$,
    \item $\mathbf{w}_{i,i+1} \in \mathbb{R}^{d_w}$ is the edge attribute vector for edge $(i \to i+1)$,
    \item $\mathbf{a}^{(l)} \in \mathbb{R}^{2d^{(l)} + d_w}$ is the learnable attention vector,
    \item $\Vert$ denotes vector concatenation,
    \item $\text{LeakyReLU}(\cdot)$ is the activation function with a negative slope (commonly 0.2).
\end{itemize}

Since each node has only one neighbor, the normalized attention coefficient reduces to $\alpha_{i,i+1}^{(l)} = 1$. The node update rule simplifies accordingly:
\[
\mathbf{v}_i^{(l+1)} = \sigma \bigl( W^{(l)} \mathbf{v}_{i+1}^{(l)} \bigr).
\]
where $\sigma$ is a non-linear activation.

\paragraph{Temporal GAT (GAT-T)}  
In this variant, the edge attribute vector \(\mathbf{w}_{(i \rightarrow i+1)}\) includes only the normalized local temporal difference (Definition~\ref{def:EdgeAttrs-time}). This scalar time gap informs the attention computation, enabling the model to capture temporal dependencies during message propagation.

\paragraph{Temporal and Transition Embedding GAT (GAT-TT)}  
This variant augments the edge attribute vector with a semantic transition embedding:
$$
\mathbf{w}_{(i \rightarrow i+1)} = [\Delta t_{i \rightarrow i+1}; \texttt{Embed}(\tau_{i \rightarrow i+1})] \in \mathbb{R}^{1 + d_\tau}, $$
where \(\tau_{i \rightarrow i+1}\) is a categorical transition label and \(\texttt{Embed}(\cdot)\) is a learnable embedding function. This enriched edge representation allows the attention mechanism to consider both temporal proximity and behavioral semantics when computing attention weights.

\subsubsection{Graph Attention Networks with Dynamic Temporal Windows}\label{GAT-TW}
To model the recency effect inherent in temporal processes, we extend the standard GAT by integrating exponential decay based on each node’s global temporal distance to the prediction point. This mechanism creates dynamic, case specific attention windows that emphasize recent events while attenuating the influence of older context. We implement two variants: one using only global temporal distance, and another that jointly considers decay and semantic edge embeddings.

\paragraph{GAT with Global Temporal Decay (GAT-TD)}
This variant uses the normalized \textit{Global Temporal Distance Vector} $\delta_i = T_{\text{final}} - T_i$ (see Definition~\ref{def:time}) to modulate attention. An exponential decay term downweights the influence of temporally distant events:
\[
\alpha_{i,i+1}^{(l)} = \text{LeakyReLU}\left(\mathbf{a}^\top [\mathbf{v}_i^{(l)} \| \mathbf{v}_{i+1}^{(l)}]\right) \cdot \exp(-\lambda \cdot \delta_i)
\]

where $\lambda$ is a temporal decay coefficient, and $|\cdot|$ denotes vector concatenation. This formulation enables recency-aware attention without requiring explicit position encodings.

\paragraph{GAT with Global Temporal Decay and Transition Embedding (GAT-TDTE)}
GAT-TDTE extends GAT-TD by incorporating semantic transition embeddings:
\[
\alpha_{i,i+1}^{(l)} = \left(\text{LeakyReLU}\left(\mathbf{a}^\top [\mathbf{v}_i^{(l)} \| \mathbf{v}_{i+1}^{(l)}]\right) + \mathbf{b}^\top \texttt{Embed}(\tau_{i \rightarrow i+1})\right) \cdot \exp(-\lambda \cdot \delta_i)
\]
where $\tau_{i \rightarrow i+1}$ is the transition type (Definition~\ref{def:EdgeAttrs-edge}), $\texttt{Embed}(\cdot)$ is a learnable embedding function, and $\mathbf{a}, \mathbf{b}$ are attention projection vectors. This allows the model to modulate attention based on both temporal proximity and behavioral transition semantics.

Both GAT-TD and GAT-TDTE retain interpretability by exposing not only the attention weights $\alpha_{i,i+1}^{(l)}$ but also their decomposed components: raw attention logits, temporal decay terms, and semantic transition contributions.

\subsubsection{GAT Models with Full Graph} \label{sec:GATARch}
\paragraph{Input and Output Structure}
In our full graph GAT models, the input is the attributed graph $\mathcal{G}_{j}$ constructed from the complete sequence $[X_1, X_2, \dots, X_n]$. The output is a shifted sequence of next event labels: $[\mathcal{I}(\mathcal{A'}_2), \mathcal{I}(\mathcal{A'}_3), \dots, \mathcal{I}(\mathcal{A'}_n), \texttt{eos}]$.
This setup enables next event prediction at each position, including forecasting the end of sequence token. Unlike prefix based models, it allows training over the entire trace in a single forward pass, leveraging global attention across the full sequence.

In addition to node attributes and local time gaps ($\Delta t$) as edge weights, each directed edge is also assigned a categorical transition-type identifier ($\tau$) that encodes the semantic relation between consecutive events. These identifiers are embedded into continuous vectors before entering the attention computation, enabling the model to learn how attention dynamics shift when transition semantics are introduced. The interaction between node position, temporal attributes, and transition embeddings is further analyzed and visualized in the case study (\ref{appendix:timeline}) and through edge attention correlation analysis (Section \ref{sec:EACor}).

\paragraph{Model Architecture}
The GAT models follow a dual stream architecture. Node label embeddings and node level attributes—augmented with broadcasted graph level features—are processed in parallel through GAT layers that incorporate edge attributes. These outputs are fused and passed through an additional GAT layer with multiple attention heads. In GAT-T and GAT-TT variants, attention coefficients are computed using edge attributes encoding local time differences and, optionally, transition type embeddings. Final predictions are produced at each node via a linear layer, enabling position wise supervision without global pooling.

\subsubsection{Summary}
Prior work has shown that temporal and semantic cues capture complementary aspects of sequential behavior. Local inter-event gaps ($\Delta t$) are widely used to encode short-range ordering \citep{venkateswaran2021robust}, while transition types ($\tau$) provide semantic context that improves process modeling and predictive accuracy \citep{comuzzi2024process}. Global temporal decay ($\delta$) has also been employed to model recency effects \citep{wahid2019predictive}, ensuring that more recent events carry greater influence. Our framework builds on these insights by systematically combining these cues within GAT variants: each variant highlights a distinct factor, and the final design (GAT-TDTE) unifies them in a principled, layered progression. Empirical results in Section~\ref{sec:RS} confirm that this design yields complementary gains rather than arbitrary stacking.

Unlike prefix-based GCNs, the GAT models operate on full event sequence graphs, capturing long-range dependencies across entire traces. Within this global setting, the variants provide a controlled comparison of how temporal gaps, decay mechanisms, and semantic transitions individually and jointly shape attention.

We first establish GAT-T as a direct baseline against prefix GCNs: both use local temporal gaps $\Delta t$, but GAT-T has access to the full trace, with $\Delta t$ capturing short-range ordering. GAT-TT extends this baseline by adding transition embeddings $\tau$, so that attention reflects not only temporal proximity but also the semantic type of each edge. This semantic encoding also enhances interpretability. Because each $\tau$ corresponds to a discrete process transition category (e.g., ``approval to rejection'' or ``submission to revision''), the learned attention weights can be directly mapped to meaningful relational patterns within the process. This enables clear visualization of how edge-level semantics influence the temporal distribution of attention and reshape the model’s predictive focus (see \ref{appendix:timeline}). Similar semantic-edge explainability strategies have been shown to improve transparency in process-aware GNNs \citep{pasquadibisceglie2024prophet, lischka2025directly}. To capture the recency effect observed in temporal data, we then move to dynamic temporal windows. GAT-TD replaces local $\Delta t$ with a decay term based on global distance $\delta$, producing recency-aware attention that downweights older events and creates case-specific attention windows. Finally, GAT-TDTE combines decay and transition semantics, integrating both *when* an event occurred (recency) and *what* transition it represents (semantics).

\begin{figure}[bp!] 
  \begin{minipage}[c]{0.48\linewidth}
    \centering
    \includegraphics[width=\linewidth]{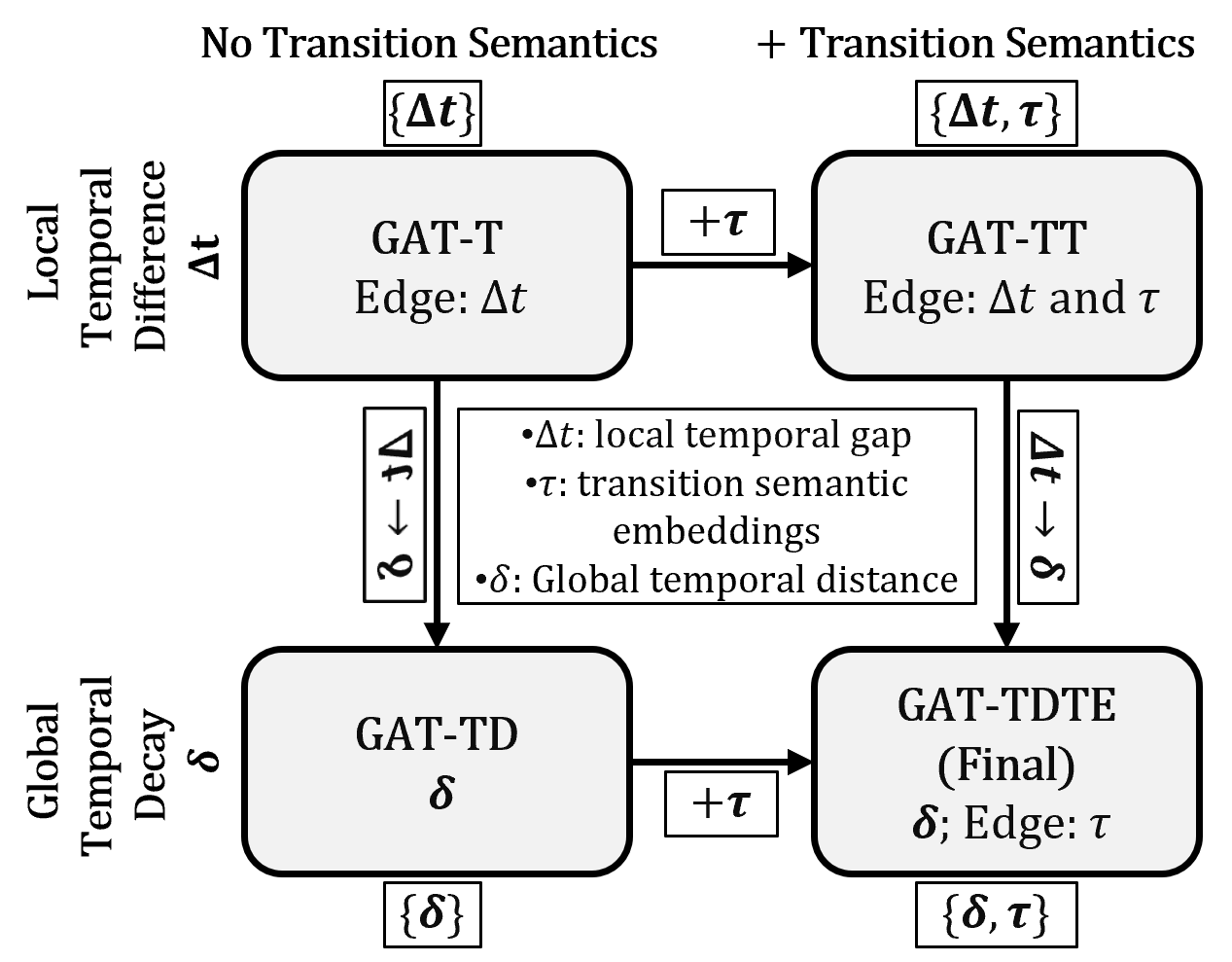}
  \end{minipage}\hfill
  \begin{minipage}[c]{0.48\linewidth}  
    \centering
    \notesize
    \setlength{\tabcolsep}{1.5pt}
    \begin{tabularx}{\linewidth}{lccc >{\raggedright\arraybackslash}X@{}}
      \multicolumn{5}{c}{Interpretability signals per variant}\\       \toprule
Variant & $\Delta t$ & $\tau$ & $\delta$ & Attention logit (pre-softmax) \\ \midrule
GAT-T     & $\checkmark$ &        &        & $z(\cdot;\,\Delta t)$ \\
GAT-TT    & $\checkmark$ & $\checkmark$ &  & $z(\cdot;\,\Delta t,\tau)$ \\
GAT-TD    &              &        & $\checkmark$ & $z(\cdot)\,e^{-\lambda\delta}$ \\
GAT-TDTE  &              & $\checkmark$ & $\checkmark$ & $(z(\cdot)+s(\tau))\,e^{-\lambda\delta}$ \\
\bottomrule
    \end{tabularx}
    \par\footnotesize
    \raggedright
\emph{Notes.} $z(\cdot)$ is the LeakyReLU logit from $[W\mathbf v_i \| W\mathbf v_{i+1}]$. Edges are $(i\!\to\! i{+}1)$. 
  \end{minipage}
  
  \caption{Variants arranged by temporal cue (rows: local $\Delta t$ vs global decay $\delta$) and transition semantics (columns: without vs with $\tau$). Arrows denote incremental extensions: GAT-T \(\xrightarrow{+\tau}\) GAT-TT; GAT-T \(\xrightarrow{\Delta t \to \delta}\) GAT-TD; GAT-TD \(\xrightarrow{+\tau}\) GAT-TDTE (final).}
  \label{fig:gat-framework}
\end{figure}

Interpretability also grows layer by layer. Each added component introduces a distinct, inspectable term in the attention logit: raw logits $z$, local gaps $\Delta t$, global decay factors $e^{-\lambda\delta}$, and semantic embeddings $\tau$. Visualizing these terms (Fig.~\ref{fig:gat-framework}) shows how attention shifts from local timing to semantics, from local to global decay, and finally to their integration. This layered design makes the dependencies among variants explicit and demonstrates that the fusion of temporal and semantic signals is structured rather than simple feature stacking.

\section{Experiments} \label{sec:Exp}
\subsection{Data Description and Preprocessing} \label{sec:DD}
We evaluate our models on six benchmark event logs: \textbf{BPI12} \citep{vanDongen2012BPI}, \textbf{BPI12W} \citep{vanDongen2012BPI},  \textbf{BPI13i} \citep{Steeman2013BPI}, \textbf{BPI13c} \citep{Steeman2013BPIclosed}, \textbf{BPI20} \citep{vanDongen2020BPI}, and \textbf{Helpdesk} \citep{Polato2017HelpDesk}. These datasets originate from real-world business processes across multiple industries. BPI12 and BPI12W record loan applications from a Dutch financial institution, with BPI12W including work item logs. BPI13i (incident) and BPI13c (closed problem) stem from Volvo IT’s service management system. The Helpdesk log captures ticket handling in an Italian software company, while BPI20 (prepaid travel cost) covers two years of travel reimbursement requests from a Dutch university, including both domestic and international trips with multi-level approvals. Together, these datasets span diverse temporal scales, process complexities, and attribute richness (Table~\ref{tab:data}), providing a comprehensive testbed for our modeling pipeline.

\begin{itemize}[leftmargin=1.5em]
\item \textbf{Temporal Variability:} The logs range from short operational windows (Helpdesk) to multi year case durations (BPI13c). This temporal diversity is essential for validating our decay based GAT models, which rely on time aware attention mechanisms that adapt across different time spans.
\item \textbf{Process Complexity:} Long and structurally diverse traces in BPI12 and BPI12W challenge the ability of global GAT models to capture long range dependencies. In contrast, the shorter traces in BPI13i and BPI13c allow for evaluating localized prefix GCNs.
\item \textbf{Attribute Richness:} The BPI13 variants include multiple event and sequence level attributes, which support our dual input fusion design. Simpler logs such as BPI12 and Helpdesk test the model's performance under sparse input conditions.
\item \textbf{High Class Cardinality:} BPI20 introduces a demanding setting with 29 prediction classes but only 2,099 traces. Its sequences are of moderate length (median 8, max 21), with minimal event-level features but several sequence-level attributes. This profile stress-tests model generalization when class space is large and per-event signal is sparse.
\item \textbf{Realism and Scale:} These real world logs exhibit class imbalance, irregular time intervals, and varying volumes (from 1.4K to over 13K traces), providing realistic challenges for prediction under both low and high data regimes.
\end{itemize}

\begin{table}[htb!]
\centering
\setlength{\tabcolsep}{3pt}
\caption{Statistics of the Datasets Used in the Experiments}
\label{tab:data}
\notesize
\begin{tabularx}{\linewidth}{cccccccccccX}
\hline
\textbf{Dataset} & \textbf{\#Seq} & \textbf{Min(len)} & \textbf{Max(len)}\footnotemark & \textbf{Med(len)} & \textbf{Min(t)} & \textbf{Max(t)} & \textbf{Med(t)} & \textbf{\#Attr(E, S)} & \textbf{\#Evt}& \textbf{\#St} & \textbf{\#Cls}\footnotemark\\
BPI12W           & 9658           & 2                 & 156               & 14                & 0               & 197538          & 12720           & (1, 1)               &7&3 & 19             \\
BPI12            & 13087          & 3                 & 175(159)              & 11                & 0               & 197538          & 1164            & (1, 1)               &24 & 3 & 36             \\
BPI13i           & 7554           & 1                 & 123(90)              & 6                 & 0               & 1110746         & 10870           & (4, 3)                &4&13& 13             \\
BPI13c           & 1487           & 1                 & 35                & 3                 & 0               & 3246981         & 118051          & (4, 3)               &4&7 & 7              \\
helpdesk         & 4580           & 2                 & 15(13)               & 4                 & 44124           & 86392           & 57397           & (1, 1)                & 14 &1 &14   \\  
BPI20				&2099		&1					&21(18)	
&8	                &0	             &467965	           &34624	&(1, 4)				&29	&1	&29   \\   
\bottomrule
\end{tabularx}
\end{table}
\footnotetext[1]{When only one sequence exists at the maximum length, we report the second-longest length in brackets (representing the maximum length found in the test dataset).}
\footnotetext[2]{len = sequence length. t = time (minutes) between first and last events. \#Attr(E, S) = number of event- and sequence-level attributes. med = median}

A key observation from our dataset analysis is the discrepancy between the number of unique event types (\#Evt) and the final classification labels (\#Cls), which results from incorporating \textit{``lifecycle:transitions''} as sub-status information (\#St). For example, in the BPI13i and BPI13c datasets, the sub status categories (13 and 7, respectively) substantially increase the label space beyond the base event vocabulary (both with 4 activity types), leading to 3.25× and 1.75× larger output spaces. Given this expanded granularity in many real-world logs, we adopt the joint event and sub status prediction formulation, more commonly used in recent PBPM graph based models \citep{rama2021deep, rama2023embedding}. This design better reflects the multi-dimensional nature of process states and supports more fine grained, actionable predictions.

All datasets are preprocessed using a unified pipeline, including min-max normalization, categorical encoding, sequence slicing with EOS labeling, etc. Details are provided in Section~\ref{sec:PD} and \ref{sec:method}.

\subsection{Experiment Setup} \label{sec:ES}
All models were implemented using PyTorch Geometric 2.5.3 \citep{FeyLenssen2019} and executed on a machine equipped with an Intel(R) Core(TM) i9-8950HK CPU (6 cores, 12 threads) and 32 GB of RAM. No GPU acceleration was employed. 

\subsubsection{Global Configuration} \label{sec:GC}
To ensure consistent comparison across all model stages, we applied a unified configuration to avoid hyperparameter induced bias. 

For the GCN prefix models, each model consists of three streams: (i) the node attribute stream uses a single GCN layer with 32 hidden units, (ii) the event label embedding is projected to 16 dimensions and passed through a GCN with 32 units, and (iii) the fusion stream combines both and applies a GCN layer with 64 units. Sequence level attributes are projected via a 32 dimensional linear layer and concatenated before the final fusion layer. Training is performed using the Adam optimizer with a learning rate of 0.001 and weight decay of $1 \times 10^{-5}$. Label smoothing (factor 0.1) is applied to improve generalization and stabilize convergence. We use fixed global batch sizes selected from {32, 64, 128, 256}, adjusted per dataset according to the number of generated prefix subgraphs( smaller prefixes produce more training samples and require larger batch sizes to maintain efficiency).

For GAT-based models, we adopt a deeper architecture to accommodate the increased model capacity. Node attributes are processed using a GAT layer with 32 hidden units, while event labels are embedded into 64 dimensional vectors and passed through a GAT layer of hidden size 128. The fusion layer uses 256 hidden units, with 4 attention heads applied consistently across all GAT layers. Transition aware models embed edge types into a 32 dimensional vector. A fixed learning rate of 0.001 and Adam optimizer are used. 

\subsubsection{Decay Coefficient ($\lambda$) Analysis} \label{sec:decayanalysis}
For time-decay models, we evaluate $\lambda$ over the fixed candidate set
$\{0.01, 0.1, 0.3, 0.7\}$ for each dataset. This strategy reflects the
temporal diversity of event logs, where case lengths and attribute ranges
vary substantially. In the reported results, we present the best-performing
configuration of $\lambda$ for each dataset. This procedure not only ensures
fair comparison across datasets but also allows us to assess the sensitivity
of performance to the decay factor, thereby clarifying the role of $\lambda$
under different temporal profiles.

\subsubsection{Evaluation Metrics} \label{sec:evaluation}
To evaluate model performance across all architectures, we report both classification accuracy and sequence alignment quality. Specifically:
\begin{itemize}[leftmargin=*]
\item \textbf{Accuracy (Acc):} Measures the proportion of test instances where the top-1 predicted label exactly matches the ground truth. For completeness, and to account for class imbalance across logs, we also report \textit{precision (P)}, \textit{recall (R)}, and \textit{weighted F1-score (F1)} for GAT models (see Table~\ref{tab:gat_combined}), though our primary focus remains on accuracy and sequence-level alignment metrics.
\item \textbf{Top-3 / Top-5 Accuracy:} Reported to reflect cases where the correct label appears among the top-3 or top-5 predictions. These metrics are especially relevant for event prediction tasks with multiple plausible outcomes.
\item \textbf{Damerau–Levenshtein Distance (DL):} For GAT based models generating full sequence predictions, we additionally compute DL distance between predicted and reference sequences. This edit-distance metric captures positional and structural deviations, providing a holistic measure of sequence level prediction quality.
\end{itemize}

All datasets are split into 80\% training and 20\% testing sets, using stratified sampling over sequence length (GAT) and event label (prefix GCN) to preserve distributional characteristics. Each model is trained for a maximum of 300 epochs with early stopping based on validation loss, using a patience threshold of 30 epochs. Reported results are averaged over three independent runs to ensure robustness and reduce variance introduced by random initialization.

\section{Result}
\label{sec:RS}
\subsection{Prefix-Based GCN Performance Analysis}
\begin{figure}[htbp!]
    \centering
     \caption{Baseline GCN Models Performance with Different Prefix lengths}
  \label{fig:prefix}
  \includegraphics[width=\textwidth]{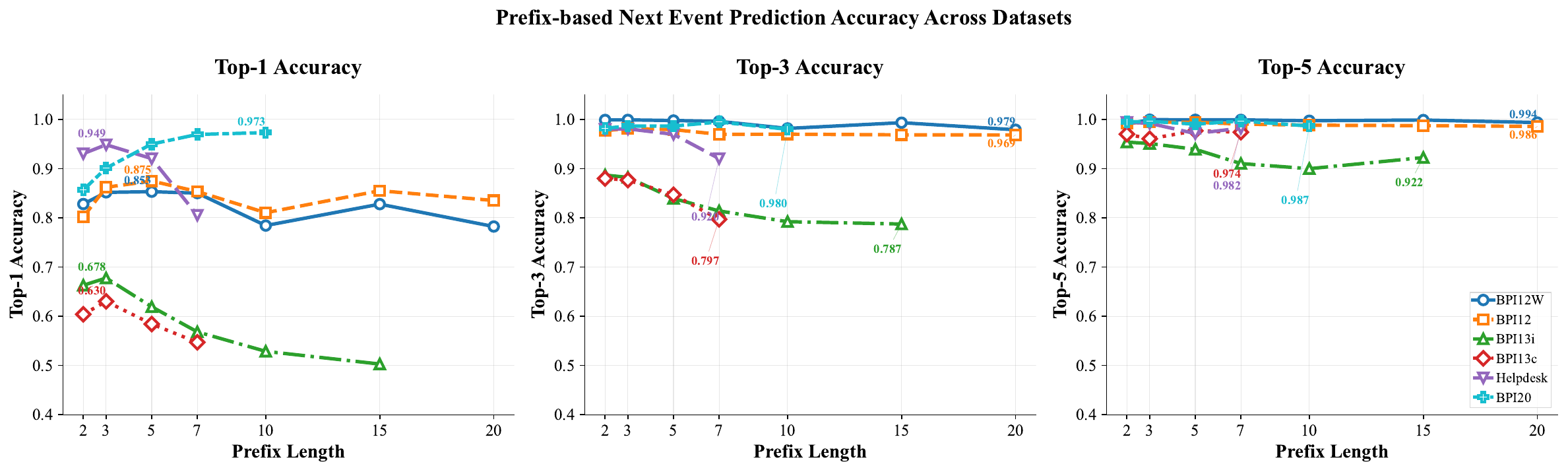}
\end{figure}

The performance of our next event prediction model was evaluated across six benchmark datasets using top-k accuracy metrics (k = 1, 3, 5), under varying prefix lengths ranging from 2 to 20 events. Figure. \ref{fig:prefix} and Table. \ref{tab:prefix_summary} present a comprehensive overview of the results, highlighting consistent trends across datasets.

\sisetup{
  round-mode          = places,
  round-precision     = 3,
  detect-weight=true,
  detect-family=true
}
\begin{table}[htbp!]
\setlength{\tabcolsep}{1pt}
\centering
\caption{Summary of Prefix(@P)-Level Coverage and Top-K (T1/T3/T5) Accuracy Across Datasets}
\notesize
\label{tab:prefix_summary}
\begin{tabularx}{\textwidth}{l *{7}{r} c S[table-format=1.3] c c}
\toprule
\textbf{Dataset} & \textbf{@P2} & \textbf{@P3} & \textbf{@P5} & \textbf{@P7} & \textbf{@P10} & \textbf{@P15} & \textbf{@P20} 
& \textbf{Max T1@P} 
& \textbf{Avg T1} & \textbf{Rng(T3)} & \textbf{Rng(T5)} \\
\midrule
BPI12w   & 160449 & 150791 & 133665 & 119235 & 100053 & 73169  & 52499  
         & 0.8748(5)      & 0.8406 & 0.969-0.982 & 0.986-0.994 \\
BPI12    & 249113 & 236026 & 213281 & 193965 & 171529 & 138985 & 109904 
         & 0.8531(5)      & 0.8253 & 0.979-0.999 & 0.994-0.999 \\
BPI13i   & 57979  & 50426  & 37101  & 27709  & 18187  & 9804   & 5656   
         & 0.6776(3)      & 0.5922 & 0.787-0.888 & 0.910-0.954 \\
BPI13c   & 5173   & 3687   & 1957   & 957    & 420    & 110    & 38     
         & 0.6301(3)      & 0.5913 & 0.797-0.880 & 0.961-0.977 \\
Helpdesk & 16641  & 12061  & 3140   & 565    & 61     & 1      & 0      
         & 0.9486(3)      & 0.9012 & 0.920-0.981 & 0.971-0.994 \\
BPI20	 &3230	&2811	  &1979	    &1179	&299 	&12	     &1				&0.9732(10)	      &0.9301	&0.982-0.995 &	0.990-0.997\\
\bottomrule
\end{tabularx}
\end{table}

Overall, most datasets achieved peak prediction accuracy with relatively short prefixes (typically 3-5 events), regardless of the underlying sequence length distributions. This indicates that the next event is most influenced by the most recent few steps in the trace, and that longer histories may introduce diminishing returns or even noise. For instance, in BPI12W, performance dropped from 0.8531 at prefix 5 to 0.7843 and 0.7823 at prefixes 10 and 20, respectively. A similar trend was observed across other datasets, reflecting two interacting factors. First, as prefix length increases, the number of available samples decreases due to fewer long traces, introducing sample bias and instability. Especially, the fixed-size output space remains the same and large(e.g., 36 for BPI12, 19 for BPI12W) Second, the GCN model lacks an attention mechanism, treating all input nodes equally. As a result, important past events may be undervalued, while noisy or less relevant events may be overemphasized.

BPI20 presents an exception: accuracy remains high at longer prefixes (e.g., 0.9732 at prefix 10), rather than dropping. This pattern reflects its structural profile—short maximum length (21 events), but relatively large median length (8 events), meaning that most traces are already moderately long. In addition, BPI20 contains 29 prediction classes and richer sequence-level attributes but minimal event-level features. These factors together make longer prefixes informative rather than noisy, since extending the input better exposes the case-level context needed to discriminate among many classes. Thus, BPI20 complements the other benchmarks by highlighting conditions where longer-range prefixes strengthen rather than weaken predictive performance.

We also observe substantial gains from Top-1 to Top-3 accuracy, particularly in datasets with more complex process structures. The average improvements from Top-1 to Top-3 were 0.151 (BPI12W), 0.113 (BPI12), 0.201 (BPI13i), 0.244 (BPI13c), 0.038 (Helpdesk), 0.058 (BPI20). This reflects greater ambiguity in next event prediction under complex workflows, where the top prediction may miss but the correct label still appears in a small candidate set. Notably, Top-5 accuracy remains high across all prefix lengths and datasets (e.g., BPI12W: 0.9995 at prefix 3, 0.9937 at prefix 20), supporting the model's utility for decision support even in low confidence contexts.

In summary, GCN model with sub graph prefix setup is most effective with short prefix lengths and delivers high Top-k accuracy across diverse process settings, though it may undervalued the differences raised from global interactions within entire sequence due to architectural limitations and data sparsity at longer prefixes.

\subsection{GAT Variants Performance Analysis}
Table \ref{tab:gat_combined} reports the performance of four GAT variants evaluated across six datasets. Metrics include Top-1/3/5 accuracy, precision, recall, weighted F1, and Damerau–Levenshtein (DL) distance. Best-performing values are highlighted for accuracy and DL measures.

\begin{table}[htbp!]
\begin{threeparttable}
\centering
\caption{Top-k accuracy, Damerau–Levenshtein distance (DL), Precision(P), Recall(R) and weighted F1 (F1) of GAT Model(M) variants. }
\setlength{\tabcolsep}{1.3pt}
\notesize
\begin{tabularx}{\linewidth}{lcccccccccccccc}
\toprule
& \multicolumn{7}{c}{\textbf{BPI12W}} & \multicolumn{7}{c}{\textbf{BPI12}} \\
\textbf{M}  & Top-1 & Top-3 & Top-5 & DL &P &R & F1 & Top-1 & Top-3 & Top-5 & DL &P &R & F1\\
\cmidrule(r){1-1} \cmidrule(r){2-8} \cmidrule(r){9-15}
T     & 0.8749 & 0.9995 & 0.9998 & \textbf{0.8590} & 0.8604	& 0.8744	 & 0.868 & 0.8541 & \textbf{0.9845} & \textbf{0.9954} & 0.7793 &0.8719 &	0.8511	& 0.8501 \\
TD    & 0.8298 & 0.9660 & 0.9884 & 0.6794 &0.818 &0.8215 & 0.8118 & 0.8408 & 0.9695 & 0.9950 & 0.7732 &0.8288 &0.832 & 0.8352 \\
TT   & \textbf{0.8763} & \textbf{0.9995} & \textbf{1.0000} & 0.8406 &0.8623 &0.8725 & \textbf{0.8694} & \textbf{0.8561} & 0.9844 & 0.9948 & \textbf{0.8179} &0.8519	 &0.8541	 &  \textbf{0.8524}\\
TDTE  & 0.8260 & 0.9598 & 0.9881 & 0.7356 &0.8383 &0.8293 &0.8156 & 0.8409 & 0.9693 & 0.9938 & 0.7775 &0.8416	&0.8364	&0.8382 \\
\midrule
 & \multicolumn{7}{c}{\textbf{BPI13i}} & \multicolumn{7}{c}{\textbf{BPI13c}} \\
\textbf{M} & Top-1 & Top-3 & Top-5 & DL &P &R & F1 & Top-1 & Top-3 & Top-5 & DL &P &R & F1\\
\cmidrule(r){1-1} \cmidrule(r){2-8} \cmidrule(r){9-15}
T     & 0.6847 & \textbf{0.9121} & 0.9706 & \textbf{0.6986} &	0.673	&	0.6847	&0.6752  & 0.6188 & \textbf{0.8665} & \textbf{0.9793} & \textbf{0.6657} &	0.624	&	0.621	&	 0.6204 \\
TD   & 0.6989 & 0.9049 & 0.9673 & 0.6895 &	0.6787	&	0.693	&	0.6807 & 0.4932 & 0.8044 & 0.9615 & 0.5069 &	0.552	&	0.5032	&	0.489 \\
TT    & 0.6922 & 0.9113 & \textbf{0.9721} & 0.6954 &	0.6847	&	0.6894	&	0.6916 & \textbf{0.6360} & 0.8637 & 0.9686 & 0.6521 &	0.5988	&	0.601	&	\textbf{0.6265} \\
TDTE  & \textbf{0.6995} & 0.8904 & 0.9577 & 0.6858 &	0.6882	&	0.6842	&	 \textbf{0.6958} & 0.4797 & 0.7852 & 0.9464 & 0.5004 &	0.5072	&	0.4912	&	0.4705 \\
\midrule
& \multicolumn{7}{c}{\textbf{Helpdesk}} & \multicolumn{7}{c}{\textbf{BPI20}} \\
\textbf{M} & Top-1 & Top-3 & Top-5 & DL &P &R & F1 & Top-1& -Top-3 & Top-5 & DL &P &R & F1\\
\cmidrule(r){1-1} \cmidrule(r){2-8} \cmidrule(r){9-15}
T     & 0.9570 & 0.9915 & \textbf{0.9967} & 0.9622 &	0.9539	&	0.9495	&	0.9482	&	0.8295	&	0.9568	&	0.9903	& 0.8184	&	0.806	&	0.8143	&	0.804 \\
TD    & 0.7531 & 0.9509 & 0.9911 & 0.7395 &	0.7782	&	0.7412	&	0.7498	&	0.7529	&	0.927	&	0.9791&	0.732	&	0.8027	&	0.7211	&	0.7458\\
TT    & \textbf{0.9584} & \textbf{0.9920} & 0.9955 & \textbf{0.9703} &	0.9586	&	0.9584	&	 \textbf{0.9582}	&	\textbf{0.8439}	&	\textbf{0.9591}	&	\textbf{0.9953}	& \textbf{0.843}	&	0.8234	&	0.8204	&	 \textbf{0.8119} \\
TDTE  & 0.7516 & 0.9457 & 0.9880 & 0.7371 &	0.7899	&	0.736	&	0.7508	&	0.7515	&	0.9056	&	0.9627	& 0.724	&	0.8024	&	0.7334	&	0.7394\\
\bottomrule
\end{tabularx}
\begin{tablenotes}
\item For GAT-TD and GAT-TDTE, we evaluate the decay coefficient at four preset values, $\lambda \in \{0.01,\,0.1,\,0.3,\,0.7\}$, on each dataset and report the best-performing value. Selected $\lambda$ per dataset: BPI12/BPI12W: 0.7; BPI13i: 0.3; BPI13c: 0.1; Helpdesk/BPI20: 0.01.The same four-value evaluation is applied uniformly across datasets to assess sensitivity; no bespoke tuning is performed.
\end{tablenotes}
\label{tab:gat_combined}
\end{threeparttable}
\end{table}

\subsubsection{Comparative Performance of GATs and Prefix-GCNs across Datasets} \label{sec:GATvsGCN}
Across five of the six datasets, at least one GAT variant outperforms prefix-GCNs in Top-1 accuracy. This improvement indicates that the attention mechanism in GAT models effectively identifies and prioritizes historically relevant events beyond the limited prefix window. By considering the full event graph, the model can capture long-range dependencies that prefix-based GCNs, limited to partial traces, may fail to exploit. The only exception is BPI20. BPI20 combines \emph{many} next-event classes (29) with a relative small sample (2,099 traces), moderate sequence length (median 8; max 21), and \emph{minimal event-level features} (versus several sequence-level attributes). In this regime, per-edge signals are thin: local time gaps and transition types carry limited discriminative power across 29 classes, so full-graph attention tends to diffuse evidence and add variance in a low-sample, high-cardinality setting. Dynamic temporal decay can also de-emphasize early approvals that remain decisive later in the case. By contrast, a \emph{long-prefix} GCN acts as a low-variance aggregator: once the prefix covers most of the trace (e.g., $P{=}10$ is near full length for BPI20), it captures the decisive late-stage context without relying on per-edge attention. This explains why short-prefix GCNs and GATs underperform on BPI20, while near-full-length GCN achieves the best Top-1 (Table~\ref{tab:prefix_summary}, Table~\ref{tab:gat_combined}). In summary, global attention helps when event-level signals are sufficiently informative to benefit from selective weighting. When class cardinality is high, samples are limited, and per-event features are sparse (as in BPI20), long-prefix GCNs remain a strong, pragmatic choice.

\subsubsection{GAT with Edge-Aware Attention: Performance Analysis}
Across all datasets, adding transition-type embeddings to edges (GAT-TT) consistently improves over time-only attention (GAT-T) in both Top-1 accuracy and weighted F1. The gains are modest but uniform ($0.1–1.6$ percentage points in accuracy across logs), confirming that semantic transitions help discriminate among plausible next steps. Weighted F1 closely tracks Top-1 trends, showing that these improvements extend beyond majority-class effects.

BPI13c illustrates this effect most clearly: while plain GAT-T lags behind prefix GCNs, 
GAT-TT recovers and attains the best GAT Top-1 and F1 on the log. This dataset has a short median sequence length of three, a wide temporal span, and low event label diversity. Under such conditions, full graph attention based solely on local time differences may offer limited utility. However, when semantic edge features are introduced, as in GAT-TT, performance improves. This suggests that effective attention in sparse and temporally stretched logs requires richer contextual information beyond temporal proximity.

BPI20 is the exception where prefix-GCNs outperform all GAT variants (see Section~\ref{sec:GATvsGCN}). Still, within the GAT family, semantics remain beneficial: GAT-TT surpasses GAT-T in both Top-1 (0.844 vs.\ 0.830) and weighted F1 (0.812 vs.\ 0.804). This indicates that while global attention is not the best paradigm under high-class cardinality, low sample size, and minimal event-level features, semantic transitions continue to strengthen per-edge signals.

In summary, the results demonstrate the advantages of full graph attention and underscore the complementary role of transition semantics in enhancing prediction accuracy, particularly in settings with temporal and structural heterogeneity.

\subsubsection{GAT with Dynamic Temporal Windows Performance Analysis}
The performance of GAT models with dynamic temporal windows demonstrates a consistent trend across datasets with varying sequence lengths and temporal characteristics. In particular, BPI13i, which has a moderate median sequence length of six events and a wide temporal span, shows the highest relative improvement when using time decay mechanisms (GAT-TD and GAT-TDTE) compared to their counterparts without decay (GAT-T and GAT-TT).

These findings suggest that dynamic temporal decay is most effective when the process sequence is of moderate length, long enough to contain informative historical events but not so long that irrelevant context introduces noise. In very short sequences, attention naturally focuses on a limited set of recent events, whereas in very long sequences, distant events may dominate due to positional bias. The inclusion of a global temporal reference through $\delta_i$ enables the model to reduce the influence of older events based on actual elapsed time rather than relative position. This temporal grounding prevents events that are close in sequence but distant in time from exerting excessive influence. In sequences of moderate length with broad temporal dispersion, the model learns to selectively attend to more temporally relevant events, confirming the benefit of decay modulated attention in improving predictive accuracy.

For long sequence datasets such as BPI12 and BPI12W, we observed a modest drop in accuracy (1.4\% to 4\%) when using decay based attention mechanisms (GAT-TD, GAT-TDTE). Although this reduction may seem counterintuitive, further analysis revealed that selecting the decay strength $\lambda$ mitigates this effect. Specifically, we found that higher decay values (e.g., $\lambda = 0.7$) yielded better results on these datasets, where long traces often include irrelevant historical noise. Rather than indicating a flaw in the mechanism, this emphasizes the importance of calibrating decay to match the temporal structure of each dataset.

Moreover, BPI12, despite its long trace lengths, exhibits bimodal duration characteristics: 25\% of cases complete in under one minute (Q1 = 0.9), while outliers extend well beyond (Q3 = 20,436.5). This heterogeneity aligns well with decay modulation and results in only a 1.4\% drop in Top-1 accuracy. In contrast, BPI12W shows more uniformly distributed case durations, where aggressive decay may excessively attenuate informative but distant events, leading to greater performance loss. These results suggest that decay is most beneficial when both temporal variance and contextual noise are present, enabling the model to filter selectively.

This interpretation is further supported by the sharp accuracy drop observed in Helpdesk and BPI13c. Both datasets contain very short sequences (median length three), where most historical events are only one or two steps away. However, despite similar sequence lengths, Helpdesk has minimal time gaps between events, whereas BPI13c features larger global time differences. Consequently, a low decay strength ($\lambda = 0.01$) works best for Helpdesk, while BPI13c benefits from a slightly higher value ($\lambda = 0.1$). In both cases, decay must be delicately selected: if too strong, it suppresses even nearby useful signals. These observations confirm that the effectiveness of temporal decay hinges on both sequence depth and absolute time dispersion.

We observe a meaningful interaction between temporal decay and transition type embeddings. In the mid length dataset BPI13i, incorporating transition type edge embeddings alongside temporal decay (GAT-TDTE) results in higher accuracy compared to using decay alone (GAT-TD). In contrast, this combination leads to performance degradation in both shorter and longer sequences. These results suggest that transition based edge features may intensify the temporal attenuation effect, which is beneficial when sequence depth and temporal dispersion are well balanced, but may suppress relevant information in regimes with limited or excessive context.

Finally, BPI20 shows limited benefit from temporal decay: with short traces, high class cardinality, and sparse event-level features, decay can further dilute already weak per-edge signals. This explains why GAT-TD and GAT-TDTE underperform their non-decay counterparts, even though $\lambda=0.01$ minimizes the loss.

Overall, these results confirm that while the exact value of $\lambda$ can shift across datasets, model performance is robust within the tested range $\{0.01, 0.1, 0.3, 0.7\}$, and the observed trends validate the role of temporal decay as a consistent mechanism rather than a dataset-specific tuning artifact.

\subsubsection{Sequence-Level Robustness} Beyond Top-1 accuracy, the Top-3 and Top-5 results remain consistently high across all datasets, frequently exceeding 0.95. This suggests that the models effectively constrain the prediction space to a small set of plausible candidates, which is advantageous in real-world settings where ranked or probabilistic outputs are more practical than single-label classification. In addition, Damerau–Levenshtein (DL) scores indicate that the predicted event sequences closely match the ground truth in terms of ordering. Although DL does not capture semantic or temporal alignment, it offers a complementary sequence-level evaluation, demonstrating that the GAT based models maintain the overall process structure alongside per event predictive accuracy.

\subsection{Interpretability Analysis on Dynamic Temporal Windows with Edge Type Awareness}
\subsubsection{Dynamic Temporal Windows} To understand how the time decay mechanism modulates attention in our models, we present three targeted visualizations based on the GAT-TD model and the BPI13i dataset. These visualizations provide representative interpretability results for time decay based architectures. Similar qualitative patterns were observed for GAT-TDTE and are included in ~\ref{appendix:GATInter}. Each visualization highlights a distinct aspect of model behavior:
\begin{itemize}[leftmargin=*]
\item \textbf{Heatmap}: Shows how attention distributes across node positions and how this distribution shifts with sequence length.
\item \textbf{Critical Windows Plot}: Identifies where the model consistently concentrates attention and whether these focal points vary meaningfully across different sequence lengths.
\item \textbf{Quantitative Summary}: Aggregates attention metrics to summarize trends in peak intensity, attention span, and positional focus.
\end{itemize}

These visualizations illustrate how the model adjusts its attention window dynamically, balancing recency and relevance in response to varying temporal structures.

\begin{figure}[htbp!]
\centering
\begin{subfigure}{\textwidth}
    \includegraphics[width=0.97\textwidth, trim={0 0 0 1cm}, clip]{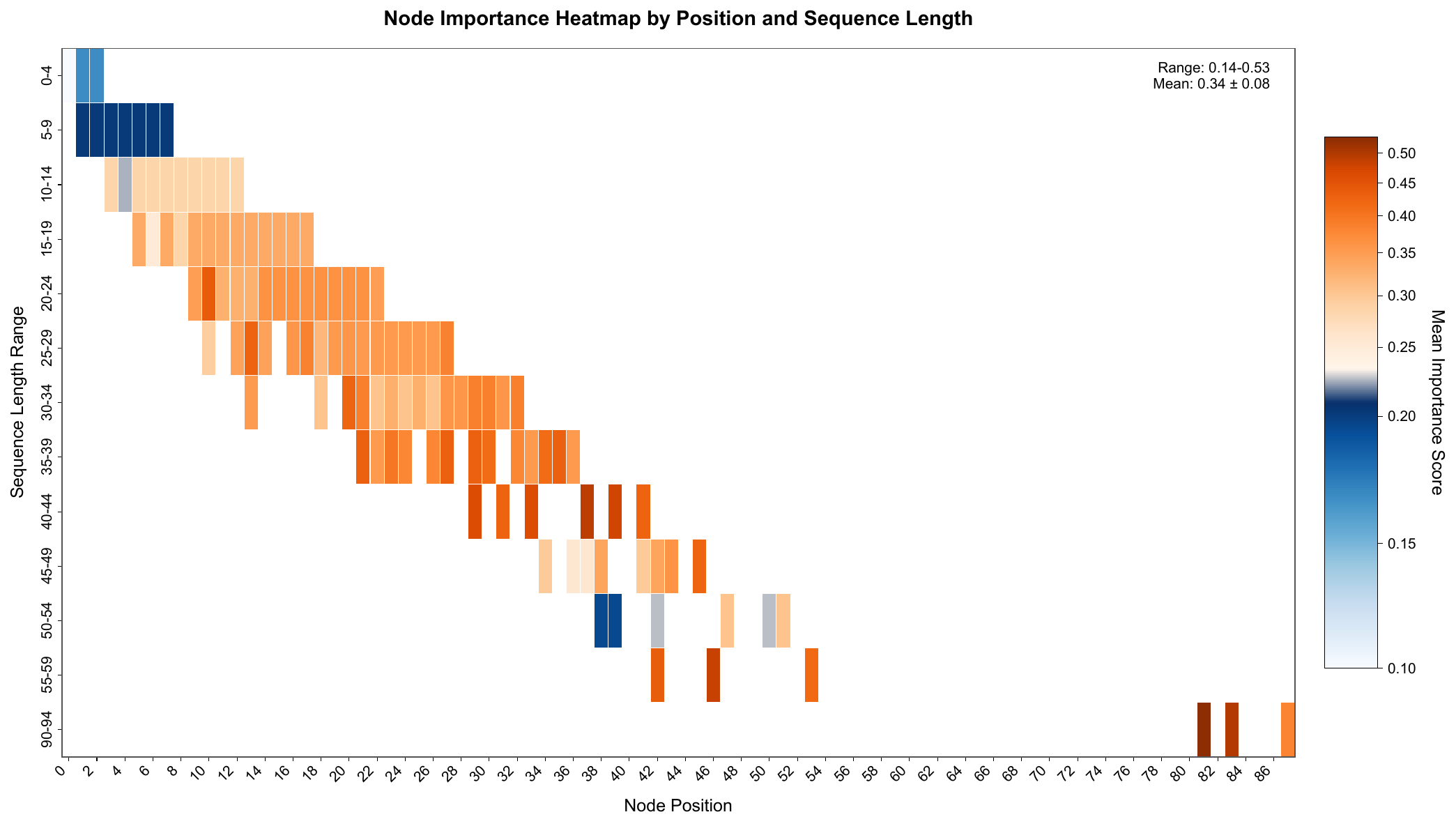}
    \subcaption{Node Importance Heatmap by Sequence Length and Position}
    \label{subfig:heatmap13i}
\end{subfigure}

\begin{subfigure}{\textwidth}
    \includegraphics[width=0.97\textwidth, trim={0 0 0 1cm}, clip]{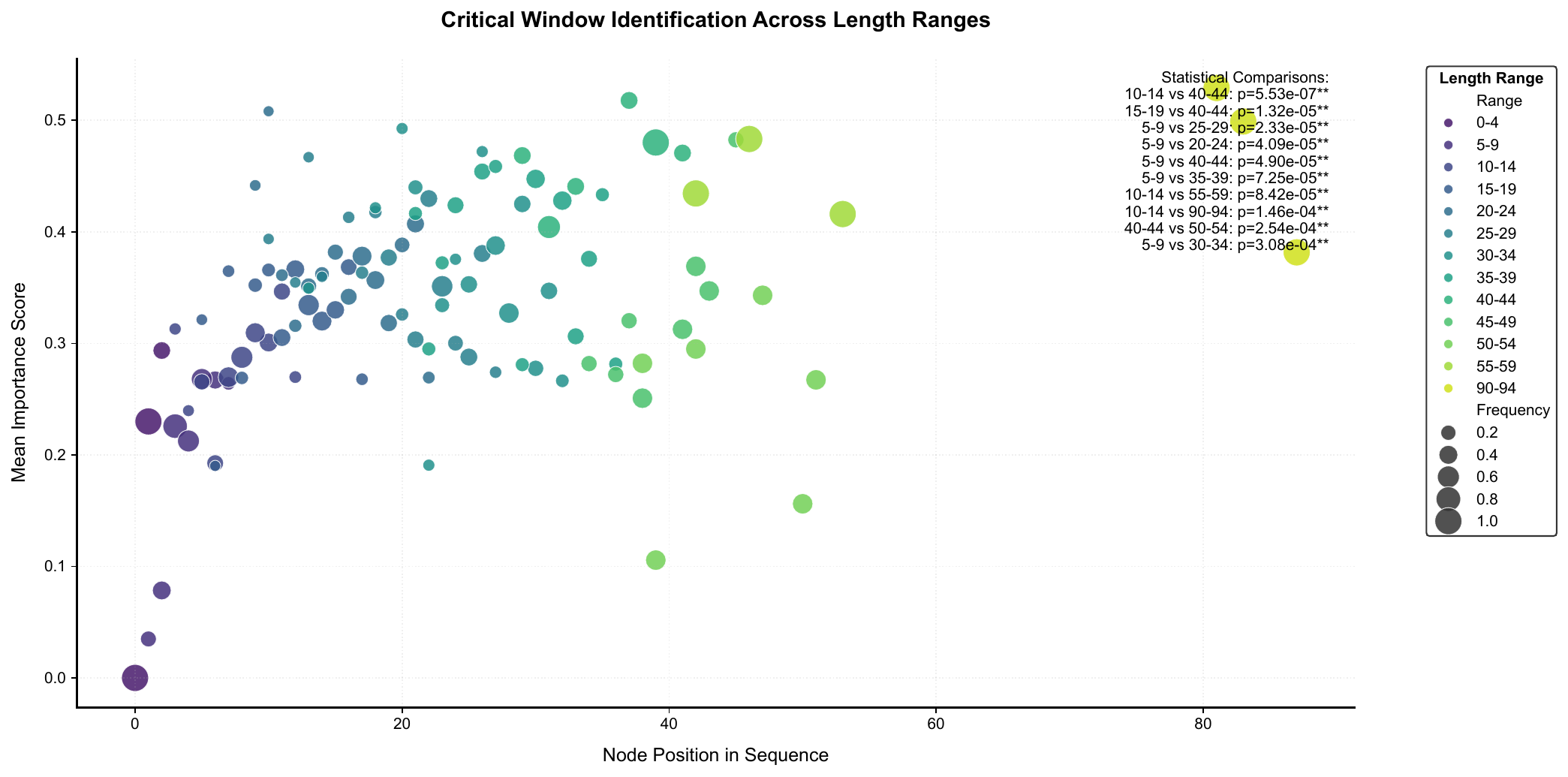}
    \subcaption{Critical Windows Identification across Length Ranges}
    \label{subfig:critwin13i}
\end{subfigure}

\begin{subfigure}{\textwidth}
    \includegraphics[width=0.97\textwidth, trim={0 0 0 1.5cm}, clip]{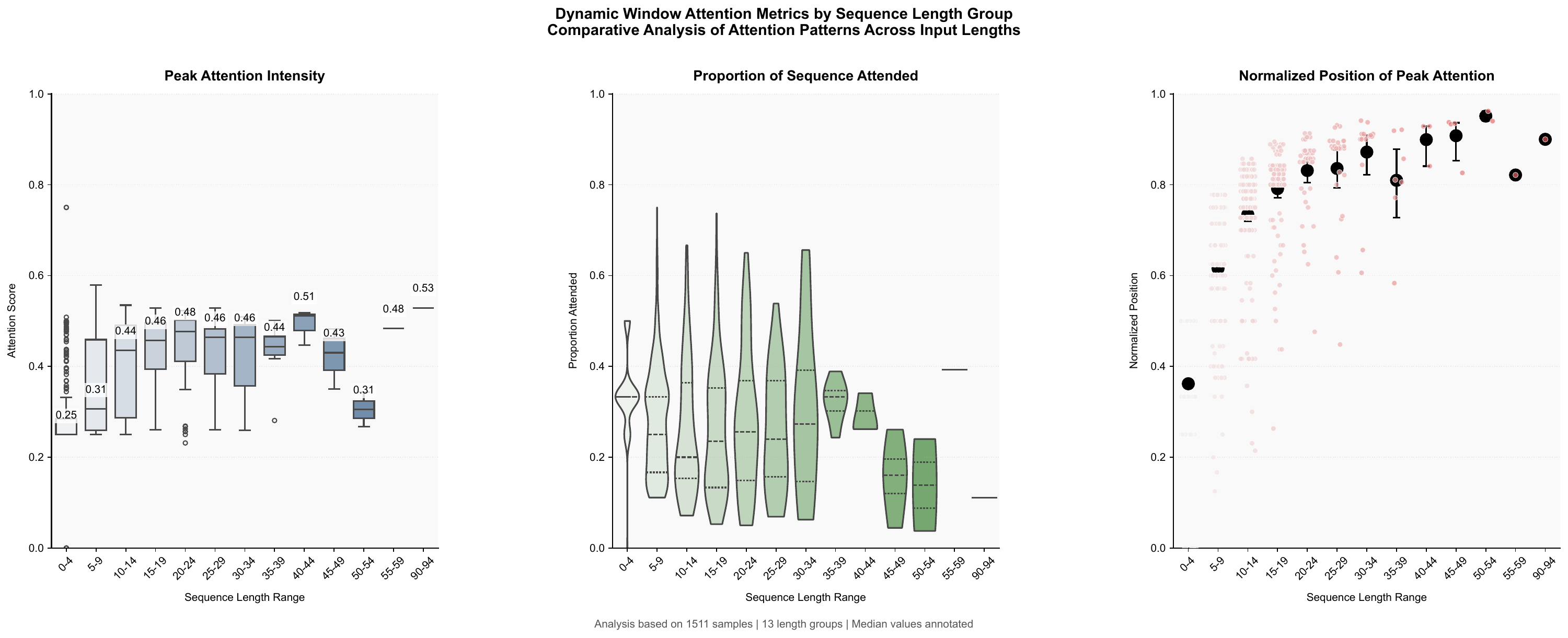}
    \subcaption{Aggregates Dynamic Temporal Windows Metrics and Comparative Analysis across Length Ranges}
    \label{subfig:attnstat13i}
\end{subfigure}

\caption{Interpretability Analysis of the GAT-TD Model on the BPI13i Dataset}
\label{fig:overall}
\end{figure}

\paragraph{Heatmap of Node Importance (Spatial Distribution)}
Figure~\ref{subfig:heatmap13i} visualizes how node importance, computed by combining attention weights with the time decay factor, varies across node positions (x axis) and sequence lengths (y axis). For interpretability, we divide sequences into discrete length bins of six events, ranging from 0 to 4 up to 90 to 94. The attention consistently concentrates on the final six to ten events, forming prediction focused windows regardless of total sequence length. For shorter sequences (10 to 29 events), attention is moderately distributed across the mid to late regions. In contrast, longer sequences (over 55 events) display sharp, localized focus on the last two to three events, with peak importance scores exceeding 0.6. This spatial shift confirms that the time decay mechanism constructs prediction optimized attention windows by amplifying recent events while attenuating distant historical context.

\paragraph{Critical Windows Analysis (Adaptive Temporal Attention)}
Figure~\ref{subfig:critwin13i} assesses the adaptability of temporal attention by visualizing each node position's average importance (y axis), frequency of Top 3 appearances (bubble size), and sequence length bin (color coded). Short sequences (0 to 4 events) show concentrated attention on starting positions (0  or 1). Medium length sequences (5 to 24 events) begin distributing attention more broadly across intermediate positions, capturing mid range dependencies. For longer sequences (25 events and above), the model selectively emphasizes specific positions near the end of the sequence, indicating an ability to isolate and prioritize informative historical signals even within extended temporal contexts. Pairwise statistical comparisons (e.g., $p = 7.93 \times 10^{-8}$ between bins 5 to 9 and 30 to 34) further confirm that attention distributions differ significantly across sequence lengths, supporting the presence of dynamic temporal adaptation rather than fixed positional biases. Detailed rank based dominance analysis (see Figure~\ref{fig:rank13i}) further substantiates these positional trends.

\paragraph{Quantitative Summary of Attention Dynamics (Aggregated Metrics)}
Figure~\ref{subfig:attnstat13i} provides a quantitative overview of attention behavior across sequence lengths using three aggregated metrics: (i) Peak attention intensity, defined as the maximum attention score, reflects the strength of the model's focus; (ii) Attention span, measured as the proportion of nodes receiving attention above 0.05, indicates the effective window size; (iii) Normalized peak position, representing the relative location of the most attended node, captures the temporal focus of attention.

The results offer clear evidence of adaptive windowing behavior. Peak attention intensity remains consistently high across all lengths (0.43 to 0.53), confirming the model's stable capacity to prioritize salient nodes. Attention span progressively narrows from approximately 34\% in short sequences to below 20\% in sequences with more than 90 events, reflecting dynamic reduction of the attention window. The normalized peak position shifts systematically toward the end of the sequence (from 0.36 to 0.87), reinforcing the model’s recency bias in longer cases. Additionally, sequences exceeding 30 events exhibit increased variability in attention span ($\text{standard deviation} = 0.18$), suggesting that selective truncation of historical context may further improve attention efficiency in extended sequences.

\paragraph{Summary} These visualizations provide strong evidence that the GAT-TD and GAT-TDTE models construct adaptive attention windows centered around prediction. The models consistently prioritize events that are both semantically relevant and temporally proximate, adjusting their attention allocation based on the temporal characteristics of each sequence. This behavior is essential for achieving robust and interpretable event prediction across diverse process settings.

\subsubsection{Edge Attention Correlation (Global Sensitivity)} \label{sec:EACor}
\begin{figure}[htbp!]
    \centering
     \caption{Correlation between Edge type Scores and Attention Weights in the GAT-TDTE model on BPI13i Dataset}
  \label{fig:corr13i}
  \includegraphics[width=\textwidth,  trim={0 0 0 0.75cm}, clip]{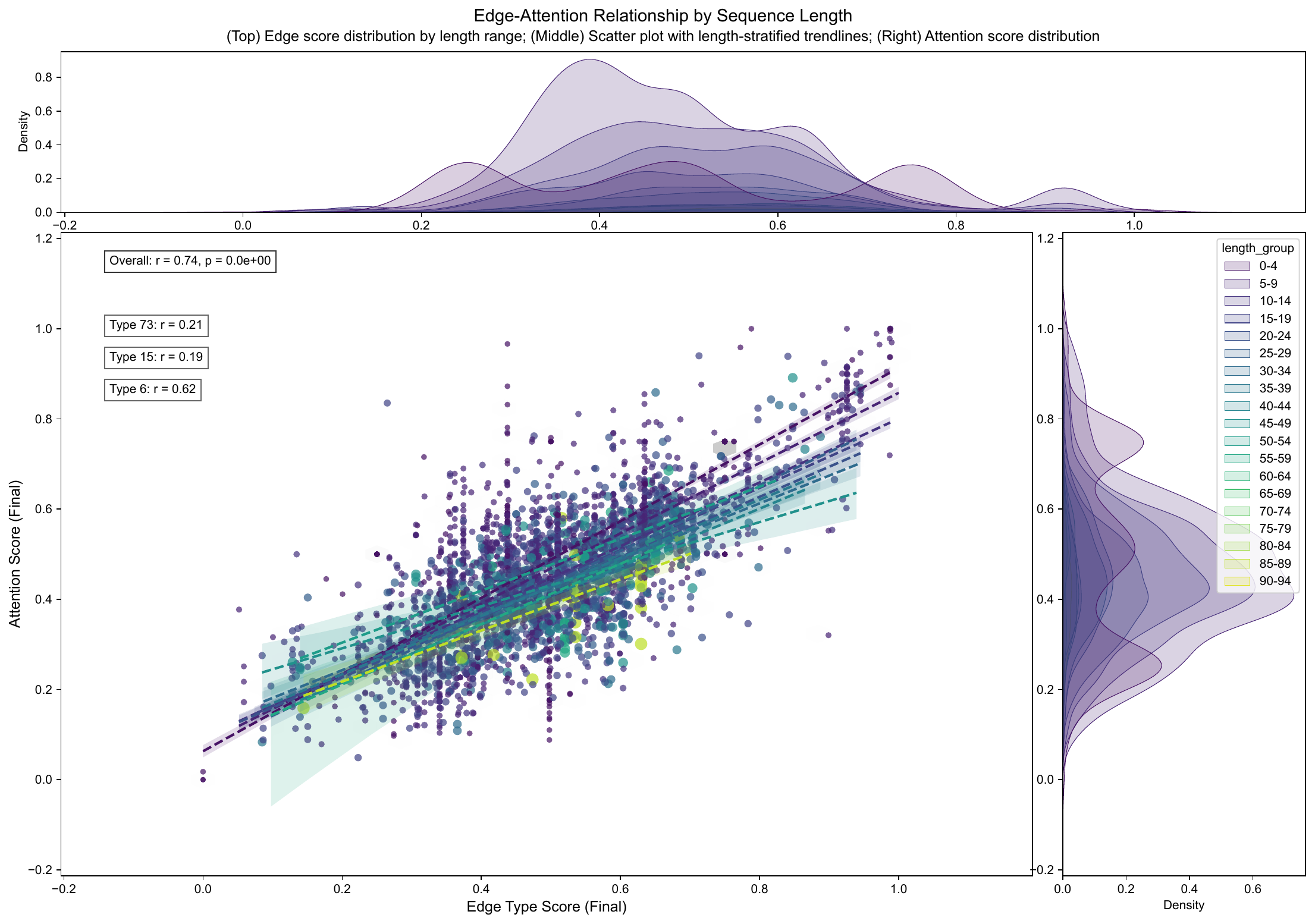}
\end{figure}

To extend the interpretability analysis, we examine how edge type embeddings influence the temporal dynamics of attention in the GAT-TDTE model. Figure~\ref{fig:corr13i} visualizes the relationship between final edge type scores (x axis) and learned attention weights (y axis) across sequences of varying lengths. Each point represents an edge, colored according to its sequence length category and scaled by total sequence length. Dashed lines represent per group regression trends, and marginal density plots summarize the distributions of edge type scores and attention weights.

The scatter and marginal plots reveal a strong overall correlation between edge type scores and attention weights ($r = 0.74$), indicating that the model meaningfully incorporates edge type semantics when allocating attention. This relationship is stronger in sequences of shorter length, where node level signals are limited, and weakens in longer sequences as node level dynamics become more dominant. Regression slope variation across length bins supports this adaptive behavior. Among the most frequent edge types, Type 6 shows a particularly strong correlation with attention weights ($r = 0.62$), whereas Types 15 and 73 exhibit weaker associations ($r = 0.19$ to $0.21$). These findings suggest that the model flexibly adjusts its reliance on structural edge features based on temporal context, effectively balancing structural priors and data driven attention learning.

These findings reinforce the value of integrating edge type semantics into attention computation and highlight the role of transition based edge embeddings in enhancing predictive accuracy for business process monitoring.

To further illustrate how edge type embeddings modulate attention dynamics at the sequence level, we include a representative case study in \ref{appendix:timeline}. The comparison between baseline and edge-augmented attention decompositions shows that while core attention focus remains consistent, edge-type scores refine importance weights—amplifying key transitions and sharpening predictive contrast.

Together, these results strengthen the interpretability argument: by linking attention weights to explicit transition categories, the analysis reveals which process relations consistently drive prediction outcomes. This makes the attention mechanism not only a performance enhancer but also an interpretable lens, exposing how temporal and semantic dependencies interact to shape model focus across cases.

\subsection{Comparative Discussion} \label{sec:CpDc}
We compare our model’s accuracy and Damerau–Levenshtein (DL) distance scores with prior approaches across six benchmark datasets, as reported in Table~\ref{tab:acc_comparison} and Table~\ref{tab:DL_comparison}. Our framework delivers competitive results, with particularly strong performance on the Helpdesk dataset. While TACO \citep{rama2023embedding} reports higher accuracy on structurally complex datasets such as BPI12 and BPI13, several methodological differences account for this gap and also highlight the strengths of our approach.

First, TACO employs hyperparameter tuning tailored to each dataset, whereas our framework uses a single global configuration across all datasets to enable fair comparison of architectural components such as edge embeddings and temporal decay. During training, we observed that our best performing GAT variants achieved training accuracies notably higher than their test scores (for example, by 5 to 8 percentage points on BPI13), suggesting that performance could be further improved through dataset specific optimization.

Second, TACO integrates an external process model, discovered using Split Miner, into its architecture. Through token replay on this model, it encodes control flow structures such as loops and parallel branches. This design provides a performance advantage on datasets with long or repetitive traces, such as BPI12W and BPI13i. In contrast, our model does not rely on any external process model and instead operates directly on event logs, learning edge semantics from data. This approach simplifies deployment and improves adaptability in scenarios where high quality process models are unavailable or difficult to construct. It is particularly well suited to real world environments where domain experts may be absent or where process behavior evolves too frequently for static models to remain effective.

Third, TACO combines graph convolutional networks and recurrent neural networks into a hybrid architecture that incrementally integrates temporal and structural context. In contrast, our model uses a graph attention mechanism to process the entire event trace in a single pass, which improves computational efficiency and supports interpretability. However, this design may be less effective in capturing long range dependencies in very deep traces, where sequential recurrence can offer advantages.

Despite these architectural differences, our framework delivers strong accuracy under a unified configuration and introduces several key advantages:(i) a time decay mechanism that adaptively reweights event relevance, (ii) edge embeddings informed by transition semantics, and (iii) built-in interpretability without relying on post hoc explanation methods. These design choices support a robust and generalizable PBPM solution with reduced dependence on tuning or external process models.

In addition to competitive accuracy, our models consistently achieve the highest Damerau–Levenshtein scores across all benchmark datasets (Table~\ref{tab:DL_comparison}), reflecting improved sequence alignment and predictive robustness in next event prediction tasks. Higher DL alignment implies that predicted sequences more closely reflect actual process behavior, which is essential for downstream applications such as process automation, decision support, and user-facing recommendations.

\begin{table}[htbp!]
\centering
\begin{threeparttable}
\caption{Accuracy Comparison with Previous Methods across Benchmark Datasets}
\label{tab:acc_comparison}
\begin{tabularx}{\textwidth}{l*{9}{>{\centering\arraybackslash}X}}
\toprule
\textbf{Dataset} & \textbf{Pas.} & \textbf{Tax} & \textbf{Cam.} & \textbf{Khan} & \textbf{Ev.} & \textbf{Mau.} & \textbf{Theis} & \textbf{TACO} & \textbf{Ours} \\
\midrule
Helpdesk & 0.6584 & 0.7506 & 0.7651 & 0.6913 & 0.7007 & 0.7477 & 0.6625 & 0.8520 & \textbf{0.9584} \\
BPI12    & 0.8259 & 0.8520 & 0.8341 & 0.8293 & 0.6038 & 0.8456 & 0.6423 & \textbf{0.8708} & 0.8561 \\
BPI12W   & 0.8159 & 0.8490 & 0.8329 & 0.8669 & 0.7522 & 0.8511 & 0.7616 & \textbf{0.8834} & 0.8763 \\
BPI13c   & 0.2435 & 0.6557 & 0.6062 & 0.5557 & 0.5566 & 0.5697 & 0.4769 & \textbf{0.6753} & 0.6360 \\
BPI13i   & 0.3110 & 0.6750 & 0.6801 & 0.6434 & 0.6815 & 0.7109 & 0.6351 & \textbf{0.7166} & 0.6995 \\
BPI20   & - & - & - & - & - & - & - & - & - \\
\bottomrule
\end{tabularx}

\caption{Damerau--Levenshtein Score across Benchmark Datasets}
\label{tab:DL_comparison}
\begin{tabularx}{\textwidth}{l *{6}{>{\centering\arraybackslash}X}}
\toprule
\textbf{Dataset} & \textbf{Ev.} & \textbf{Tax} & \textbf{Cam.} & \textbf{Fran.} & \textbf{Ours} \\
\midrule
Helpdesk & 0.8069 & 0.8121 & 0.917 & 0.1991 & \textbf{0.9703} \\
BPI12    & 0.1979 & 0.1213 & 0.632  & 0.1189 & \textbf{0.8179} \\
BPI12w   & 0.2769 & 0.0956 & 0.525  & 0.1013 & \textbf{0.8590} \\
BPI13c   & 0.6712 & 0.4856 & - & 0.4341 & \textbf{0.6657} \\
BPI13i   & 0.4786 & 0.2327 & - & 0.2743 & \textbf{0.6986} \\
BPI20   & - & - & - & - & - \\
\bottomrule
\end{tabularx}

\begin{tablenotes}
\small
\item \textbf{Pas.} = \cite{pasquadibisceglie2019using}, \textbf{Tax} = \cite{tax2017predictive}, \textbf{Cam.} = \cite{camargo2019learning},\textbf{Khan} = \cite{khan2018memory}, \textbf{Ev.} = \cite{evermann2017predicting}, \textbf{Mau.} = \cite{di2019activity}, \textbf{Theis} = \cite{theis2019decay}, \textbf{TACO} = \cite{rama2023embedding}, \textbf{Fran.} =\cite{di2017eye}
\end{tablenotes}
\end{threeparttable}
\end{table}

\section{Conclusion and Discussion} \label{sec:conclusion}
Modern graph neural network models for business process monitoring often focus on short local prefixes or apply global attention without incorporating temporal dynamics or semantic transitions. These limitations reduce their ability to adaptively identify prediction relevant events in complex, time varying traces. This work addresses these challenges through a unified framework that bridges local subgraph modeling and full sequence level reasoning. It introduces a time decay mechanism to construct dynamic temporal attention windows and incorporates semantically informed edge embeddings to capture meaningful event transitions.

Our models retain the full event history and jointly predict the next event and its substatus, enabling a more expressive and process aware output space compared to previous approaches. Empirical evaluations across six public datasets show that global attention through GAT improves Top-1 accuracy by an average of 5.5 percent over prefix based GCNs. The addition of time decay attention provides a further gain of 1.5 percent in sequences of medium length with high temporal variability. Edge embeddings informed by transition semantics enhance the model’s ability to distinguish structurally similar but semantically distinct traces. High Top-3 and Top-5 accuracies (exceeding 86\%), along with superior Damerau–Levenshtein alignment, confirm both event level prediction accuracy and sequence level coherence. Crucially, we also delineate the boundary of global attention: in a regime with high class cardinality, sparse event-level features, and short-to-moderate traces (BPI20), long-prefix GCNs outperform GATs. The unified framework and its interpretability decomposition make the regime explicit and provide prescriptive guidance: use GATs for longer, feature-rich traces where selective weighting pays off; use long-prefix GCNs when per-event signal is weak and the class space is large. The contribution is therefore not only improved averages, but a clear, data-driven rule for model selection across process regimes.

To support transparency and actionable decision making in PBPM, we introduce a suite of interpretability visualizations that explain how dynamic temporal attention windows and transition semantics influence predictions across sequences of varying length. Quantitative analysis shows a consistent contraction of attention span from 34 to 20 percent as sequence length increases, indicating that the model effectively suppresses irrelevant context while maintaining predictive accuracy. Edge attention correlation analysis ($r = 0.74$) confirms that transition informed edge embeddings meaningfully influence attention, particularly in sequences with structural ambiguity. These interpretability tools support practical use cases such as bottleneck detection and compliance auditing, while visual decompositions presented in ~\ref{appendix:timeline} offer domain experts a means to investigate prediction behavior on individual cases. Taken together, these contributions advance the state of predictive process monitoring by combining accuracy, interpretability, and practical deployability in a unified graph-based framework.

While the proposed framework advances PBPM through a unified design that integrates prefix-based GCNs and full-trace GATs with temporal decay and transition semantics, several limitations remain and point to future work. First, we deliberately fixed most hyperparameters globally to ensure comparability across models and datasets, but more extensive hyperparameter optimization could further improve performance. Exploring such optimization in conjunction with automated search techniques (e.g., AutoML) would help clarify which architectural components benefit most from dataset-specific tuning. Second, in this study, categorical features were encoded using one-hot representations across all models to ensure a fair and controlled comparison. This choice limits expressiveness but avoids introducing confounding variation in our architectural benchmarks. In future work, alternative encoding strategies (e.g., embeddings or frequency-based encodings) could be explored to examine which model architectures benefit most from richer feature representations. Third, while our evaluation relies on bucket-based prefix sampling to balance short, medium, and long prefixes, an alternative is to enumerate all prefixes and focus on last-event prediction. We consider this complementary setup an interesting extension for future work, particularly for scenarios where complete trace coverage is prioritized over prefix diversity. Four, one limitation of our approach is the assumption of complete and consistent transition-type labeling. While in benchmark event logs transition categories are derived deterministically from observed event pairs, in real-world applications incomplete or noisy labeling may reduce the reliability of transition embeddings. Extending the framework to handle uncertain or probabilistic transition types is an important direction for future work. Fifth, we evaluated on six benchmark datasets spanning multiple domains, but future studies could expand coverage to additional real-world logs to further stress-test generalizability. Moreover, our current models operate at the sequence level; extending them to hierarchical or multi-instance settings (e.g., nested or interacting processes) could broaden applicability. In addition, combining our semantic edge embeddings with TACO’s state accumulation strategy, such as incorporating token counts as node features, may offer complementary strengths. Integrating these advancements into practitioner oriented tools could further bridge predictive analytics with operational process optimization.

\section{Declaration of generative AI and AI-assisted technologies in the writing process}

Statement: During the preparation of this work the authors used ChatGPT in order to improve the readability and language of the manuscript in the writing process. After using this tool/service, the authors reviewed and edited the content as needed and takes full responsibility for the content of the published article.

\bibliographystyle{elsarticle-harv}
\bibliography{ref}

\appendix
\renewcommand{\thesection}{Appendix~\Alph{section}}
\renewcommand{\thefigure}{\thesection.\arabic{figure}}
\setcounter{figure}{0} 

\section{Temporal Attention Decomposition Case Study}
\label{appendix:timeline}
\begin{figure}[htbp!]
\centering
\begin{subfigure}{\textwidth}
    \includegraphics[width=\textwidth,  trim={0 0 0 0.75cm}, clip]{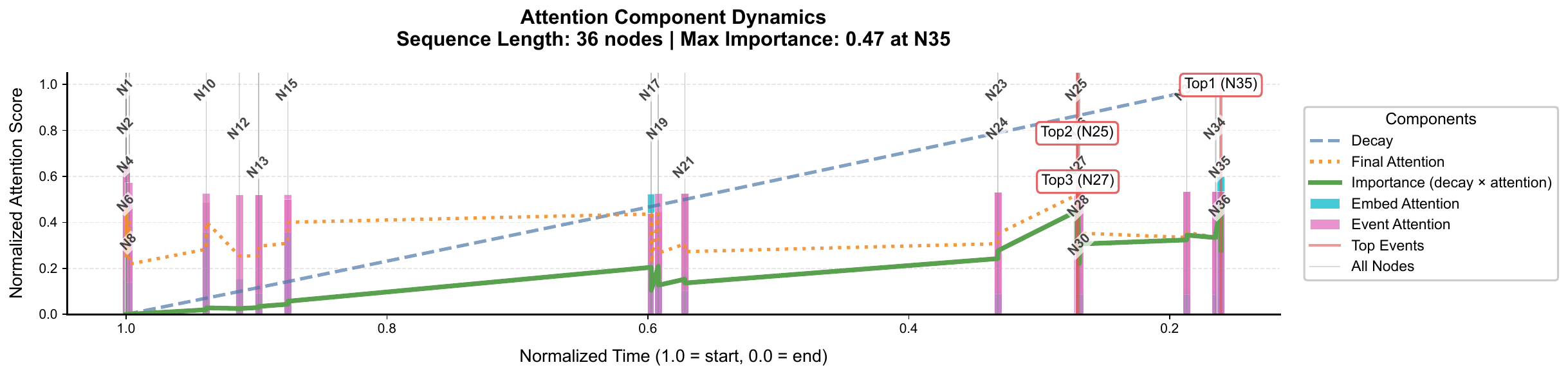}
    \caption{Dynamic Temporal Windows with Attention Decomposition}
     \label{subfig:sample13i}
\end{subfigure}

\begin{subfigure}{\textwidth}
    \includegraphics[width=\textwidth,  trim={0 0 0 0.75cm}, clip]{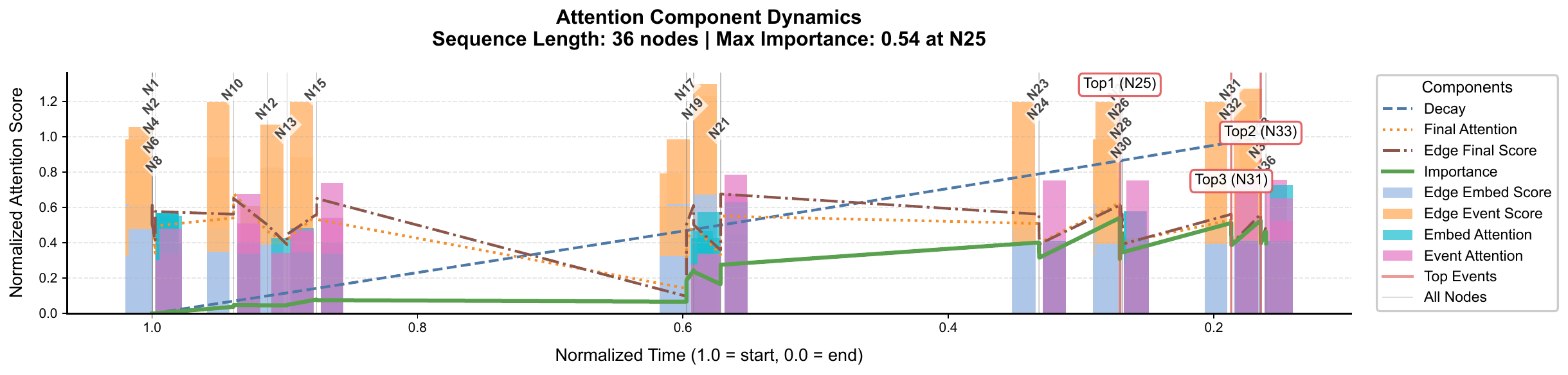}
    \caption{Dynamic Temporal Windows with Attention Decomposition Interacted with Edge Type Embedding}
    \label{subfig:transample13i}
\end{subfigure}

\caption{Temporal Attention Decomposition for a Representative Sequence (36 events) from BPI13i Using GAT-TD and GAT-TDTE Model}
\label{fig:edgeoverall}
\end{figure}

To better understand how attention is formed and modified in the GAT-TDTE model, we analyze a representative sequence of medium to long length (36 events) from the BPI13i dataset using two side by side visualizations. Figure~\ref{subfig:sample13i} shows the decomposition of attention components without edge type embeddings, while Figure~\ref{subfig:transample13i} introduces edge type scores into the attention mechanism. These visualizations illustrate how edge level semantics influence the distribution of attention across time and reshape the model’s predictive focus.

Figure~\ref{subfig:sample13i} breaks down the attention composition across the normalized timeline of the sequence. The model integrates two sources of node attention: embedding based attention (blue bars, derived from node identifiers) and event feature attention (pink bars). These components jointly contribute to the final attention weights (dotted orange), which are further modulated by a time decay function (dashed blue) to produce the importance score (green line), representing the effective contribution of each node to the prediction. Key observations from this configuration include:(i) A dominant focus on recent events, particularly in the final quarter of the sequence (e.g., node 35 with a peak importance score of 0.47);(ii) A gradual increase in attention intensity as the time decay curve amplifies proximity based weighting, reinforcing temporal locality;(ii) The top three predictive nodes are tightly clustered near the sequence end, reflecting a compressed dynamic attention window centered on the prediction point.

Figure~\ref{subfig:transample13i} extends the previous view by incorporating edge type embedding scores, visualized through three additional components: Edge Embed Score, Edge Event Score, and Edge Final Score. These edge informed values adjust the baseline attention by introducing semantic information derived from transitions between events, rather than solely from node attributes. The inclusion of edge semantics leads to three notable shifts:(i) Amplification and suppression: Certain nodes (e.g., nodes 25 and 33) increase in importance, while others decrease, despite having similar decay positions. This indicates that edge semantics refine relevance beyond what temporal proximity alone captures. (ii) Sharper attention profile: The overall importance curve (green) becomes steeper and exhibits more distinct peaks, suggesting that edge features function as gating mechanisms that sharpen the model’s predictive focus. (iii) Alignment between edge and node attention: The Edge Final Score closely follows the standard final attention, demonstrating that the model effectively integrates both node and edge modalities when determining attention weights for prediction.

This case study demonstrates that incorporating edge type embeddings enhances both the interpretability and precision of attention dynamics. While the core attention focus remains consistent across both visualizations— with the same top three nodes identified— their relative importance is reweighted to reflect transition level semantics. This enables the model to distinguish between structurally similar but semantically different event transitions, which is particularly important in process data where sequences may exhibit temporal and structural ambiguity.

\setcounter{figure}{0}
\section{Rank Based Node Dominance Analysis on BPI13i}
\label{appendix:rank}
\begin{figure}[htbp!]
    \centering
     \caption{Top Node Dominance by Rank and length Range}
  \label{fig:rank13i}
  \includegraphics[width=\textwidth, trim={0 0 0 1cm}, clip]{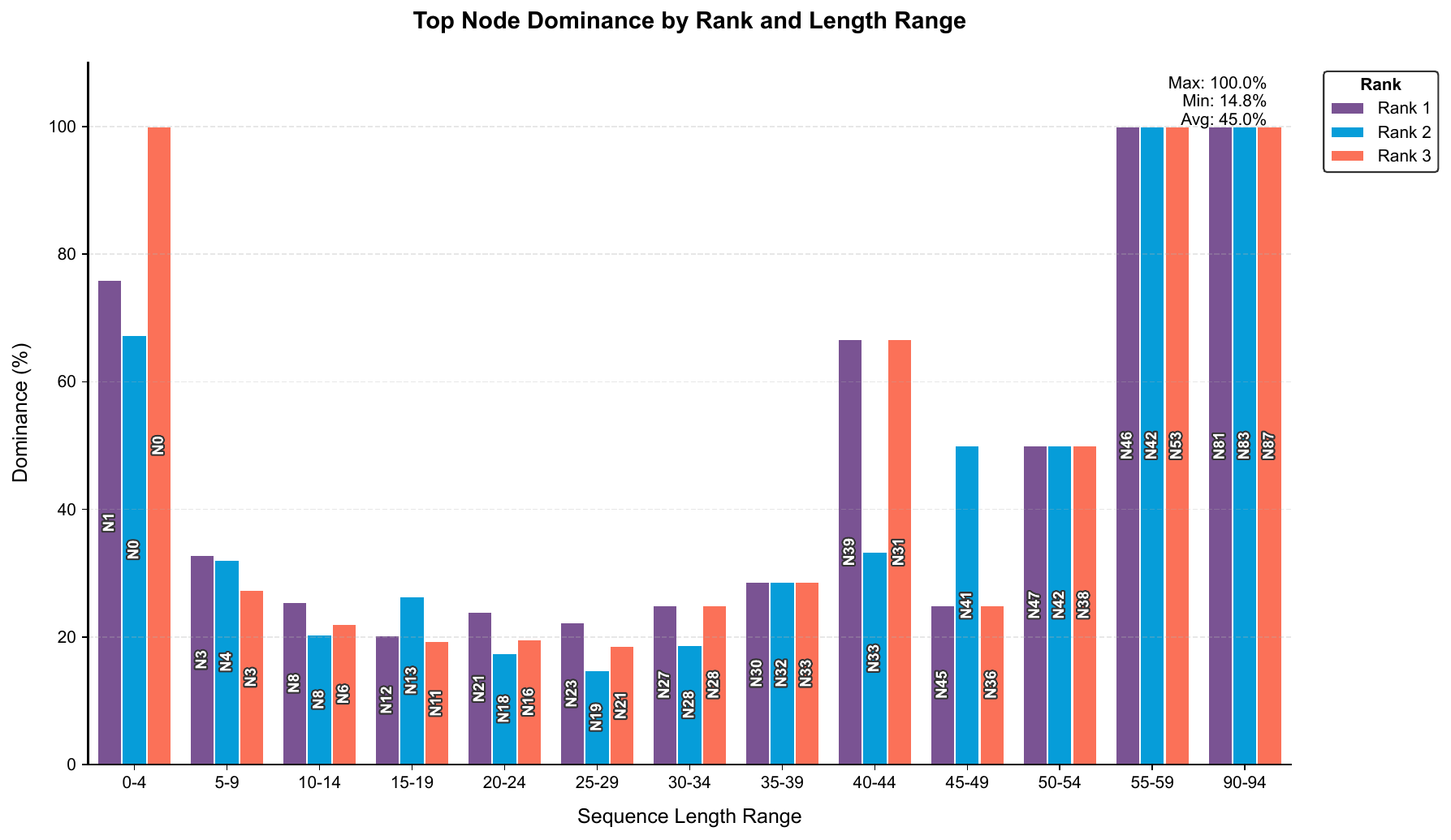}
\end{figure}

Rank based node dominance analysis (see Figure ~\ref{fig:rank13i}) examines positional consistency by ranking nodes according to their dominance, defined as frequency of appearance in the Top-3 positions across different sequence lengths. In longer sequences, specific late node positions exhibit strong dominance; for example, node 81 appears in the Top-3 with 100 percent frequency in sequences of length 90 to 94. In contrast, medium length sequences distribute attention more broadly among recent nodes. For very short sequences (0 to 4 events), the attention focus is concentrated on immediate recency, with nodes 0 and 1 achieving over 75.98 percent Top-3 frequency and importance scores exceeding 0.278.

This supplementary analysis supports the patterns observed in Figure~\ref{subfig:critwin13i}, confirming that the time decay attention mechanism generates adaptive prediction windows that align with sequence characteristics and consistently identify the most relevant temporal positions for prediction.

\setcounter{figure}{0}
\section{Cross-Dataset Interpretability Visualizations}
\label{appendix:GATInter}

\begin{figure}[htbp!]
\centering

\begin{subfigure}{0.48\textwidth}
    \includegraphics[width=\textwidth, trim={0 0 0 1cm}, clip]{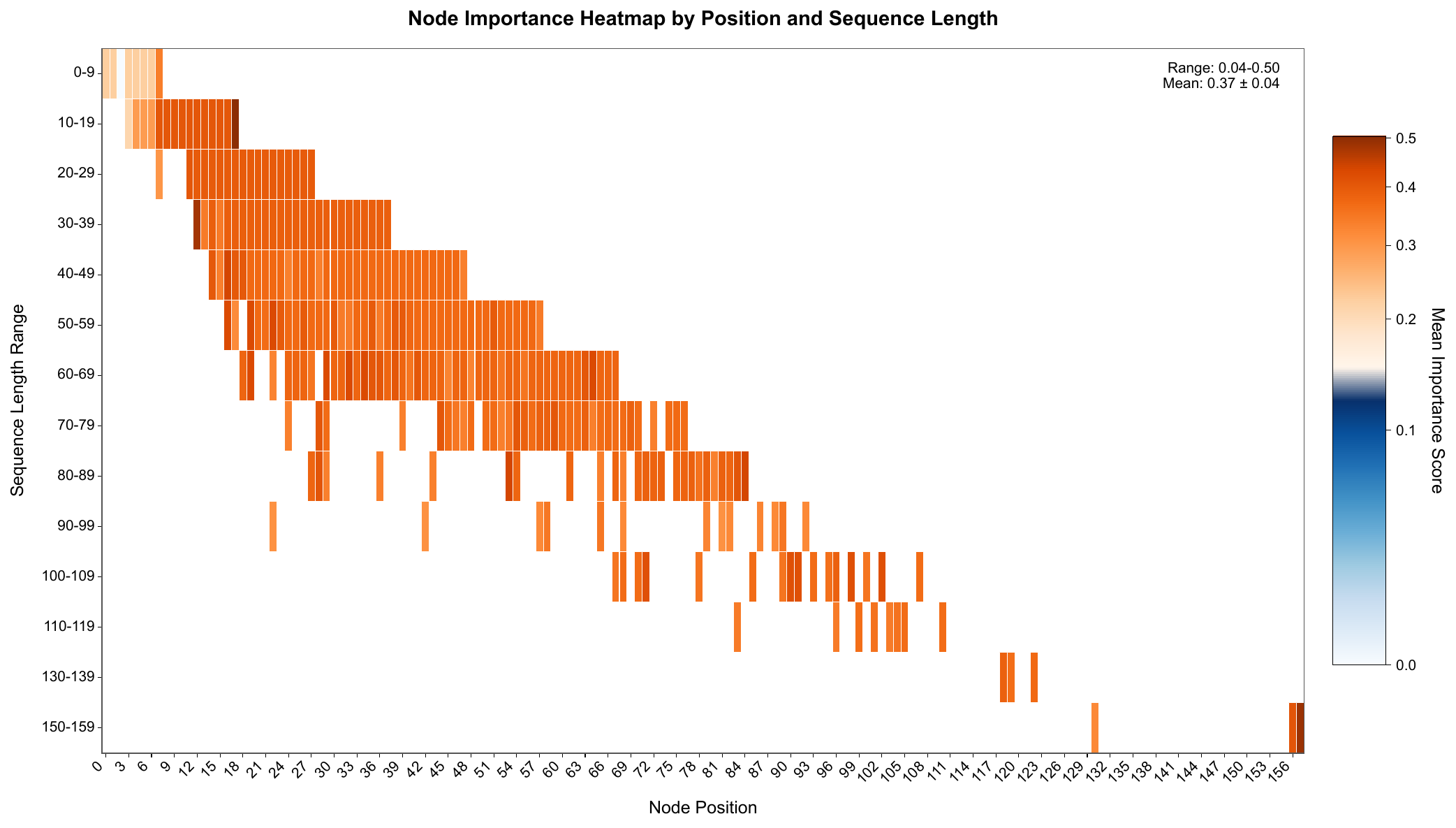}
    \caption{Dataset: BPI12; Model: GAT-TD}
    \label{app_subfig:hm12}
\end{subfigure}
\hfill
\begin{subfigure}{0.48\textwidth}
    \includegraphics[width=\textwidth, trim={0 0 0 1cm}, clip]{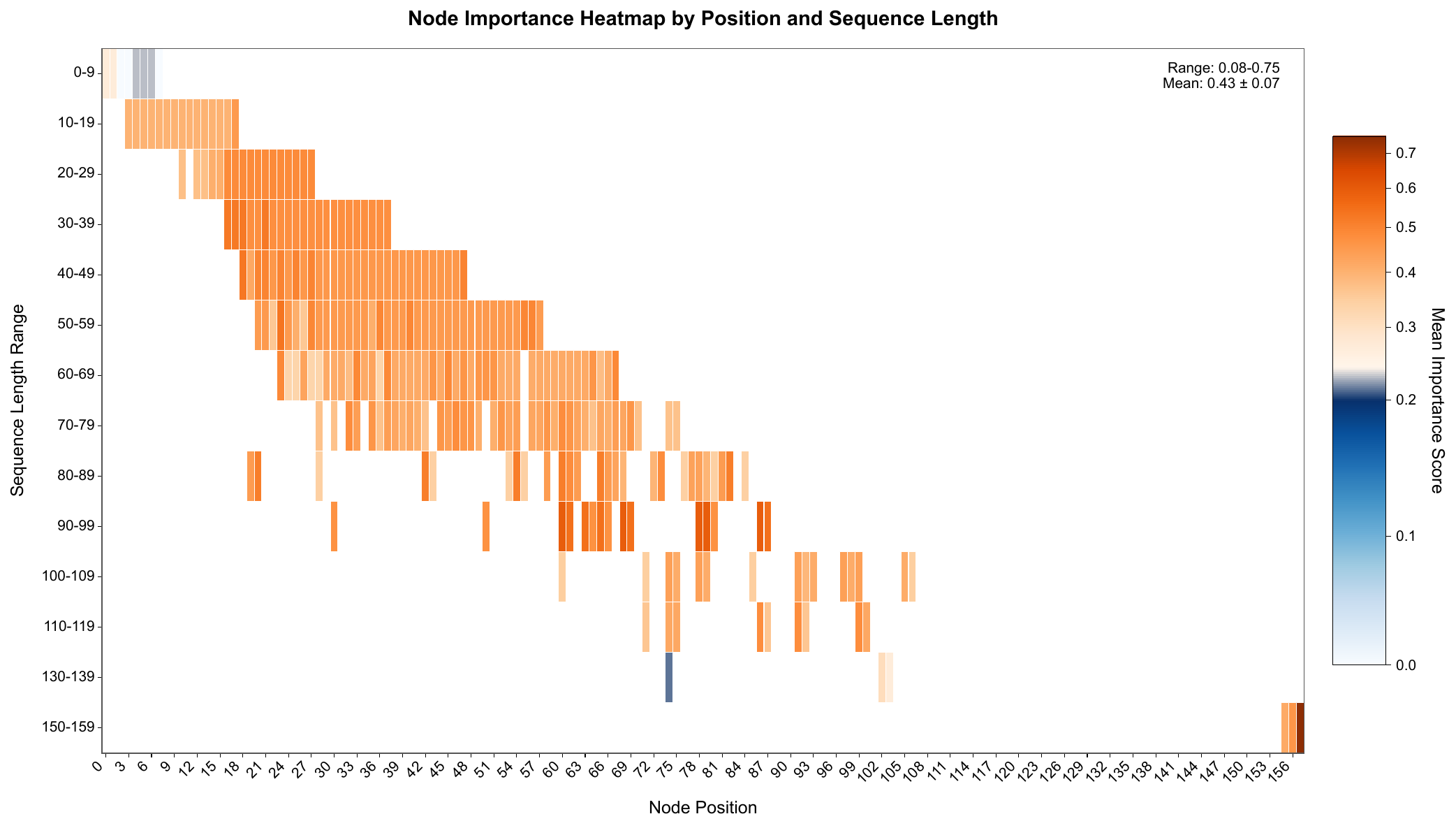}
    \caption{Dataset: BPI12; Model GAT-TDTE;}
    \label{app_subfig:hm12t}
\end{subfigure}

\vspace{0.5em}
\begin{subfigure}{0.48\textwidth}
    \includegraphics[width=\textwidth, trim={0 0 0 1cm}, clip]{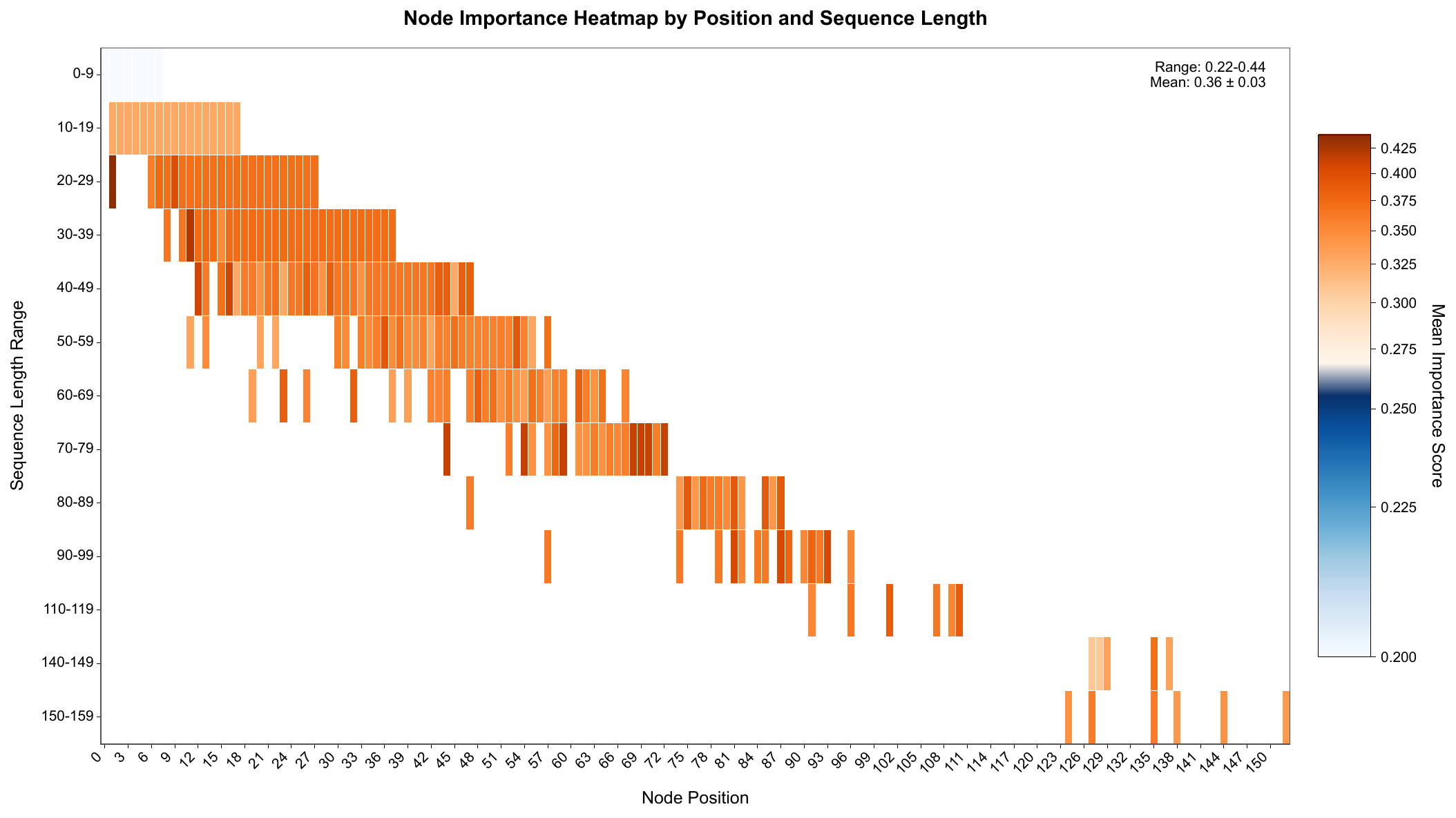}
    \caption{Dataset: BPI12W; Model: GAT-TD}
    \label{app_subfig:hm12w}
\end{subfigure}
\hfill
\begin{subfigure}{0.48\textwidth}
    \includegraphics[width=\textwidth, trim={0 0 0 1cm}, clip]{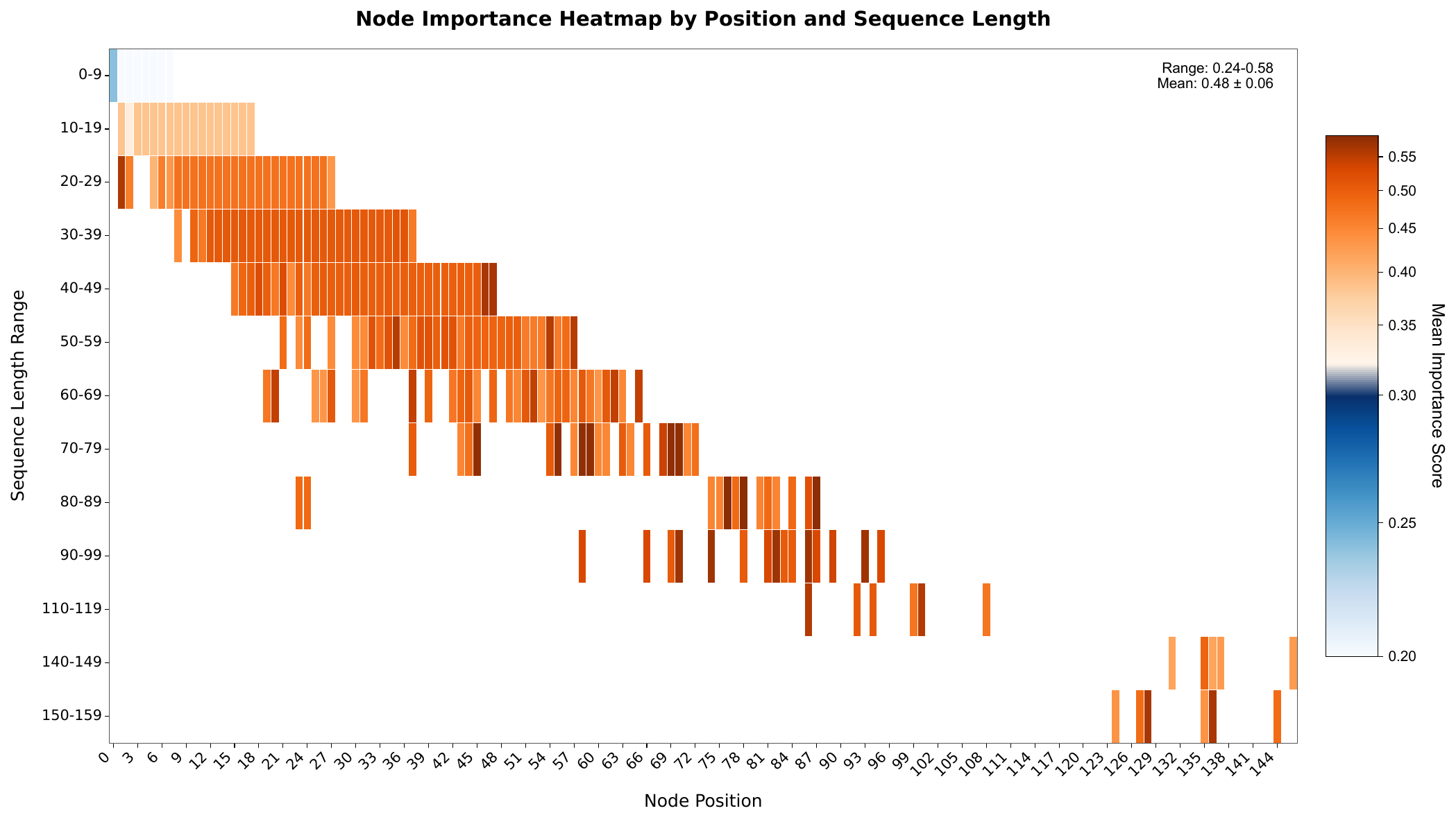}
    \caption{Dataset: BPI12W; Model GAT-TDTE}
    \label{app_subfig:hm12wt}
\end{subfigure}

\vspace{0.5em}
\begin{subfigure}{0.48\textwidth}
    \includegraphics[width=\textwidth, trim={0 0 0 1cm}, clip]{BPI13i_timedgedecay_heatmap.pdf}
    \caption{Dataset: BPI13i; Model: GAT-TD}
    \label{app_subfig:hm13i}
\end{subfigure}
\hfill
\begin{subfigure}{0.48\textwidth}
    \includegraphics[width=\textwidth, trim={0 0 0 1cm}, clip]{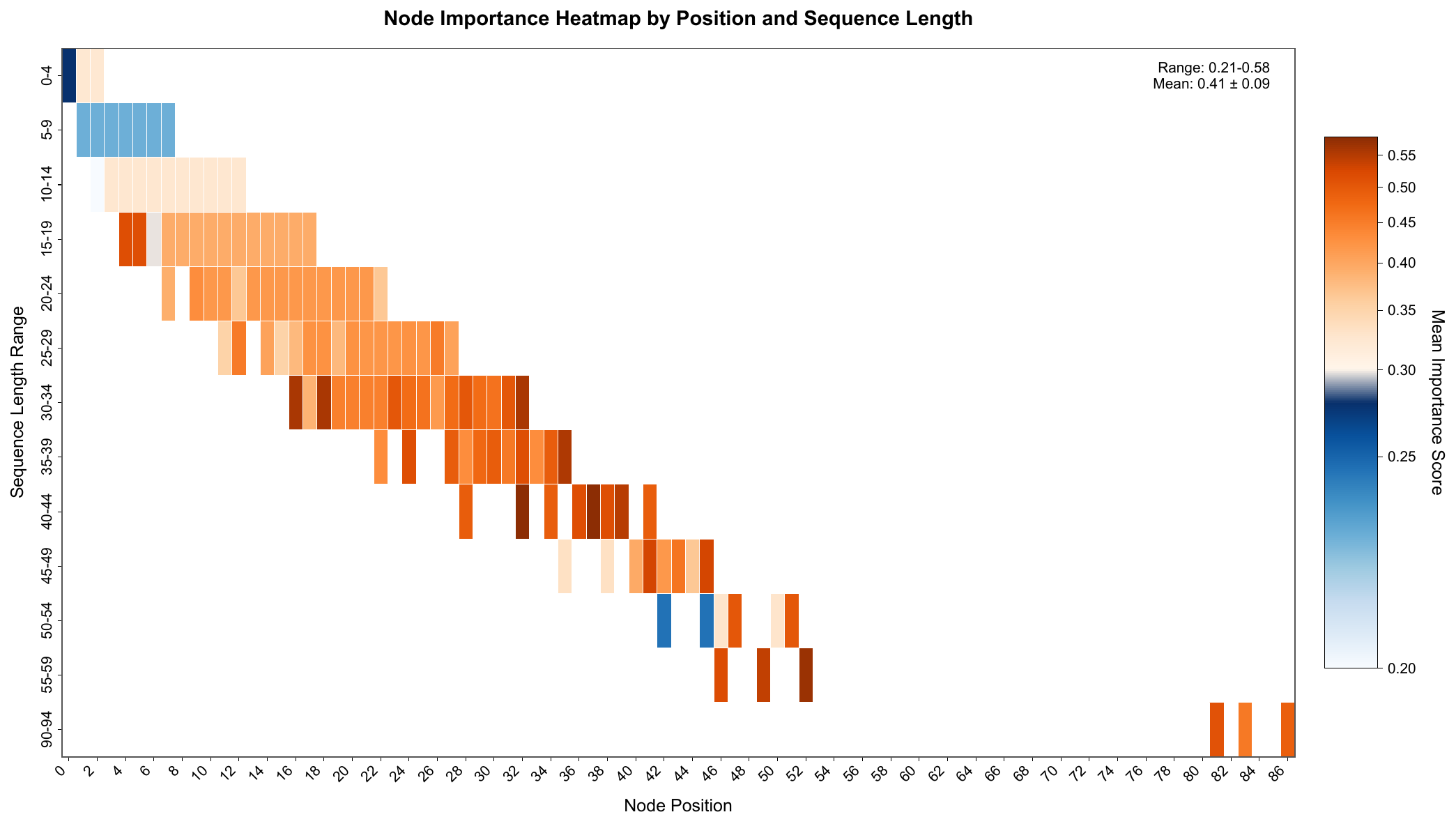}
    \caption{Dataset: BPI13i; Model GAT-TDTE}
    \label{app_subfig:hm13it}
\end{subfigure}

\vspace{0.5em}
\begin{subfigure}{0.48\textwidth}
    \includegraphics[width=\textwidth, trim={0 0 0 1cm}, clip]{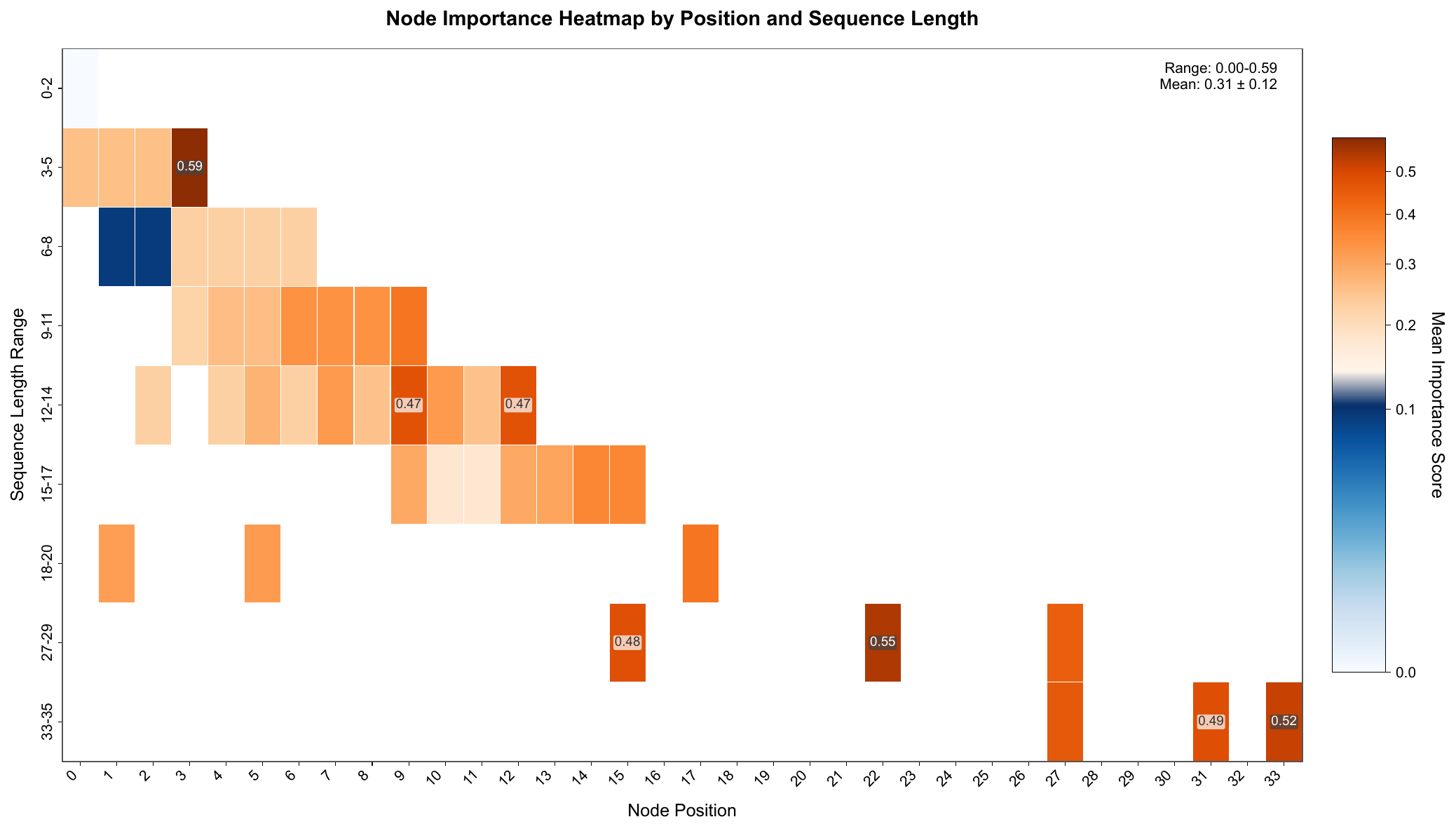}
    \caption{Dataset: BPI13c; Model: GAT-TD}
    \label{app_subfig:hm13c}
\end{subfigure}
\hfill
\begin{subfigure}{0.48\textwidth}
    \includegraphics[width=\textwidth, trim={0 0 0 1cm}, clip]{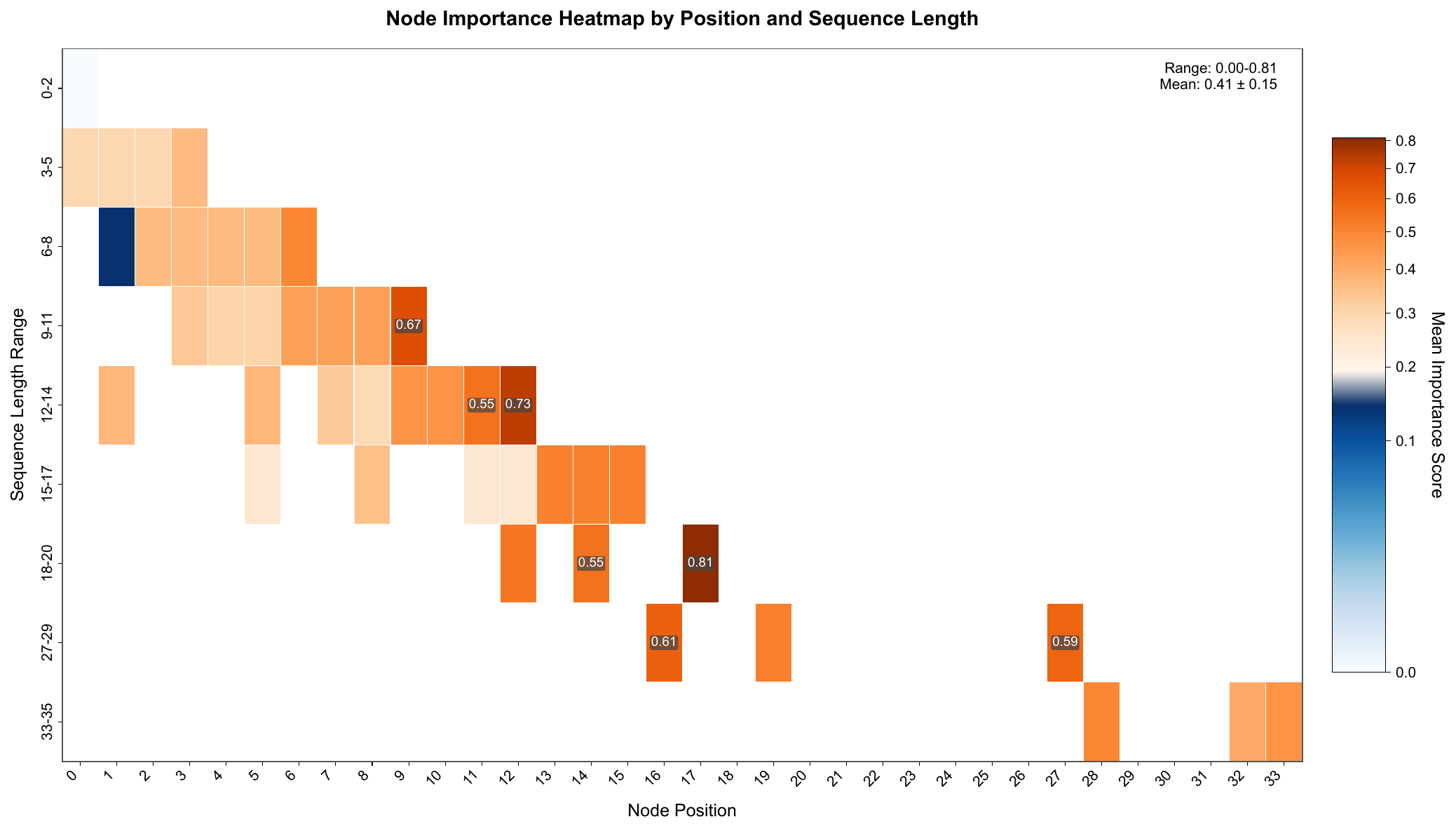}
    \caption{Dataset: BPI13c; Model GAT-TDTE}
    \label{app_subfig:hm13ct}
\end{subfigure}
\caption{Node Importance Heatmap by Sequence Length and Position (part 1)}
\label{fig:hm5x2grid_p1}
\end{figure}

\begin{figure}[htbp!]
\centering
\begin{subfigure}{0.48\textwidth}
    \includegraphics[width=\textwidth, trim={0 0 0 1cm}, clip]{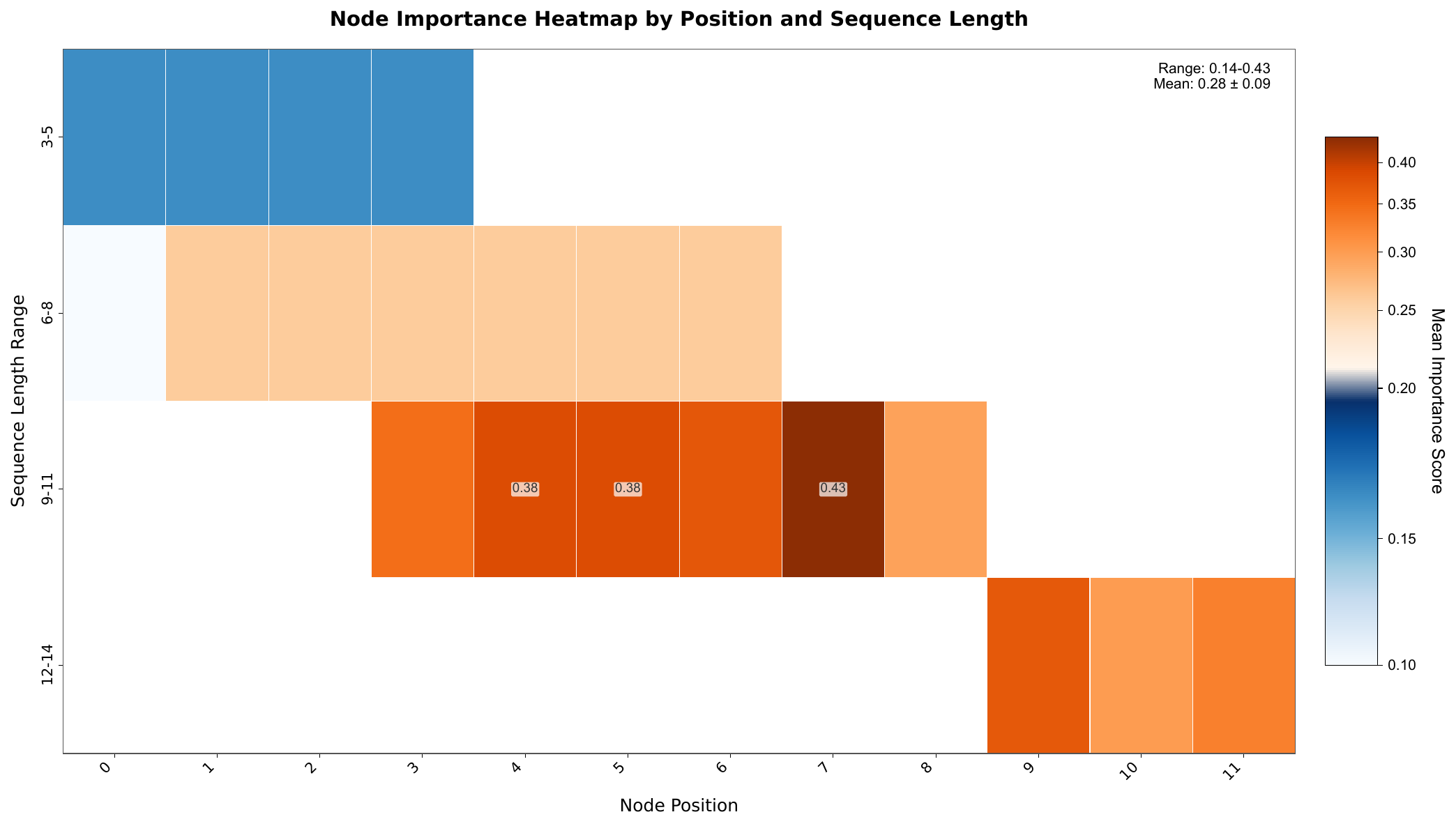}
    \caption{Dataset: Helpdeask; Model: GAT-TD}
    \label{app_subfig:hm0}
\end{subfigure}
\hfill
\begin{subfigure}{0.48\textwidth}
    \includegraphics[width=\textwidth, trim={0 0 0 1cm}, clip]{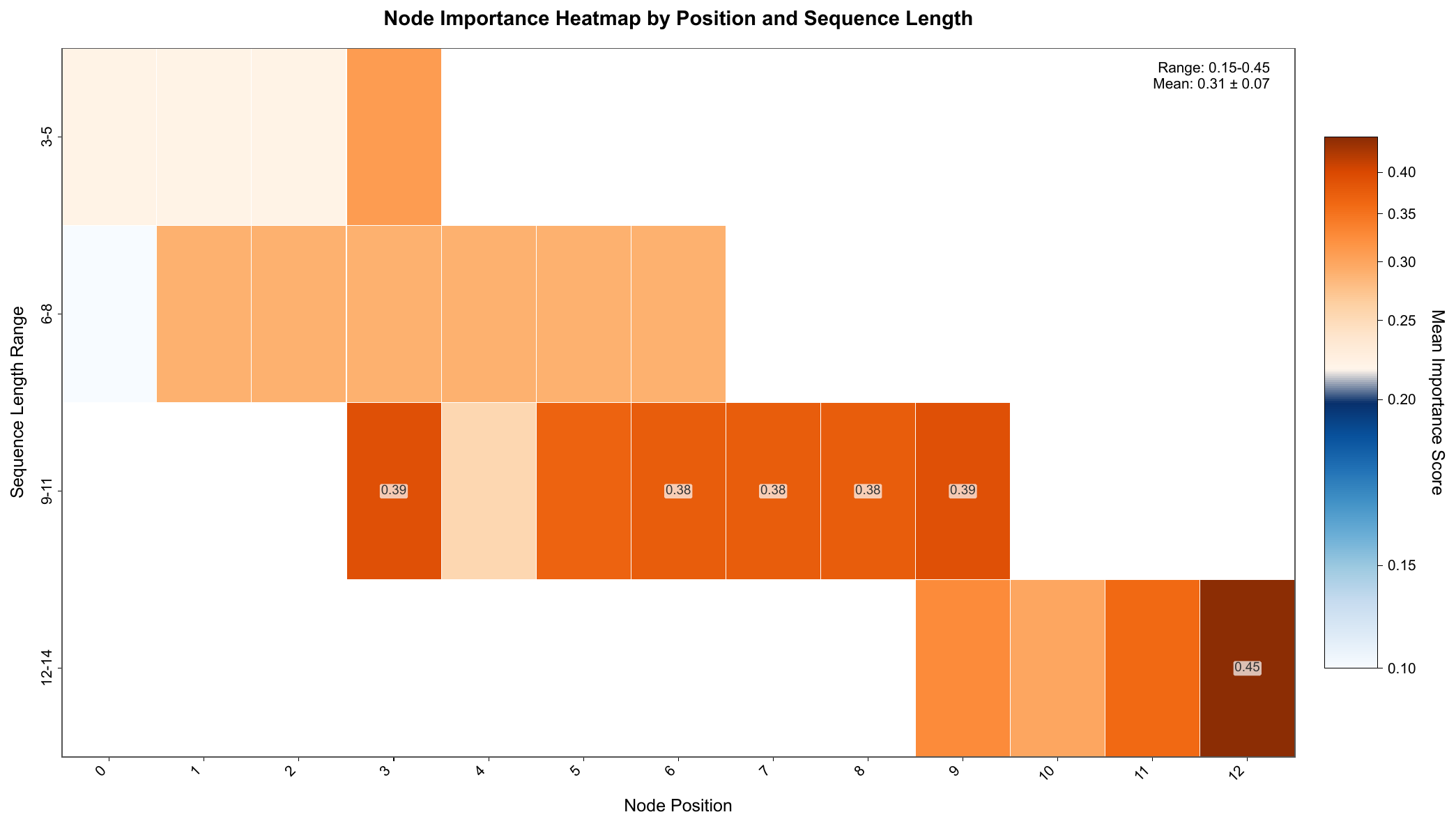}
    \caption{Dataset: Helpdesk; Model GAT-TDTE}
    \label{app_subfig:hm0t}
\end{subfigure}
\vspace{0.5em}
\begin{subfigure}{0.48\textwidth}
    \includegraphics[width=\textwidth, trim={0 0 0 1cm}, clip]{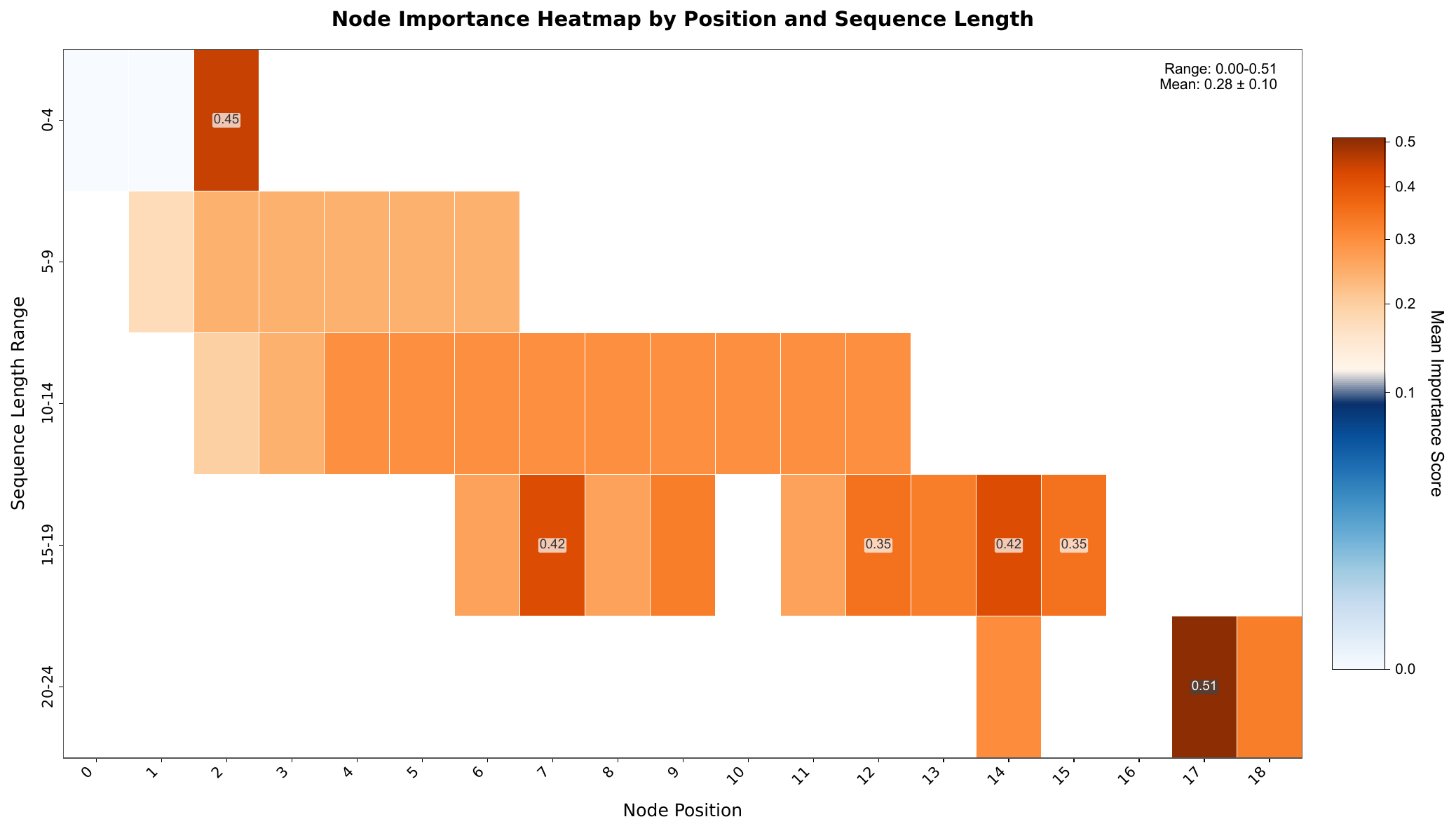}
    \caption{Dataset: BPI20; Model: GAT-TD}
    \label{app_subfig:hm20}
\end{subfigure}
\hfill
\begin{subfigure}{0.48\textwidth}
    \includegraphics[width=\textwidth, trim={0 0 0 1cm}, clip]{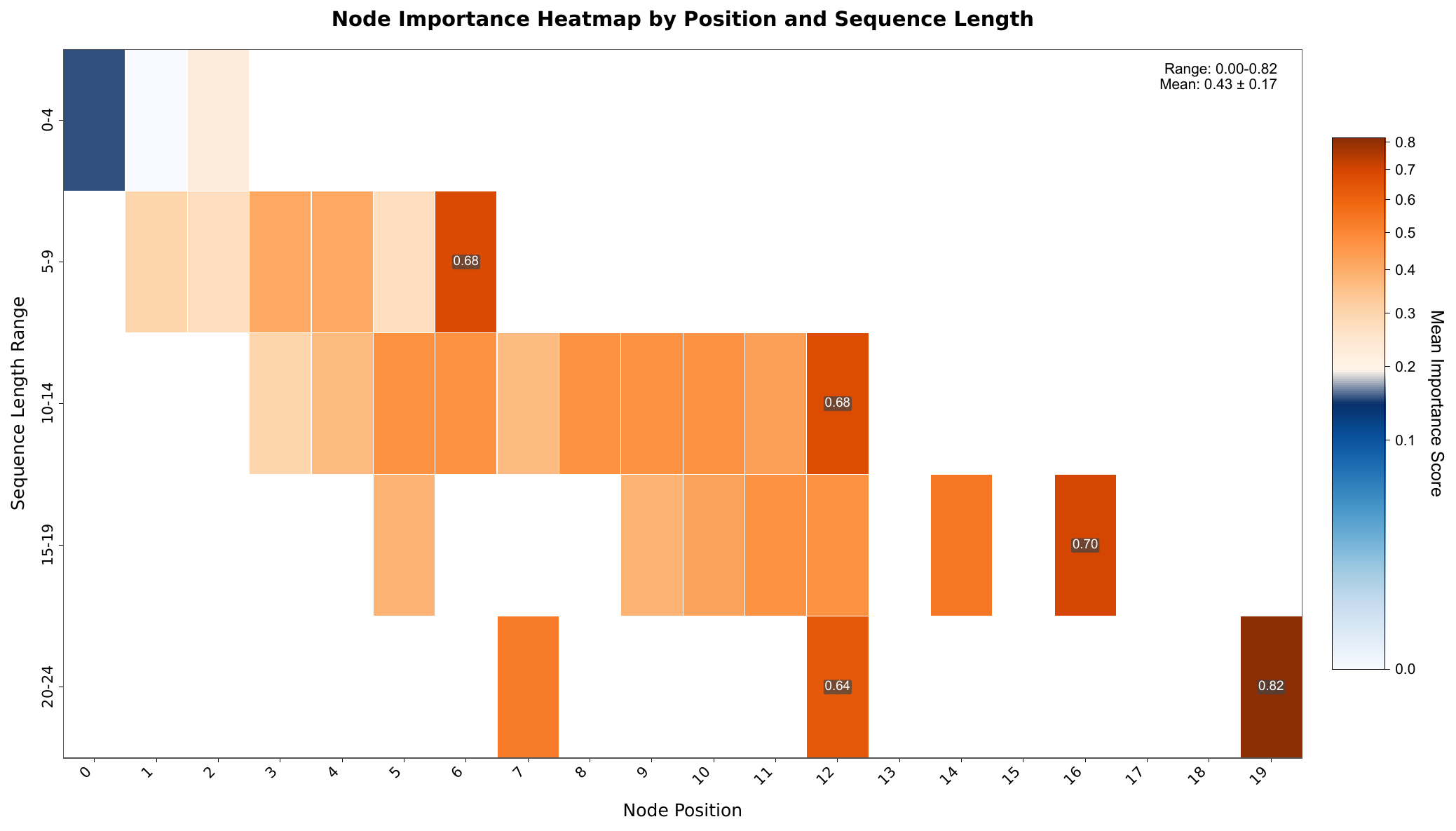}
    \caption{Dataset: BPI20; Model GAT-TDTE}
    \label{app_subfig:hm20t}
\end{subfigure}

\caption{Node Importance Heatmap by Sequence Length and Position (part 2)}
\label{fig:hm5x2grid_p2}
\end{figure}

\begin{figure}[htbp!]
\centering

\begin{subfigure}{0.48\textwidth}
    \includegraphics[width=\textwidth, trim={0 0 0 1cm}, clip]{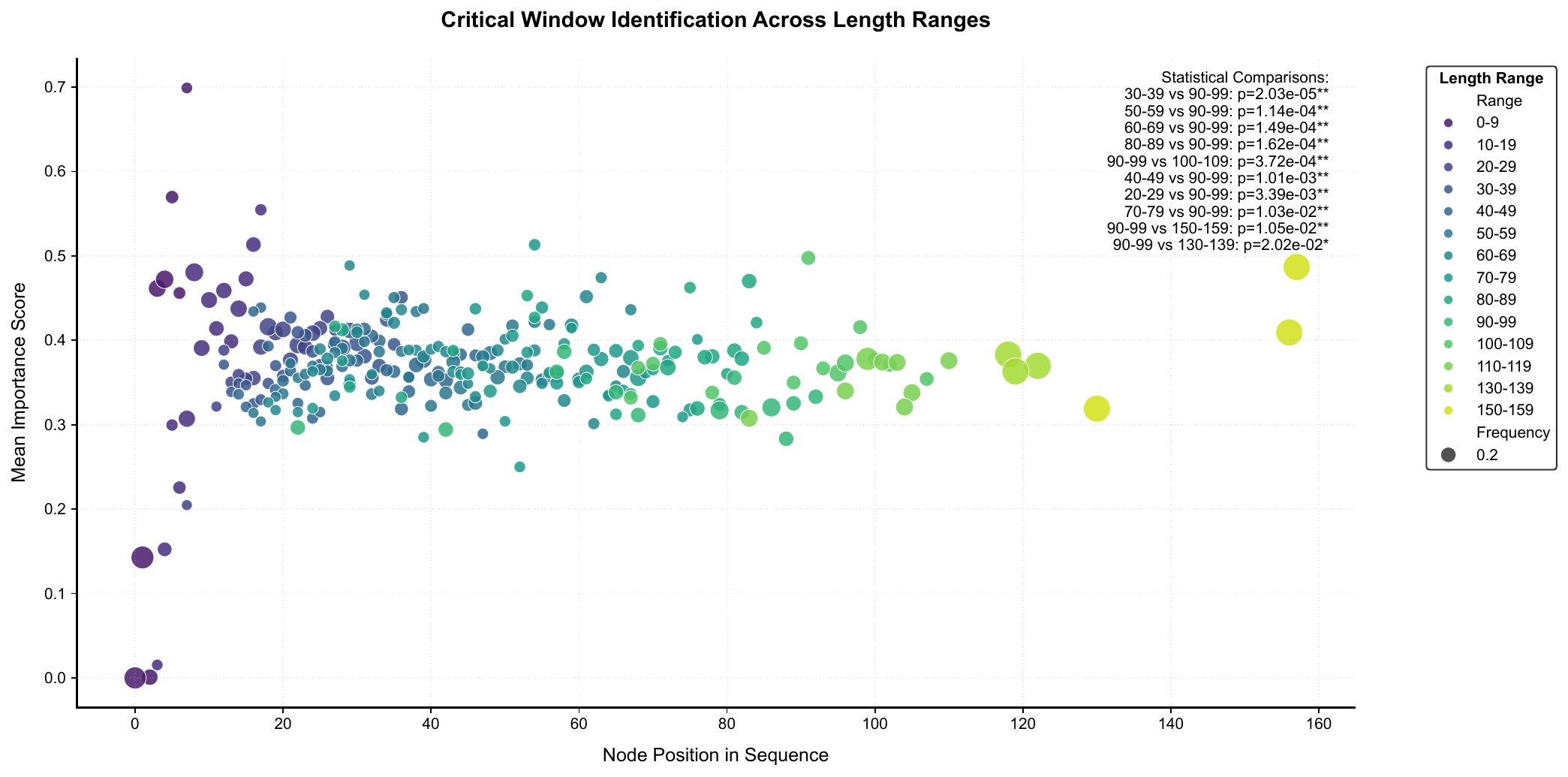}
    \caption{Dataset: BPI12; Model: GAT-TD}
    \label{app_subfig:cw12}
\end{subfigure}
\hfill
\begin{subfigure}{0.48\textwidth}
    \includegraphics[width=\textwidth, trim={0 0 0 1cm}, clip]{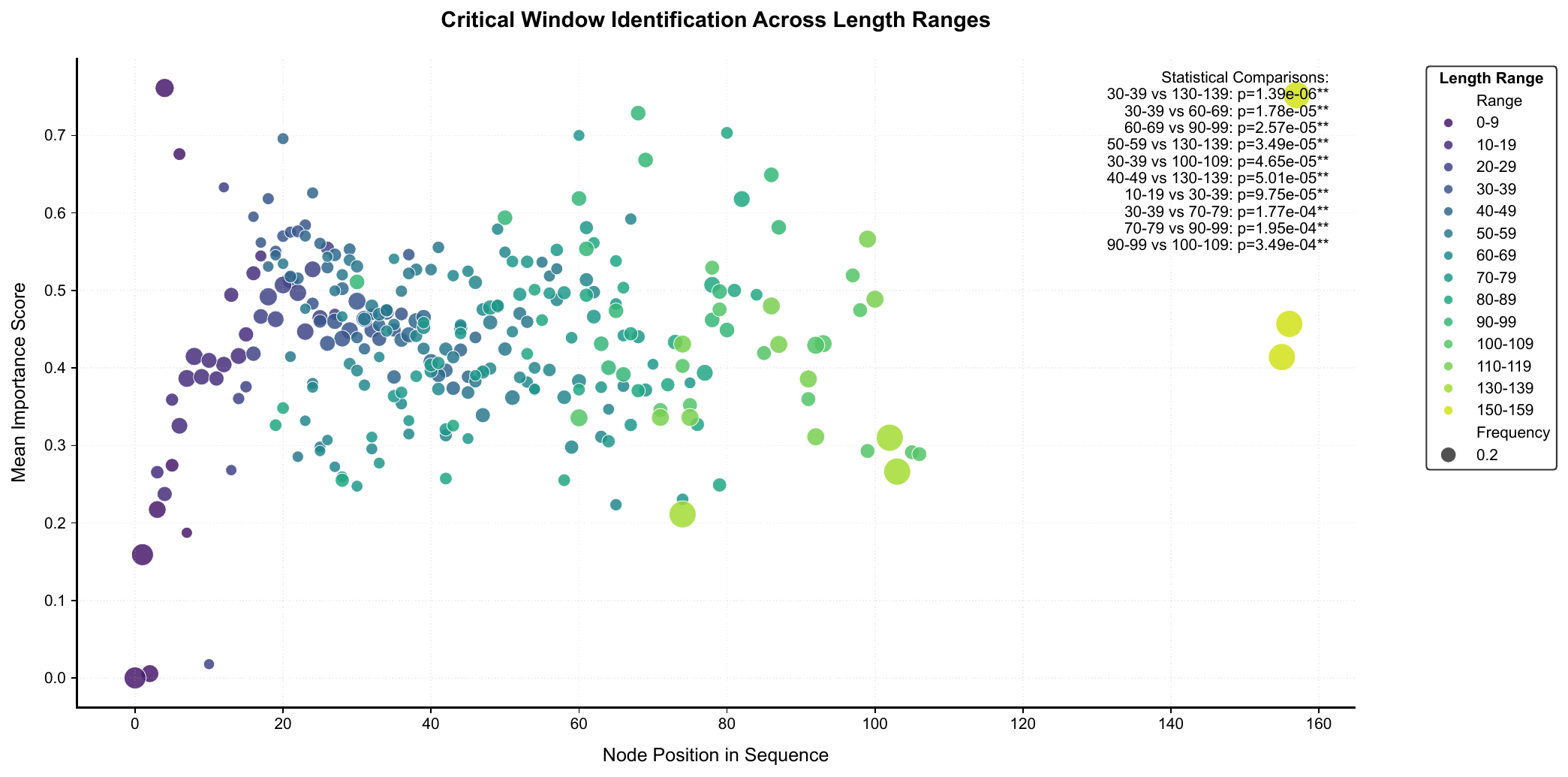}
    \caption{Dataset: BPI12; Model GAT-TDTE;}
    \label{app_subfig:cw12t}
\end{subfigure}

\vspace{0.5em}
\begin{subfigure}{0.48\textwidth}
    \includegraphics[width=\textwidth, trim={0 0 0 1cm}, clip]{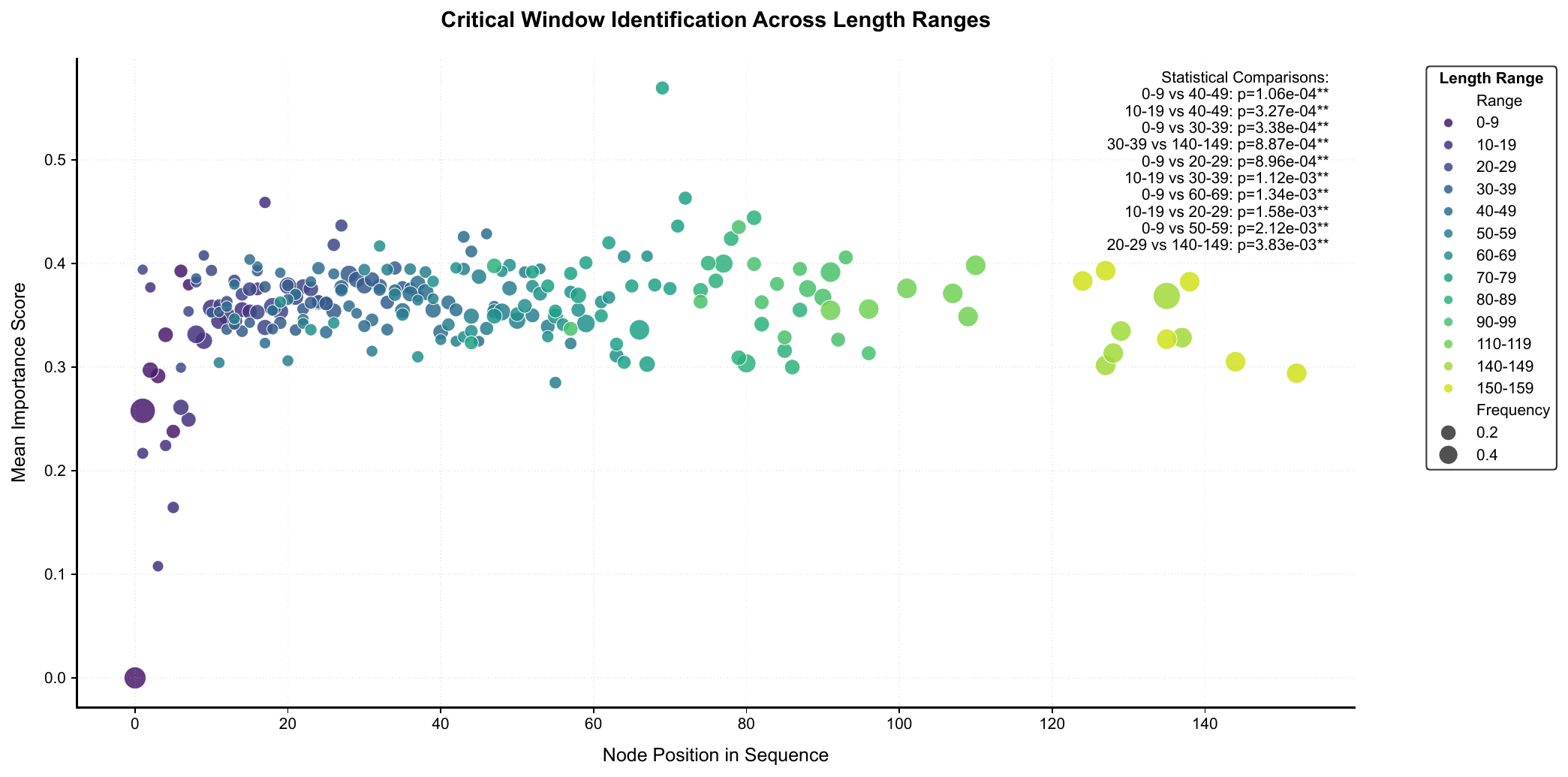}
    \caption{Dataset: BPI12W; Model: GAT-TD}
    \label{app_subfig:cw12w}
\end{subfigure}
\hfill
\begin{subfigure}{0.48\textwidth}
    \includegraphics[width=\textwidth, trim={0 0 0 1cm}, clip]{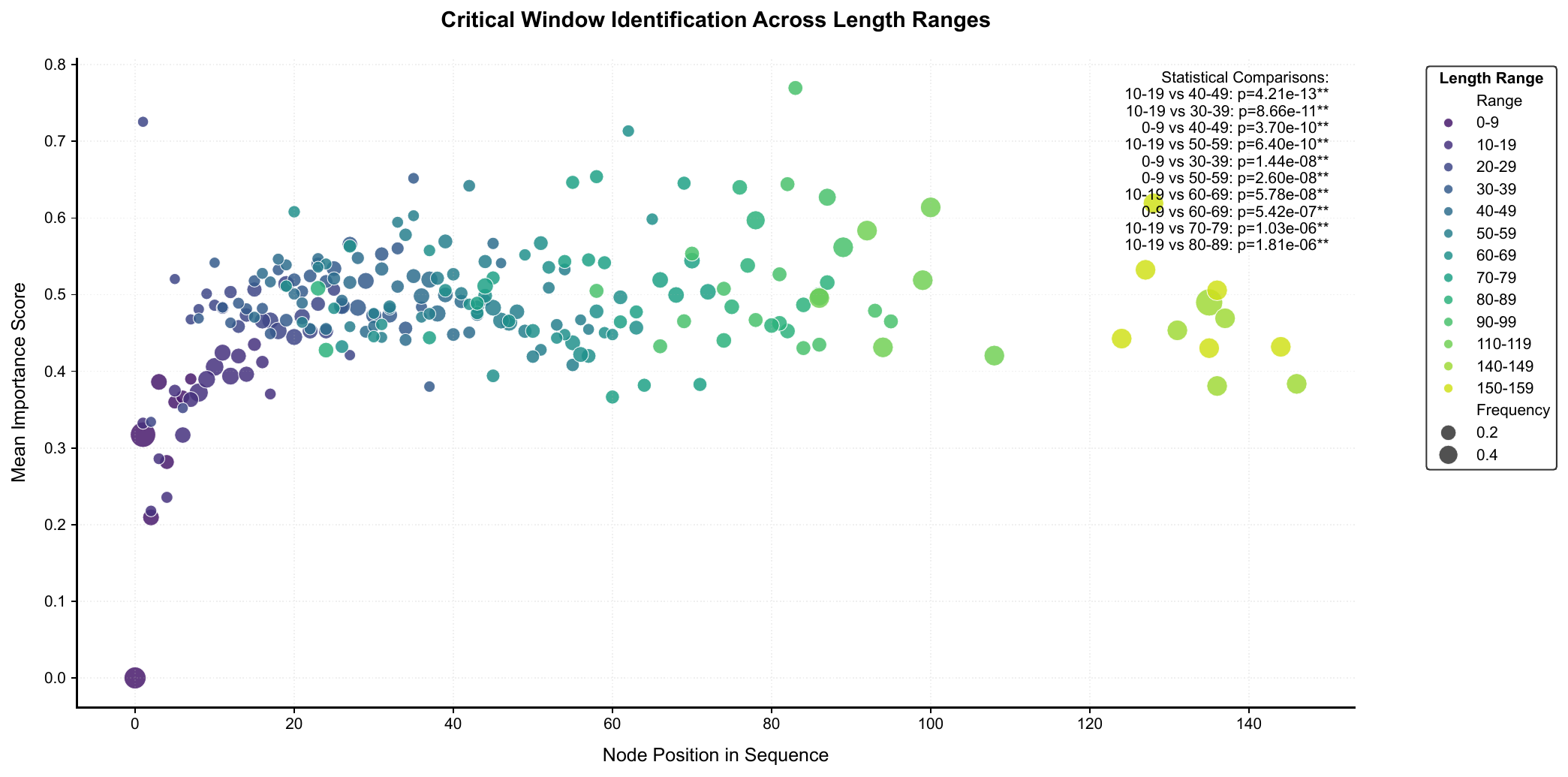}
    \caption{Dataset: BPI12W; Model GAT-TDTE}
    \label{app_subfig:cw12wt}
\end{subfigure}

\caption{Critical Windows Identification across Length Ranges (part 1)}
\label{fig:cw5x2grid_p1}
\end{figure}

\begin{figure}[htbp!]
\centering
\begin{subfigure}{0.48\textwidth}
    \includegraphics[width=\textwidth, trim={0 0 0 1cm}, clip]{BPI13i_timedgedecay_criticwin_comp.pdf}
    \caption{Dataset: BPI13i; Model: GAT-TD}
    \label{app_subfig:cw13i}
\end{subfigure}
\hfill
\begin{subfigure}{0.48\textwidth}
    \includegraphics[width=\textwidth, trim={0 0 0 1cm}, clip]{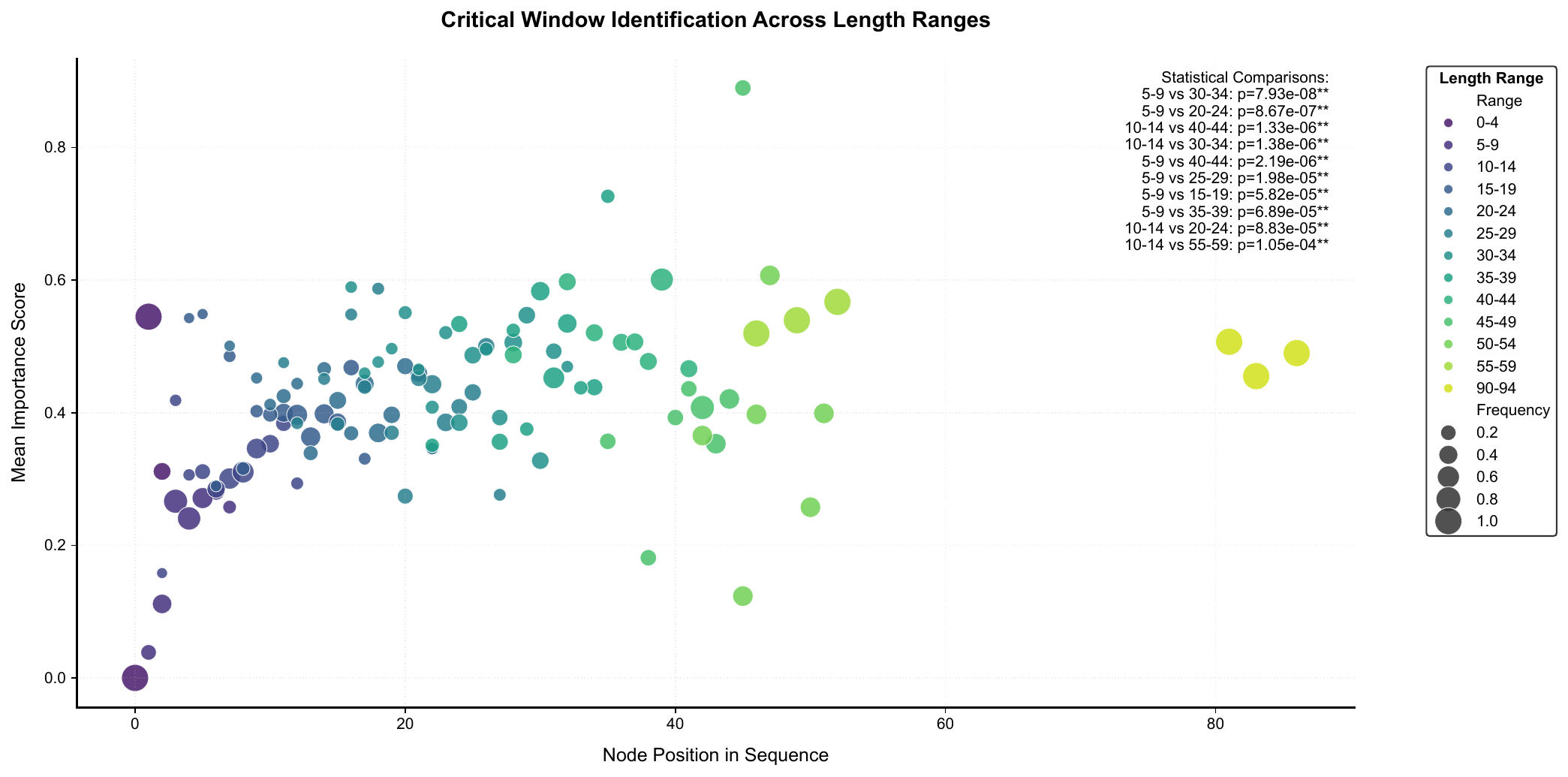}
    \caption{Dataset: BPI13i; Model GAT-TDTE}
    \label{app_subfig:cw13it}
\end{subfigure}

\vspace{0.5em}
\begin{subfigure}{0.48\textwidth}
    \includegraphics[width=\textwidth, trim={0 0 0 1cm}, clip]{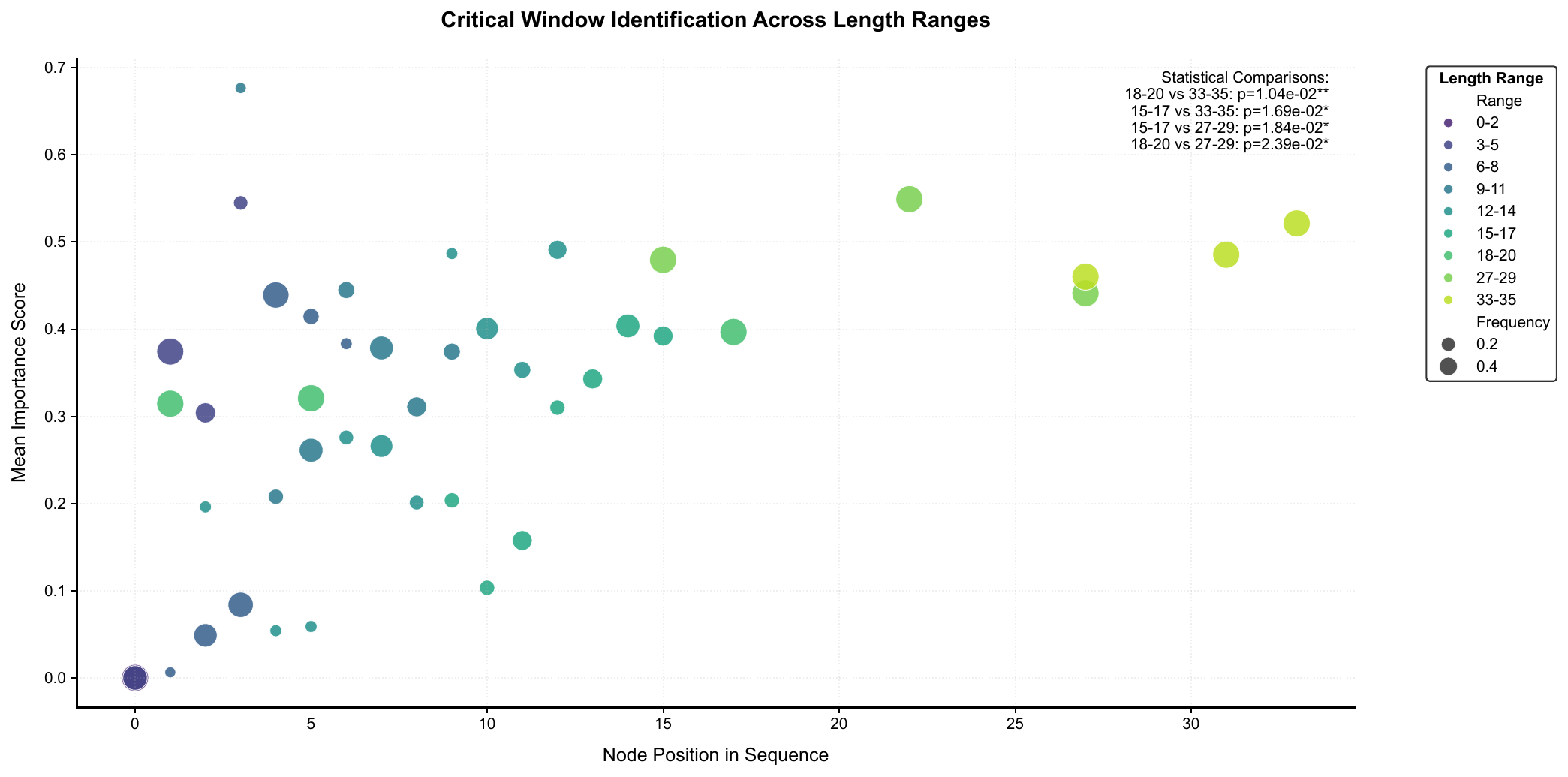}
    \caption{Dataset: BPI13c; Model: GAT-TD}
    \label{app_subfig:cw13c}
\end{subfigure}
\hfill
\begin{subfigure}{0.48\textwidth}
    \includegraphics[width=\textwidth, trim={0 0 0 1cm}, clip]{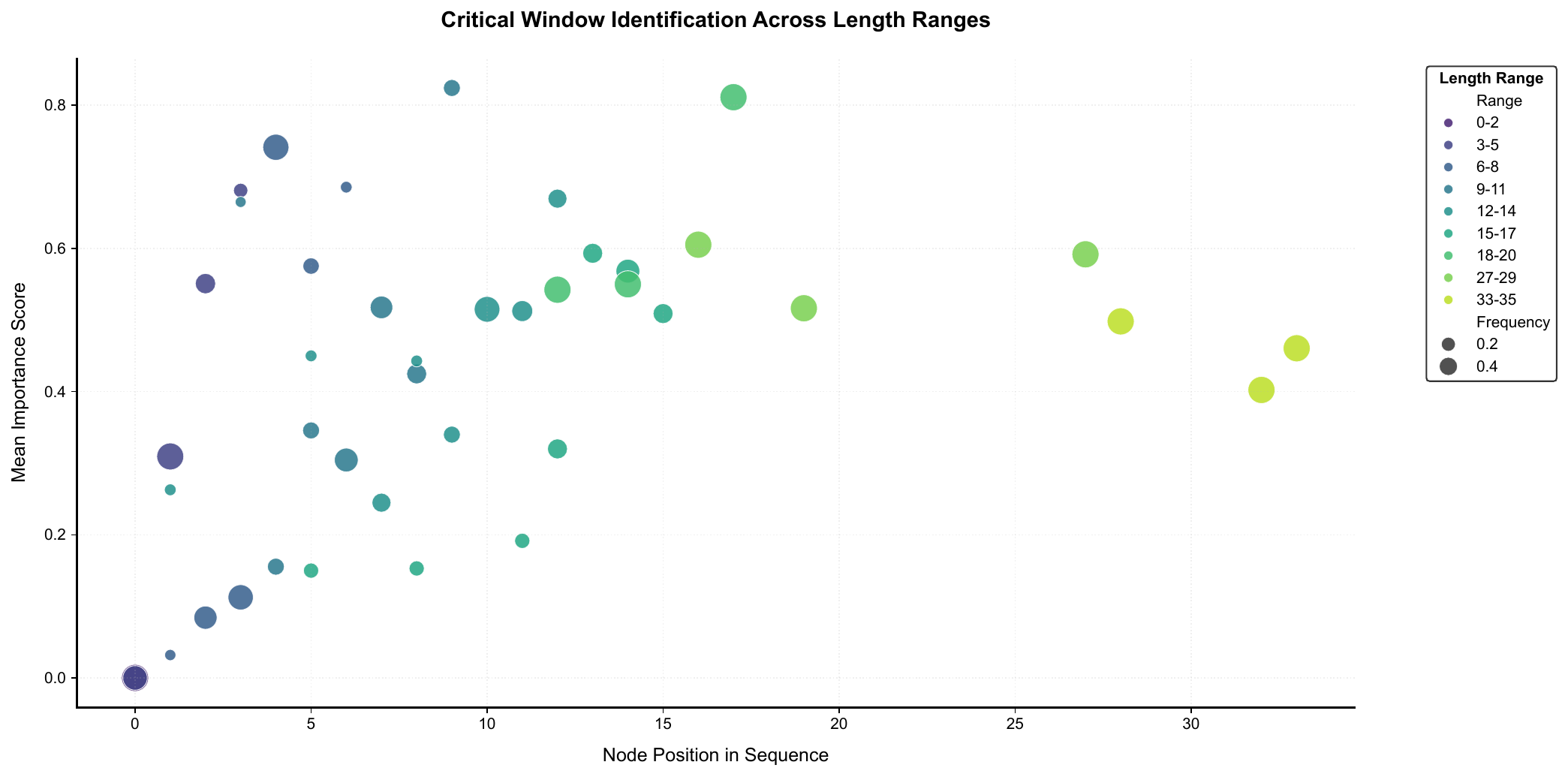}
    \caption{Dataset: BPI13c; Model GAT-TDTE}
    \label{app_subfig:cw13ct}
\end{subfigure}

\vspace{0.5em}
\begin{subfigure}{0.48\textwidth}
    \includegraphics[width=\textwidth, trim={0 0 0 1cm}, clip]{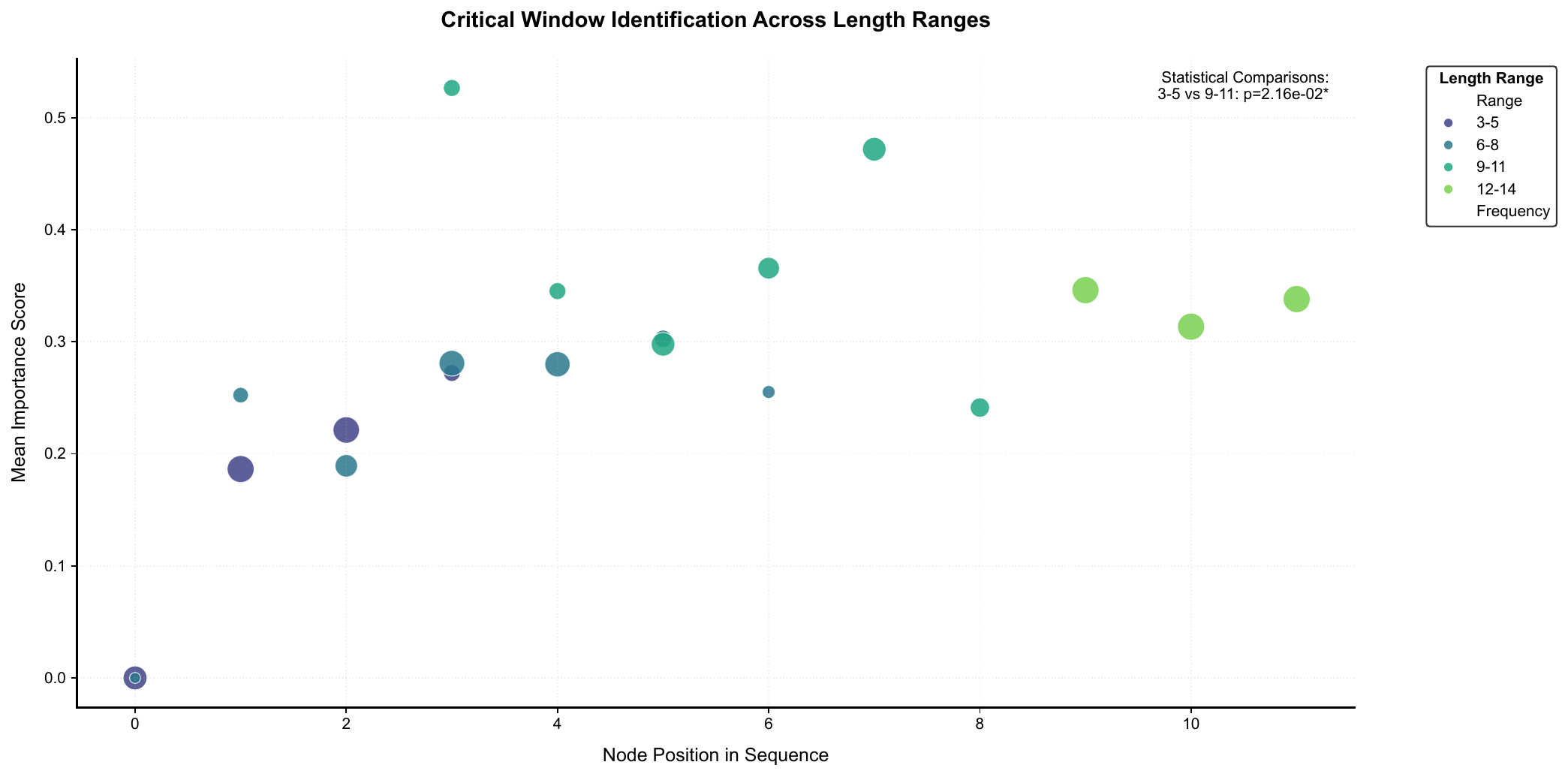}
    \caption{Dataset: Helpdeask; Model: GAT-TD}
    \label{app_subfig:cw0}
\end{subfigure}
\hfill
\begin{subfigure}{0.48\textwidth}
    \includegraphics[width=\textwidth, trim={0 0 0 1cm}, clip]{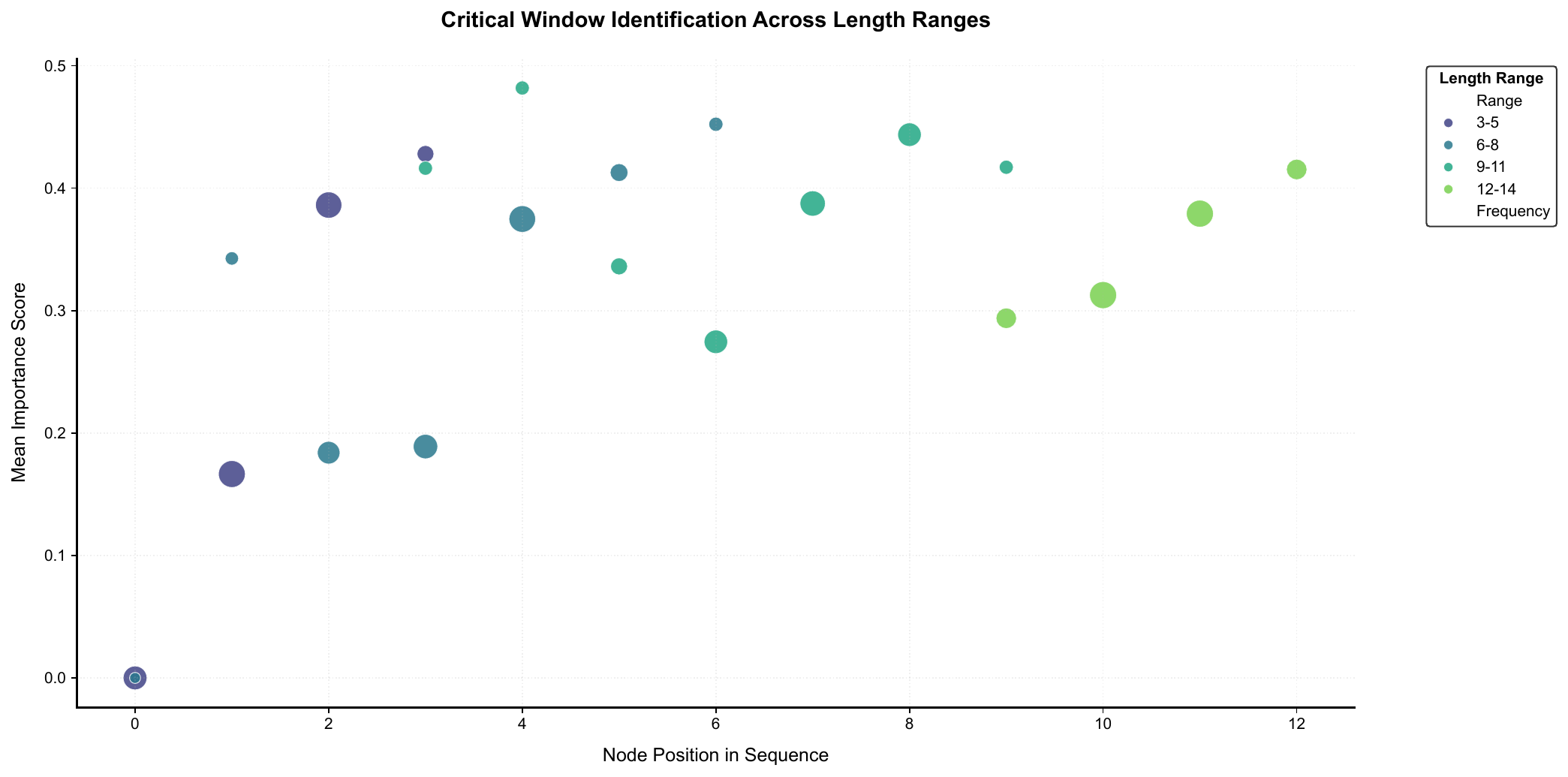}
    \caption{Dataset: Helpdesk; Model GAT-TDTE}
    \label{app_subfig:cw0t}
\end{subfigure}

\vspace{0.5em}
\begin{subfigure}{0.48\textwidth}
    \includegraphics[width=\textwidth, trim={0 0 0 1cm}, clip]{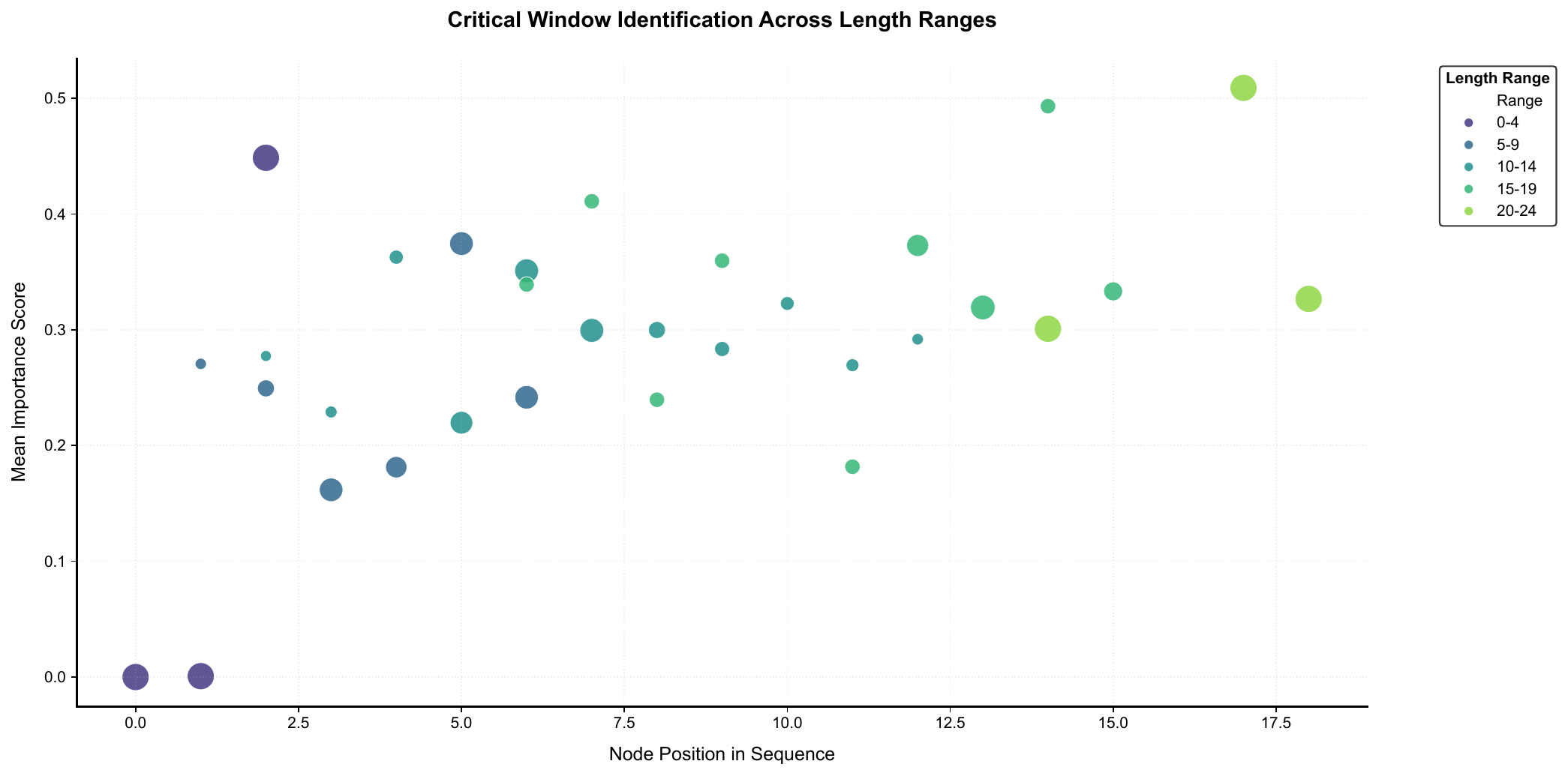}
    \caption{Dataset: BPI20; Model: GAT-TD}
    \label{app_subfig:cw20}
\end{subfigure}
\hfill
\begin{subfigure}{0.48\textwidth}
    \includegraphics[width=\textwidth, trim={0 0 0 1cm}, clip]{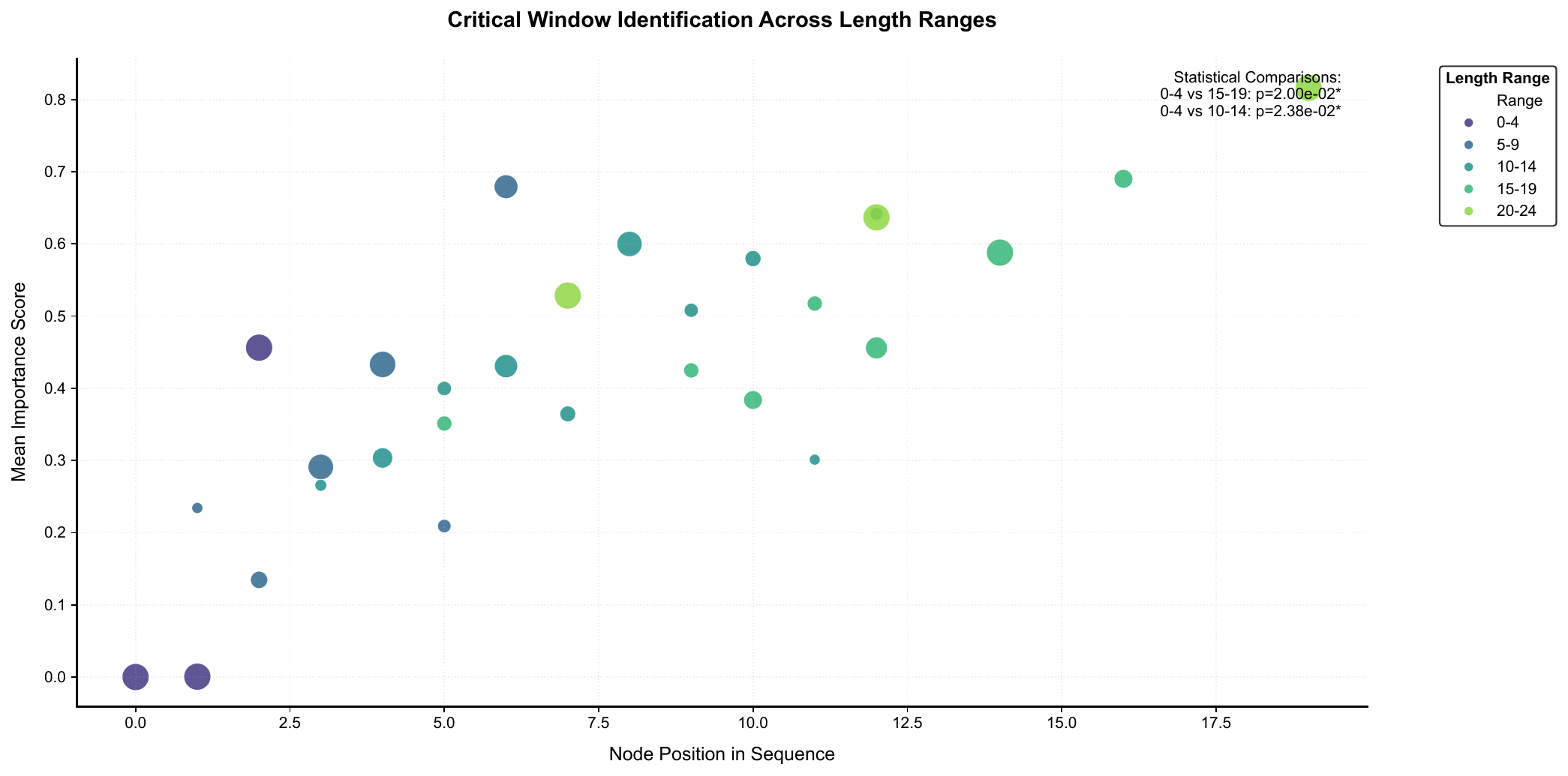}
    \caption{Dataset: BPI20; Model GAT-TDTE}
    \label{app_subfig:cw20t}
\end{subfigure}

\caption{Critical Windows Identification across Length Ranges (part 2)}
\label{fig:cw5x2grid_p2}
\end{figure}

\begin{figure}[htbp!]
\centering

\begin{subfigure}{0.48\textwidth}
    \includegraphics[width=\textwidth, trim={0 0 0 1.5cm}, clip]{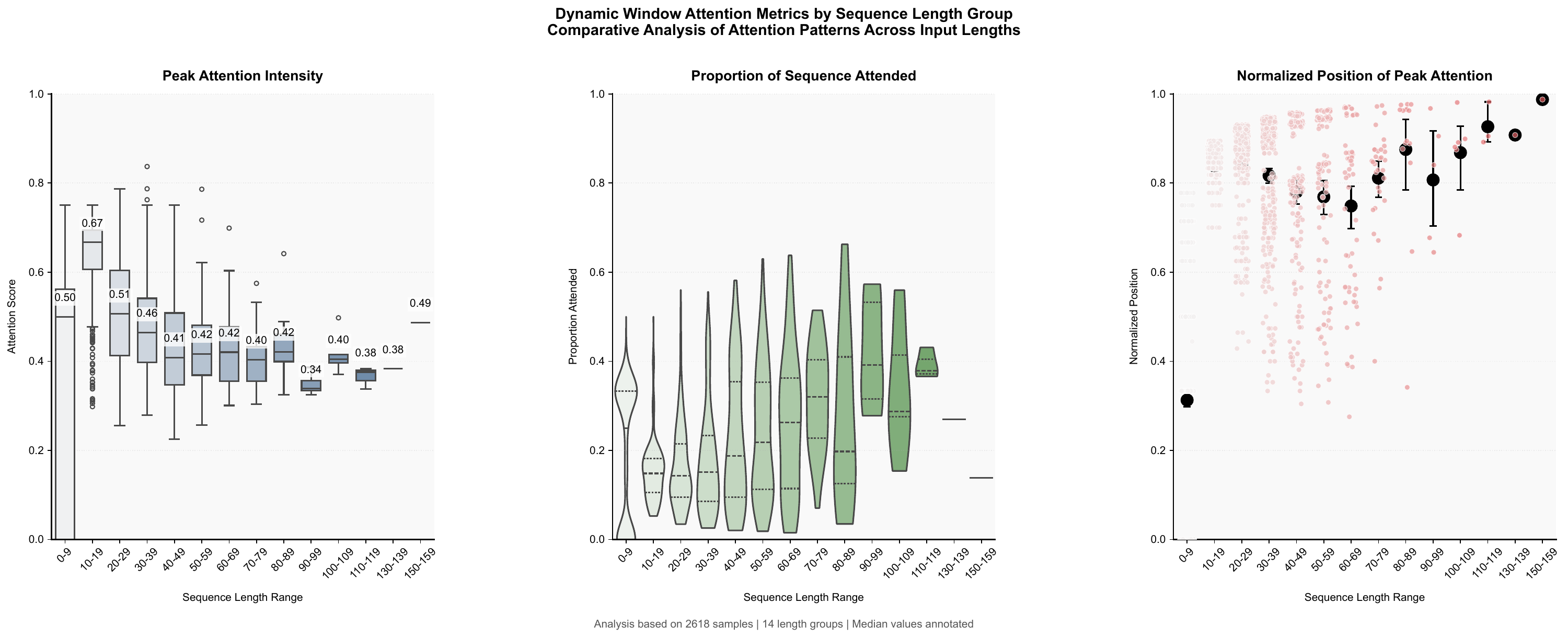}
    \caption{Dataset: BPI12; Model: GAT-TD}
    \label{app_subfig:at12}
\end{subfigure}
\hfill
\begin{subfigure}{0.48\textwidth}
    \includegraphics[width=\textwidth, trim={0 0 0 1.5cm}, clip]{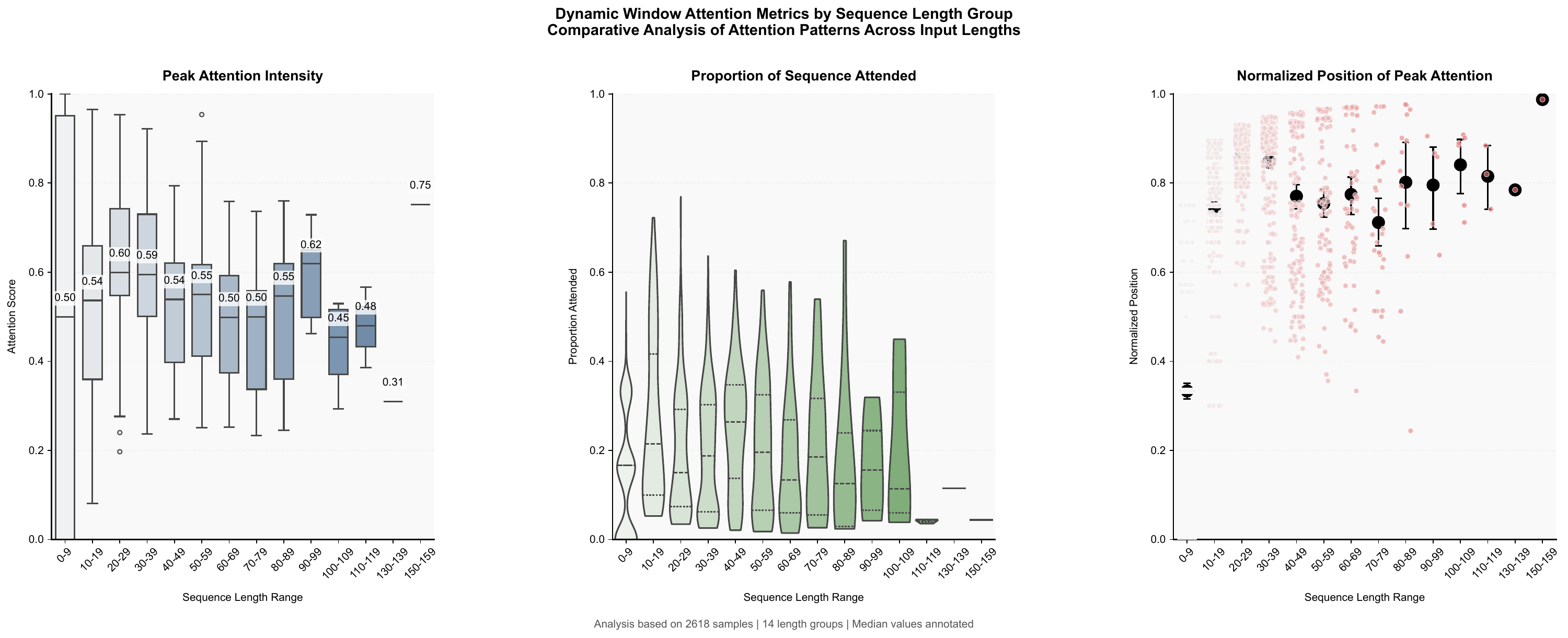}
    \caption{Dataset: BPI12; Model GAT-TDTE;}
    \label{app_subfig:at12t}
\end{subfigure}

\vspace{0.5em}
\begin{subfigure}{0.48\textwidth}
    \includegraphics[width=\textwidth, trim={0 0 0 1.5cm}, clip]{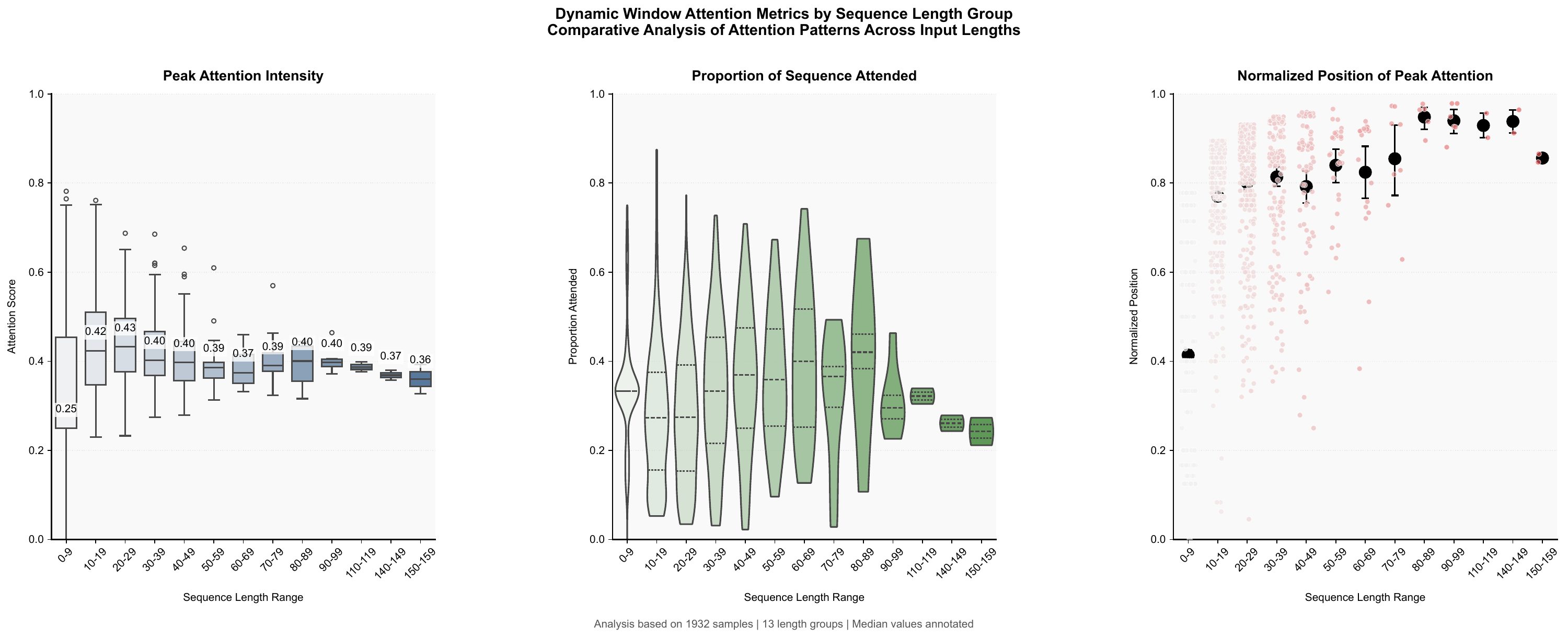}
    \caption{Dataset: BPI12W; Model: GAT-TD}
    \label{app_subfig:at12w}
\end{subfigure}
\hfill
\begin{subfigure}{0.48\textwidth}
    \includegraphics[width=\textwidth, trim={0 0 0 1.5cm}, clip]{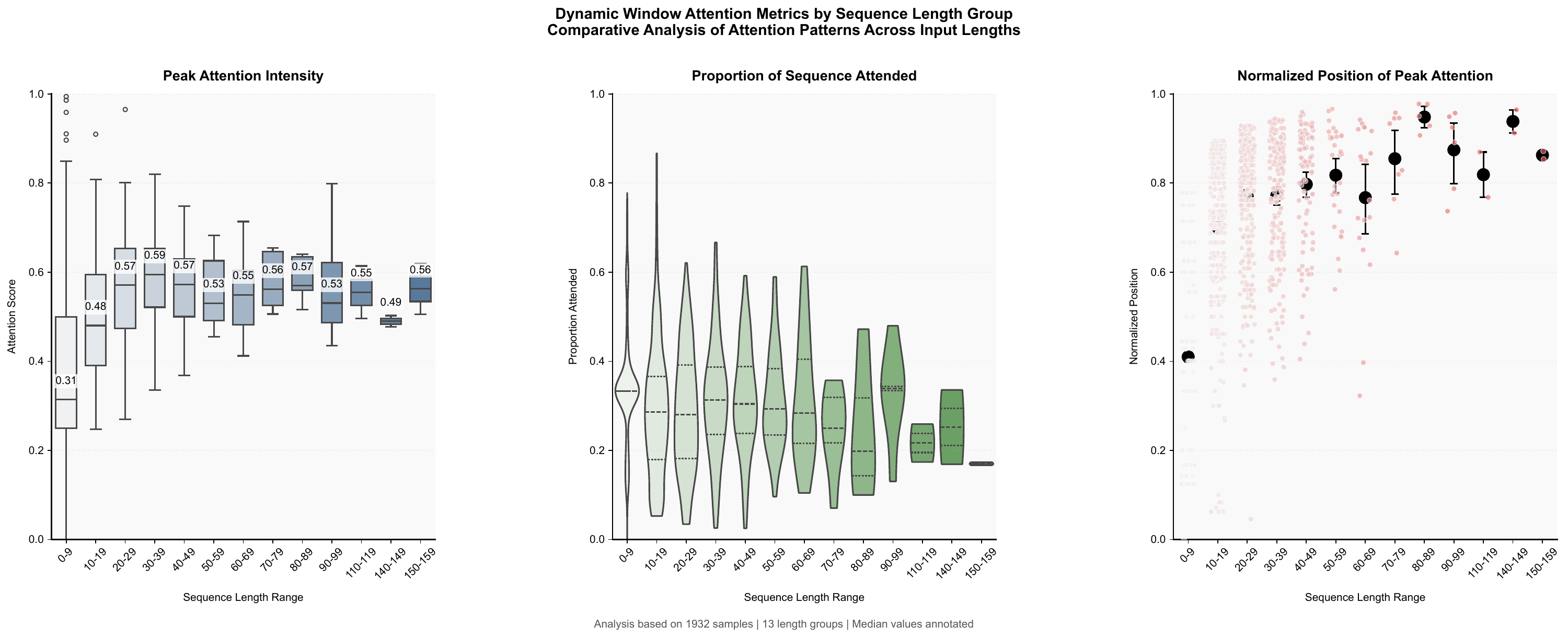}
    \caption{Dataset: BPI12W; Model GAT-TDTE}
    \label{app_subfig:at12wt}
\end{subfigure}

\vspace{0.5em}
\begin{subfigure}{0.48\textwidth}
    \includegraphics[width=\textwidth, trim={0 0 0 1.5cm}, clip]{BPI13i_timedgedecay_attentionstatis.pdf}
    \caption{Dataset: BPI13i; Model: GAT-TD}
    \label{app_subfig:at13i}
\end{subfigure}
\hfill
\begin{subfigure}{0.48\textwidth}
    \includegraphics[width=\textwidth, trim={0 0 0 1.5cm}, clip]{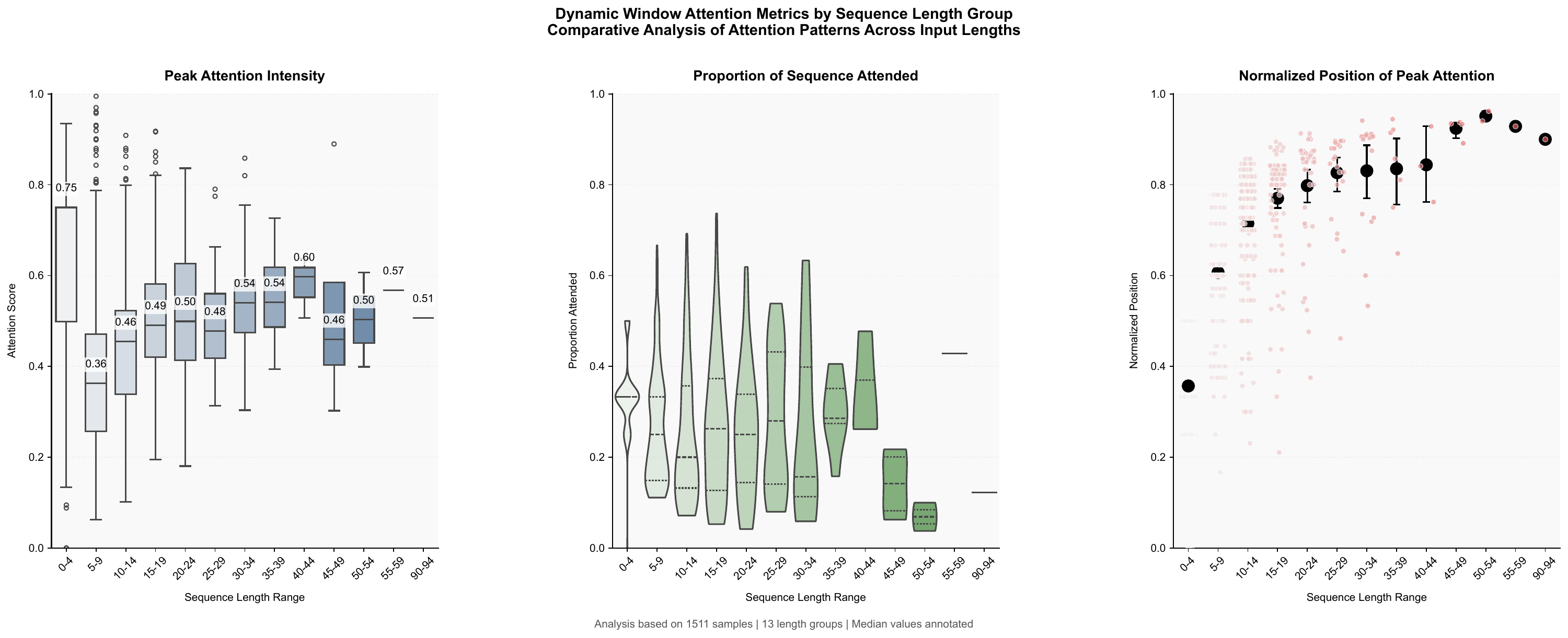}
    \caption{Dataset: BPI13i; Model GAT-TDTE}
    \label{app_subfig:at13it}
\end{subfigure}

\vspace{0.5em}
\begin{subfigure}{0.48\textwidth}
    \includegraphics[width=\textwidth, trim={0 0 0 1.5cm}, clip]{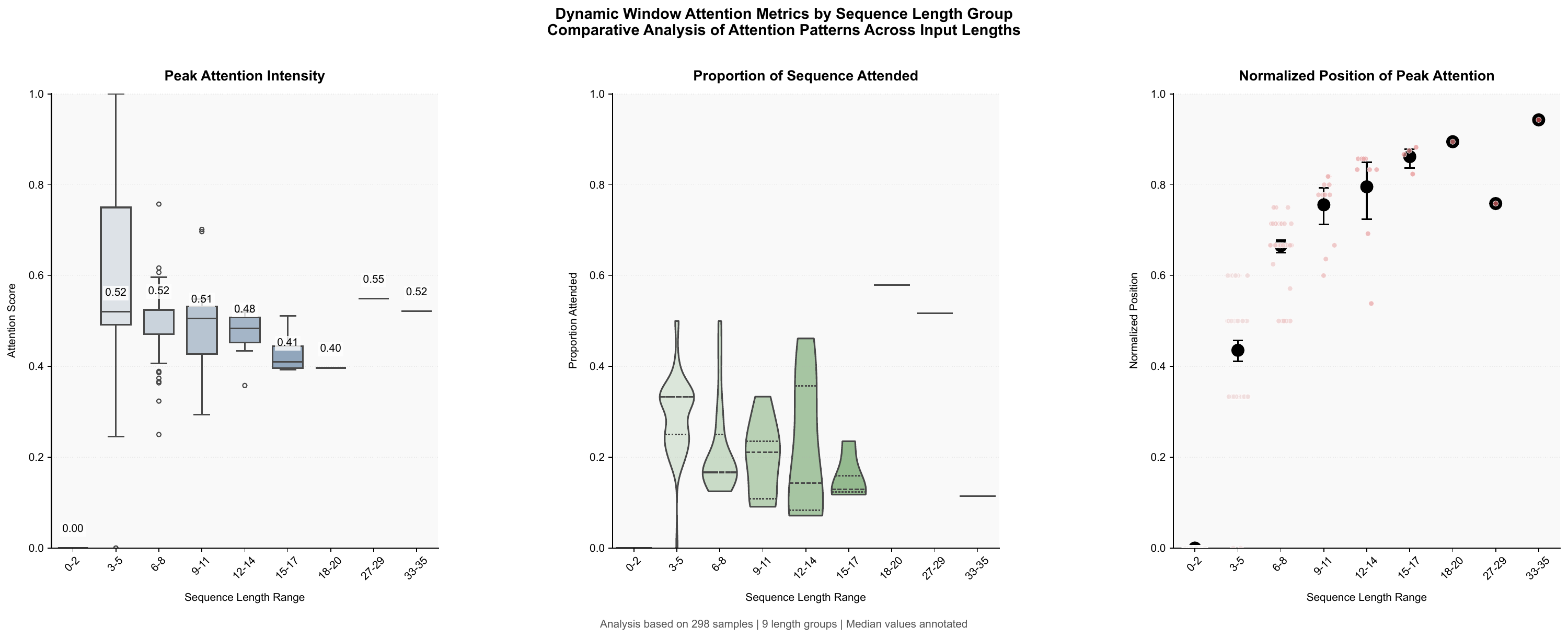}
    \caption{Dataset: BPI13c; Model: GAT-TD}
    \label{app_subfig:at13c}
\end{subfigure}
\hfill
\begin{subfigure}{0.48\textwidth}
    \includegraphics[width=\textwidth, trim={0 0 0 1.5cm}, clip]{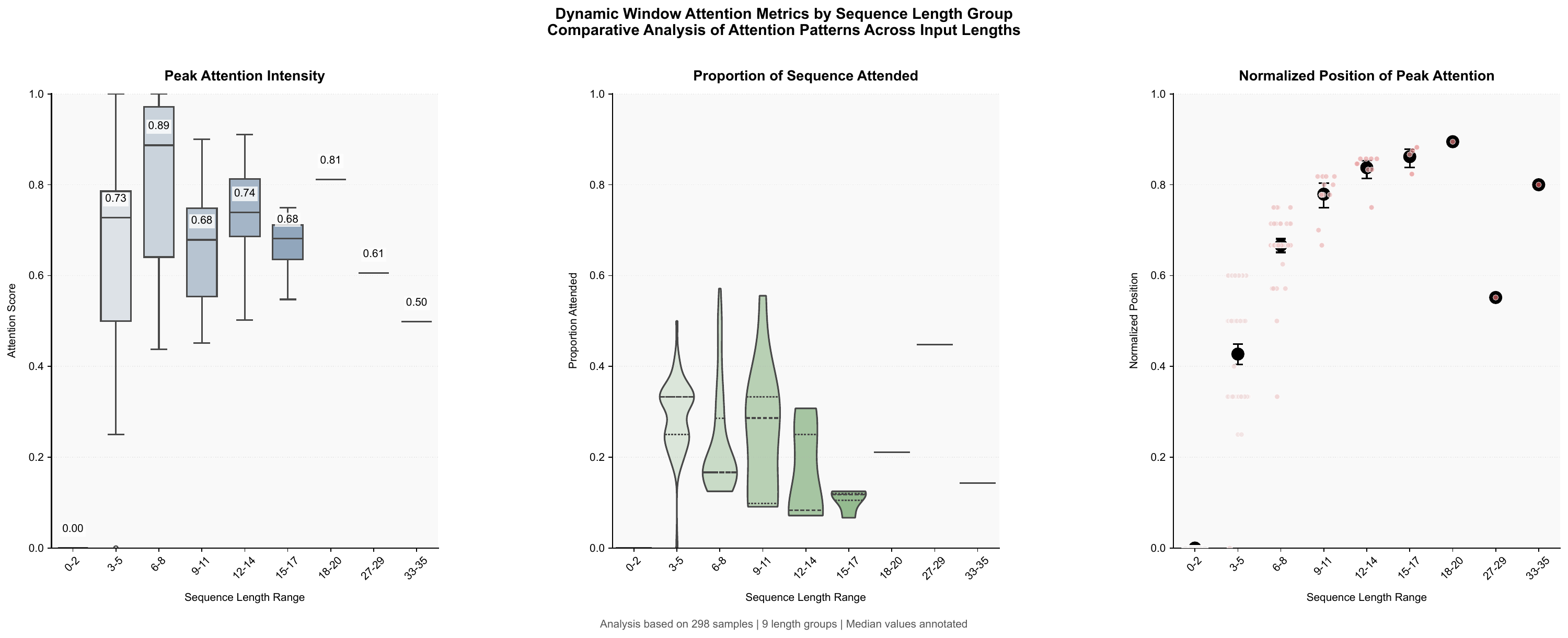}
    \caption{Dataset: BPI13c; Model GAT-TDTE}
    \label{app_subfig:at13ct}
\end{subfigure}

\vspace{0.5em}
\begin{subfigure}{0.48\textwidth}
    \includegraphics[width=\textwidth, trim={0 0 0 1.5cm}, clip]{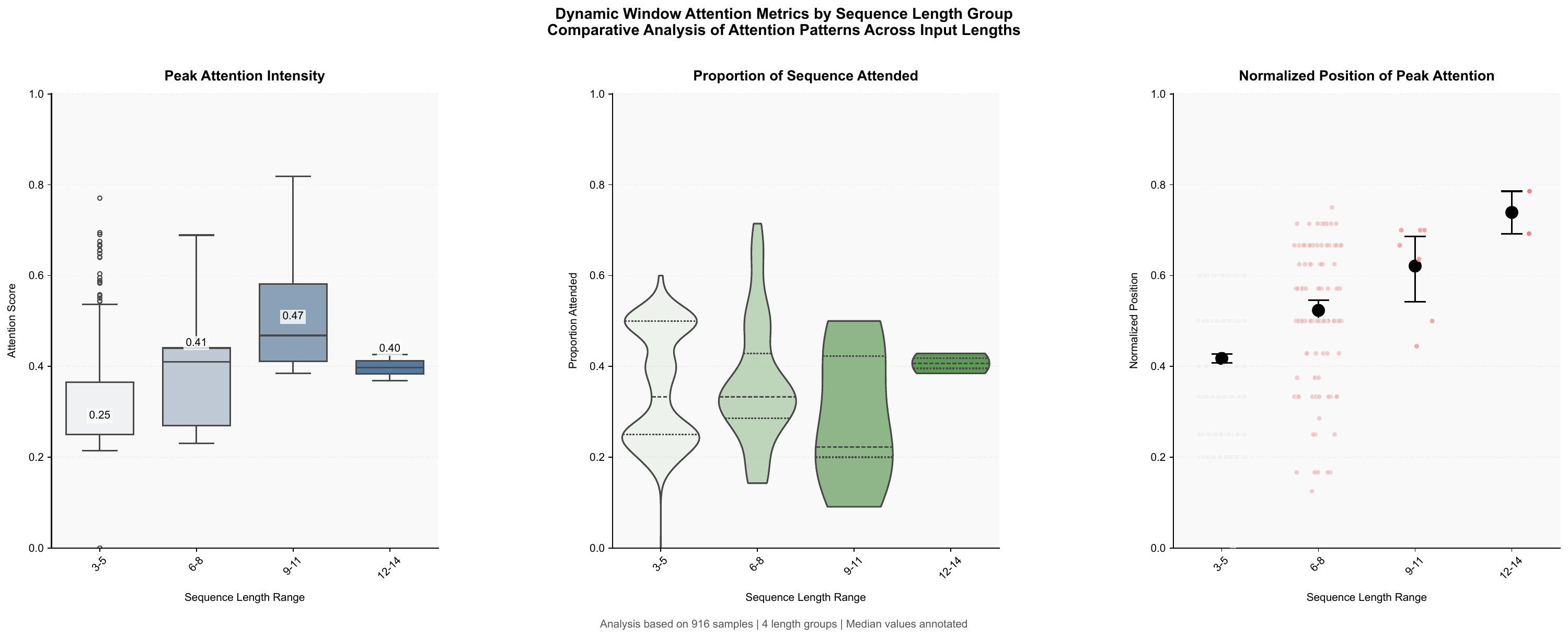}
    \caption{Dataset: Helpdeask; Model: GAT-TD}
    \label{app_subfig:at0}
\end{subfigure}
\hfill
\begin{subfigure}{0.48\textwidth}
    \includegraphics[width=\textwidth, trim={0 0 0 1.5cm}, clip]{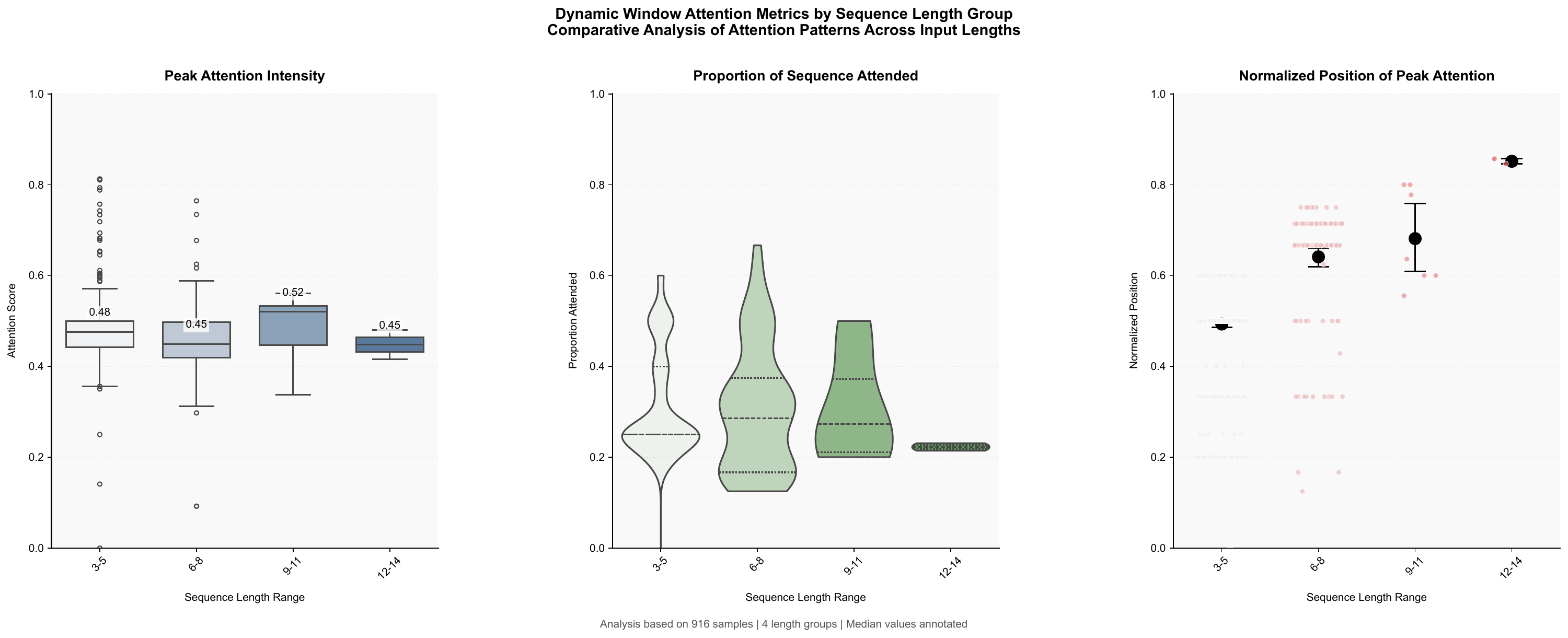}
    \caption{Dataset: Helpdesk; Model GAT-TDTE}
    \label{app_subfig:at0t}
\end{subfigure}

\vspace{0.5em}
\begin{subfigure}{0.48\textwidth}
    \includegraphics[width=\textwidth, trim={0 0 0 1.5cm}, clip]{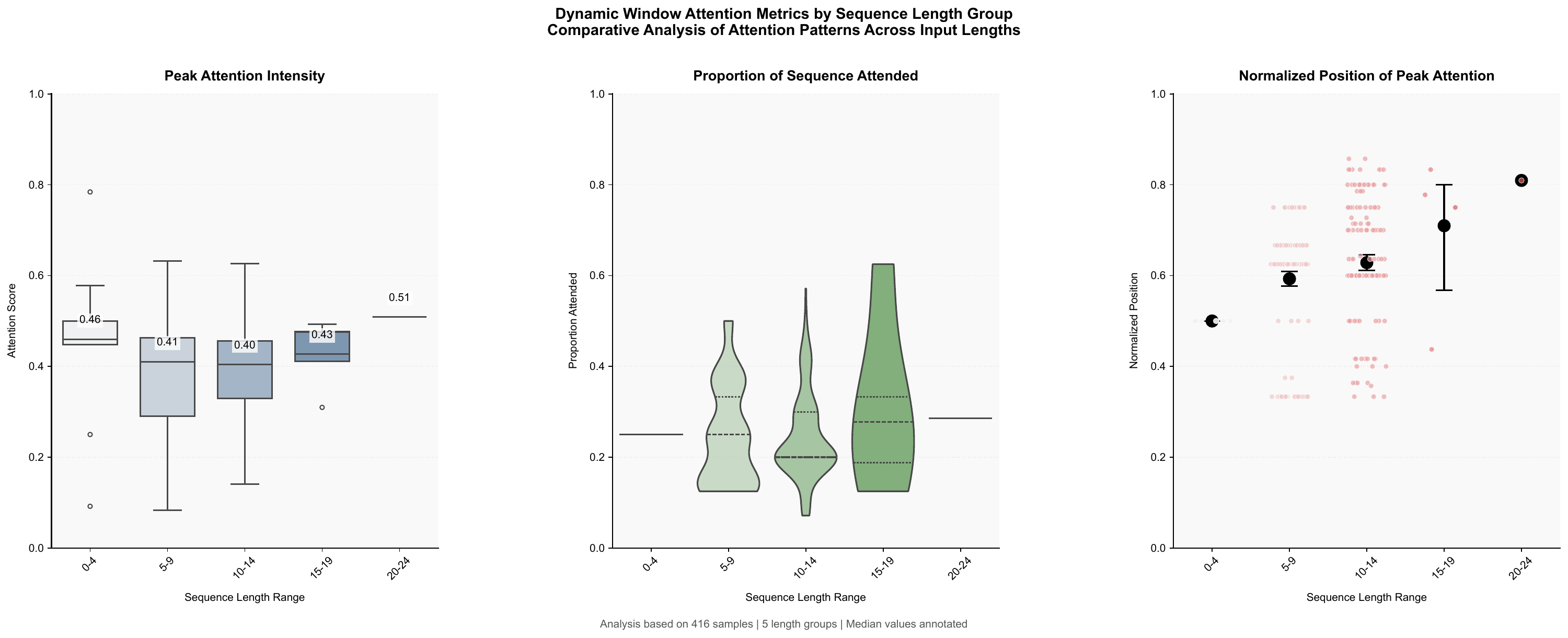}
    \caption{Dataset: BPI20; Model: GAT-TD}
    \label{app_subfig:at20}
\end{subfigure}
\hfill
\begin{subfigure}{0.48\textwidth}
    \includegraphics[width=\textwidth, trim={0 0 0 1.5cm}, clip]{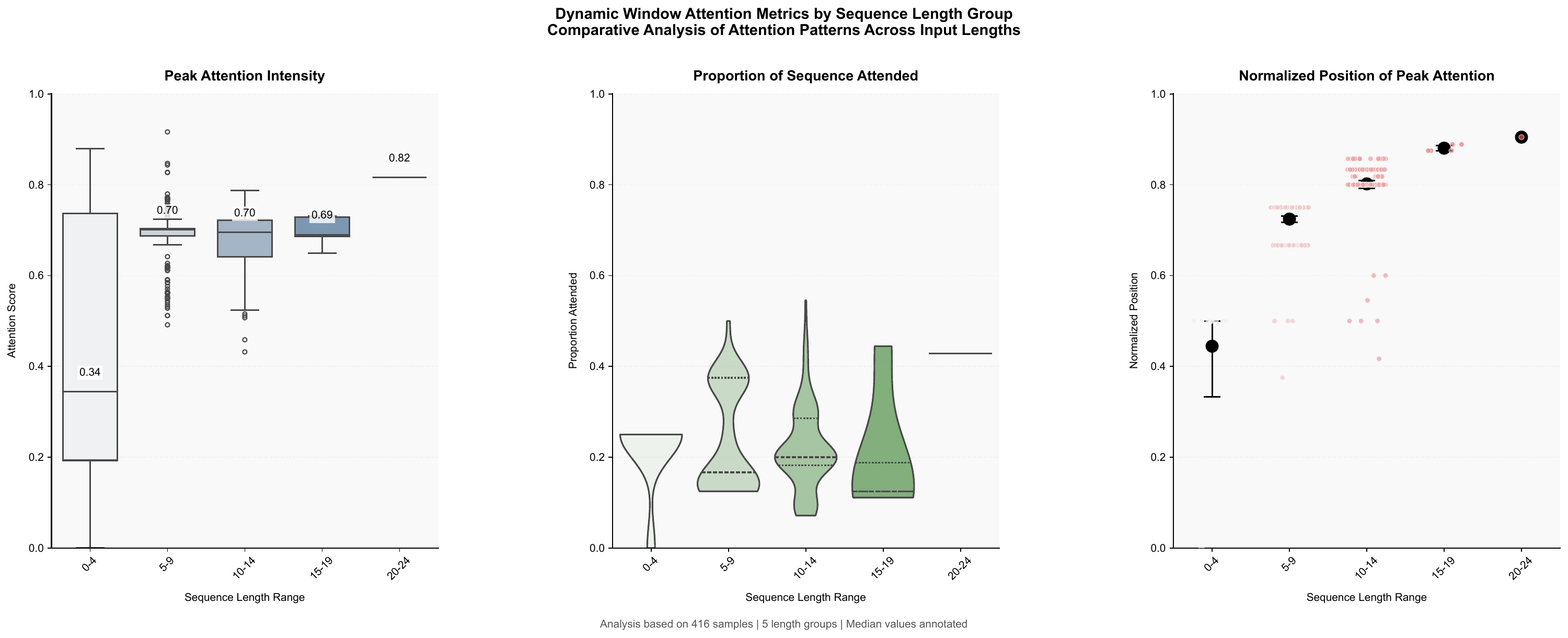}
    \caption{Dataset: BPI20; Model GAT-TDTE}
    \label{app_subfig:at20t}
\end{subfigure}

\caption{Critical Windows Identification across Length Ranges}
\label{fig:at5x2grid}
\end{figure}

\begin{figure}[htbp!]
\centering

\begin{subfigure}{0.48\textwidth}
    \includegraphics[width=\textwidth, trim={0 0 0 1cm}, clip]{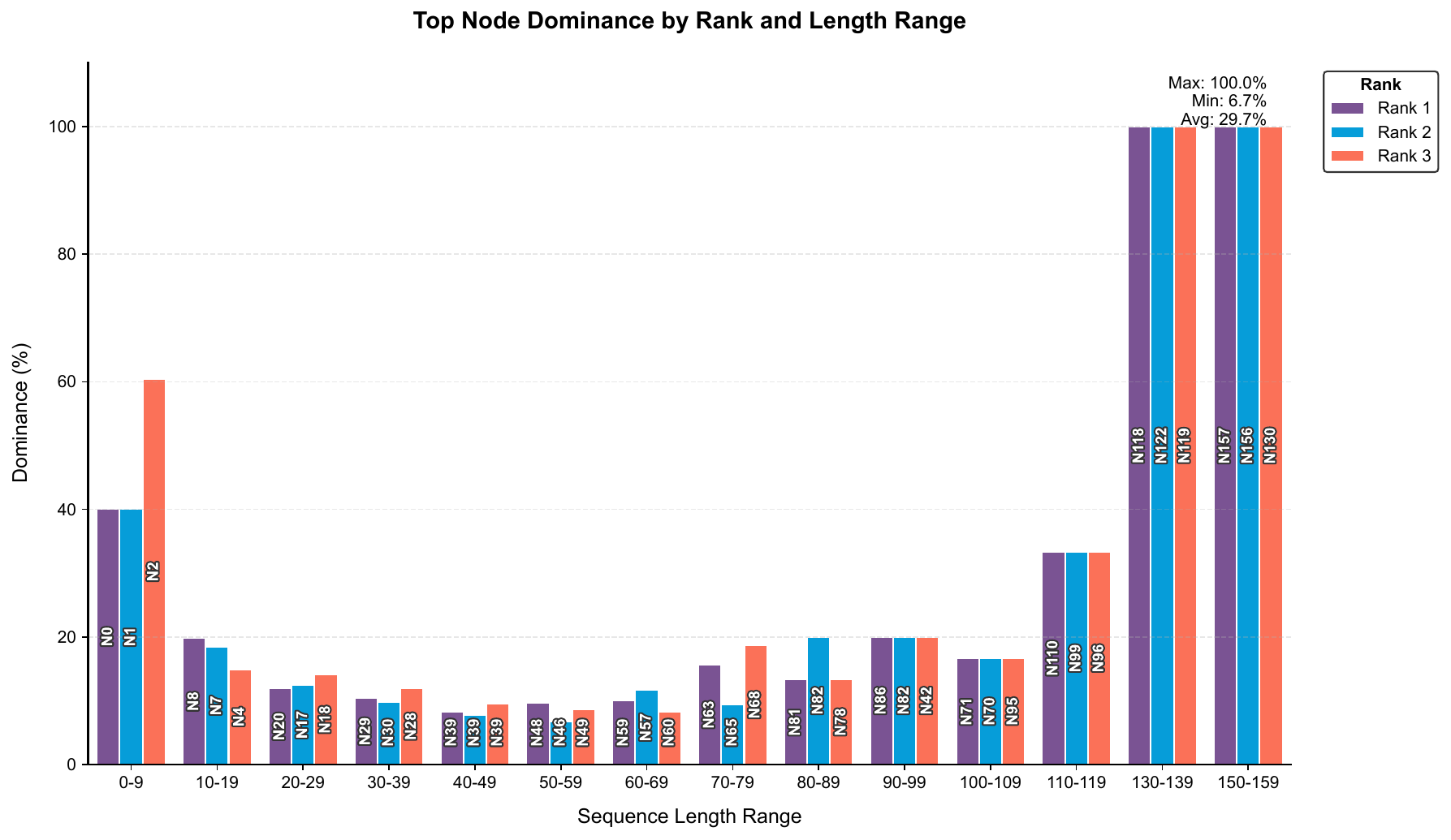}
    \caption{Dataset: BPI12; Model: GAT-TD}
    \label{app_subfig:rk12}
\end{subfigure}
\hfill
\begin{subfigure}{0.48\textwidth}
    \includegraphics[width=\textwidth, trim={0 0 0 1cm}, clip]{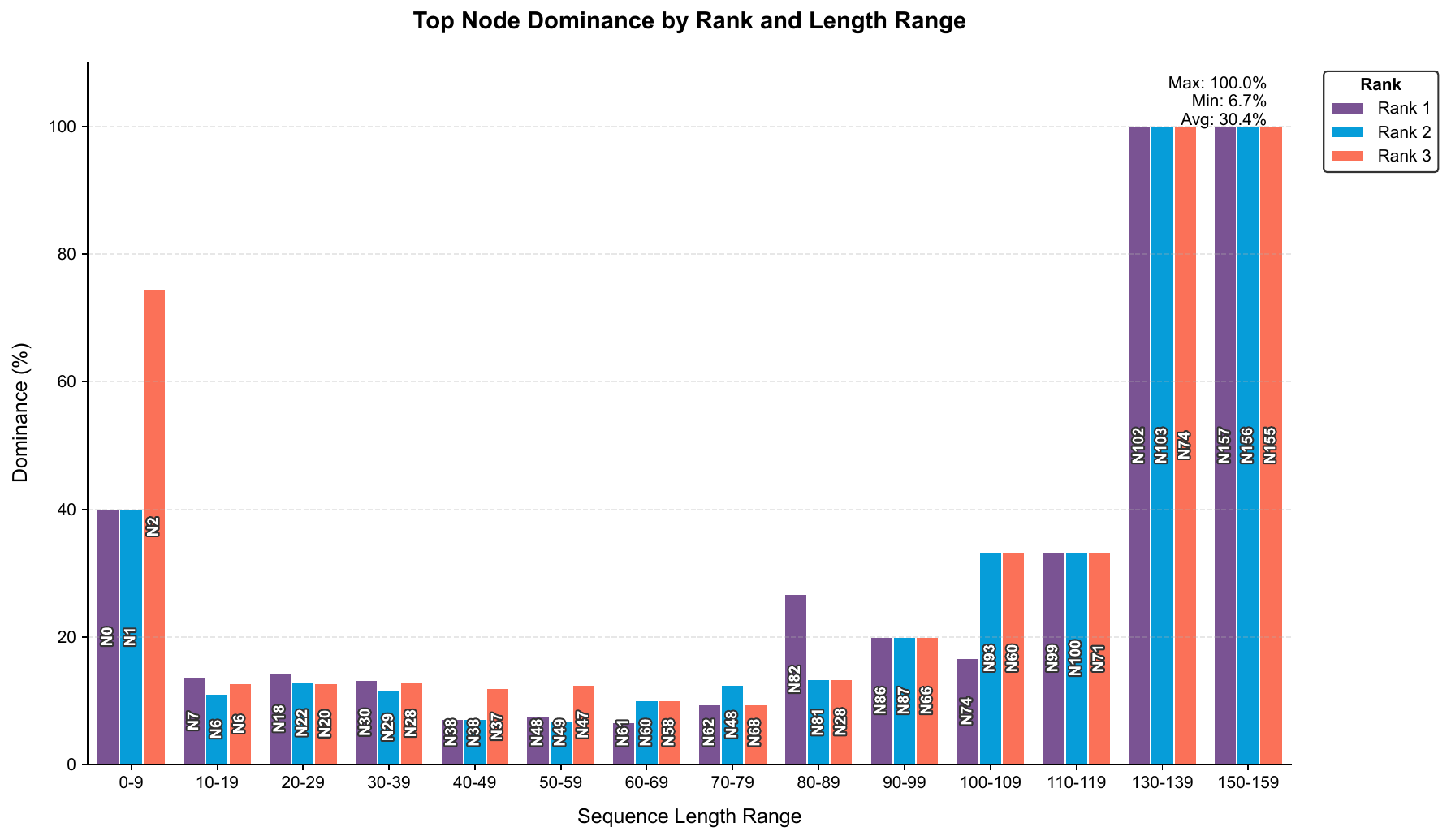}
    \caption{Dataset: BPI12; Model GAT-TDTE;}
    \label{app_subfig:rk12t}
\end{subfigure}

\vspace{0.5em}
\begin{subfigure}{0.48\textwidth}
    \includegraphics[width=\textwidth, trim={0 0 0 1cm}, clip]{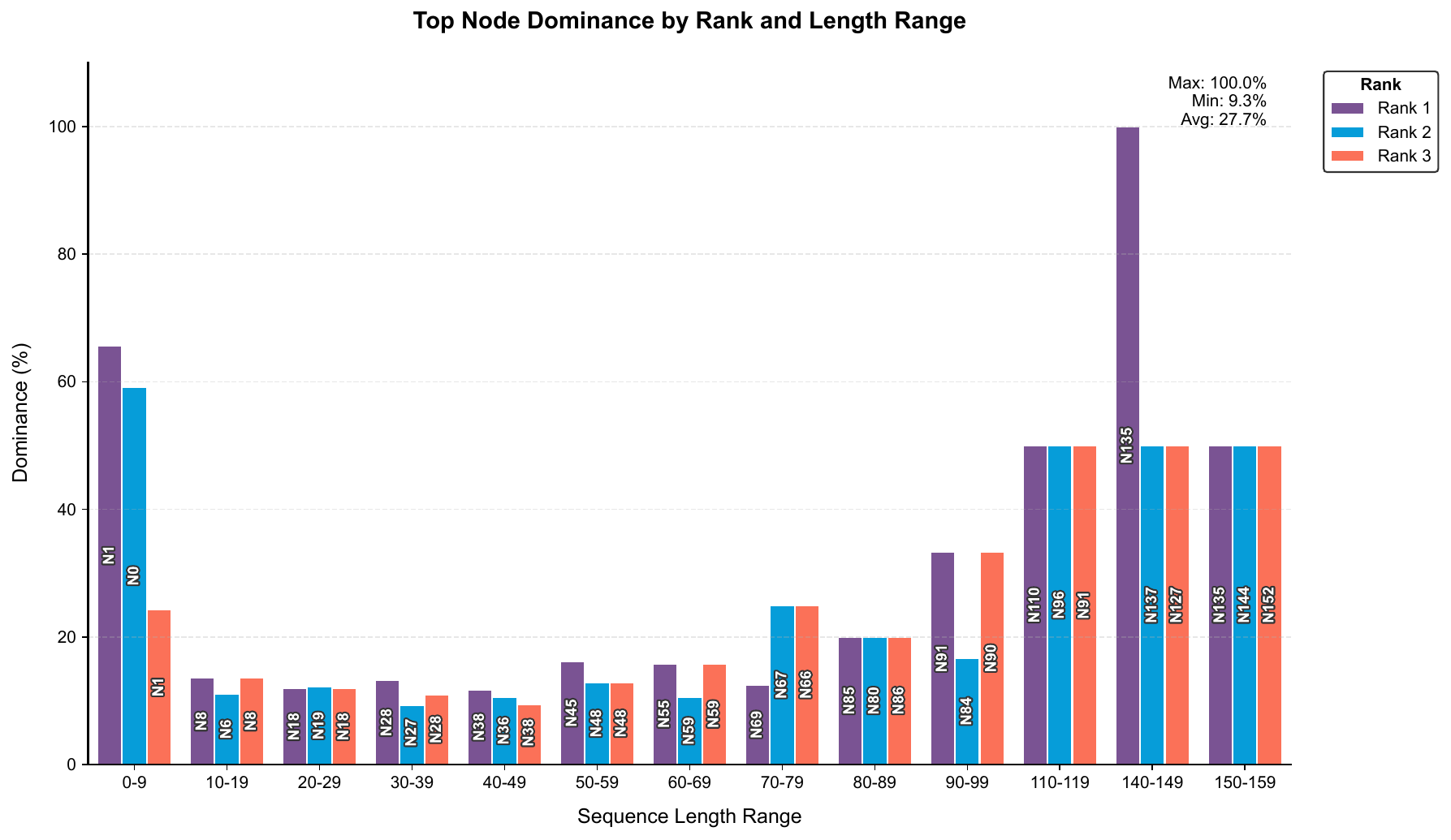}
    \caption{Dataset: BPI12W; Model: GAT-TD}
    \label{app_subfig:rk12w}
\end{subfigure}
\hfill
\begin{subfigure}{0.48\textwidth}
    \includegraphics[width=\textwidth, trim={0 0 0 1cm}, clip]{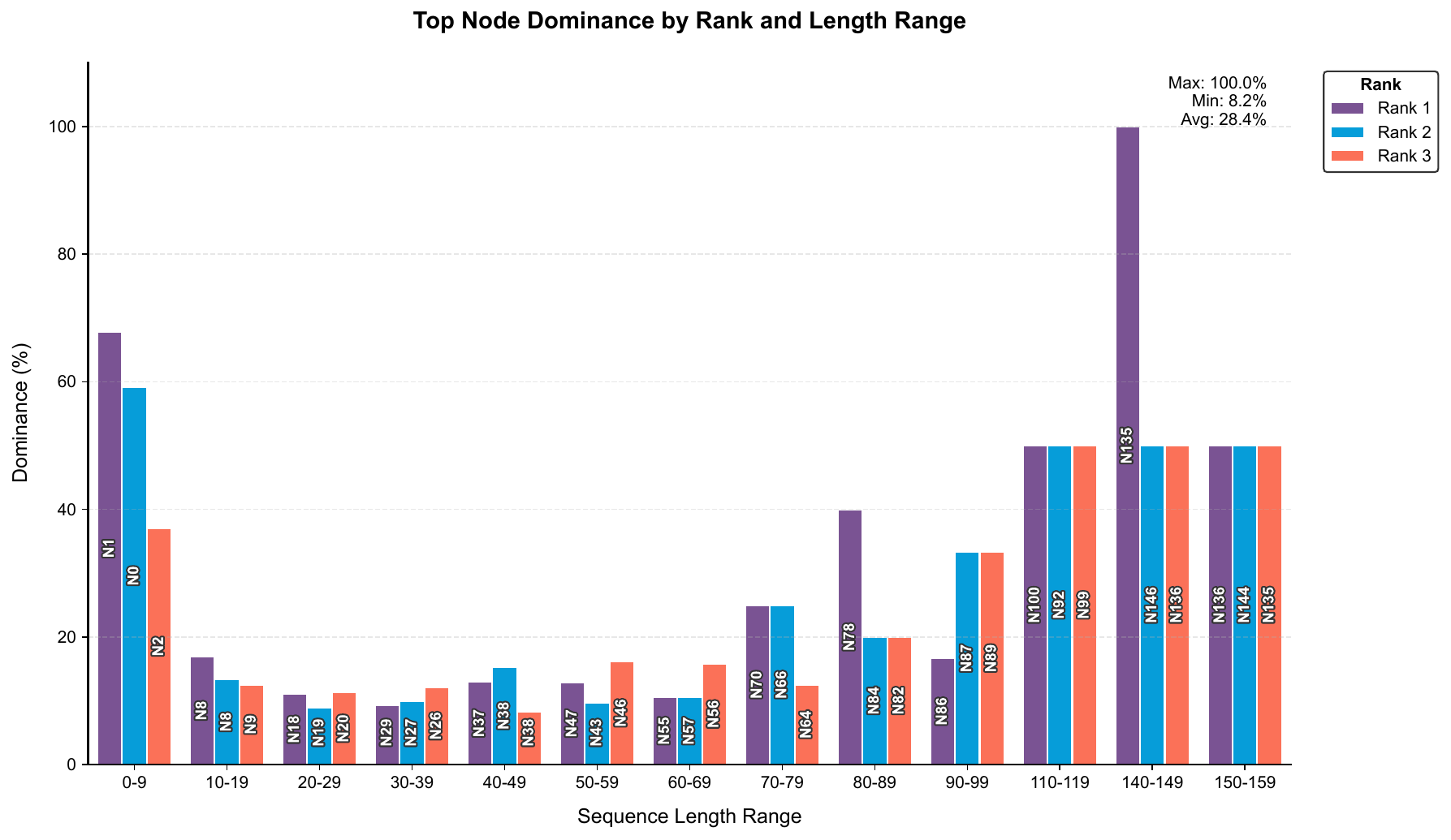}
    \caption{Dataset: BPI12W; Model GAT-TDTE}
    \label{app_subfig:rk12wt}
\end{subfigure}

\vspace{0.5em}
\begin{subfigure}{0.48\textwidth}
    \includegraphics[width=\textwidth, trim={0 0 0 1cm}, clip]{BPI13i_timeedgedecay_noderankbar.pdf}
    \caption{Dataset: BPI13i; Model: GAT-TD}
    \label{app_subfig:rk13i}
\end{subfigure}
\hfill
\begin{subfigure}{0.48\textwidth}
    \includegraphics[width=\textwidth, trim={0 0 0 1cm}, clip]{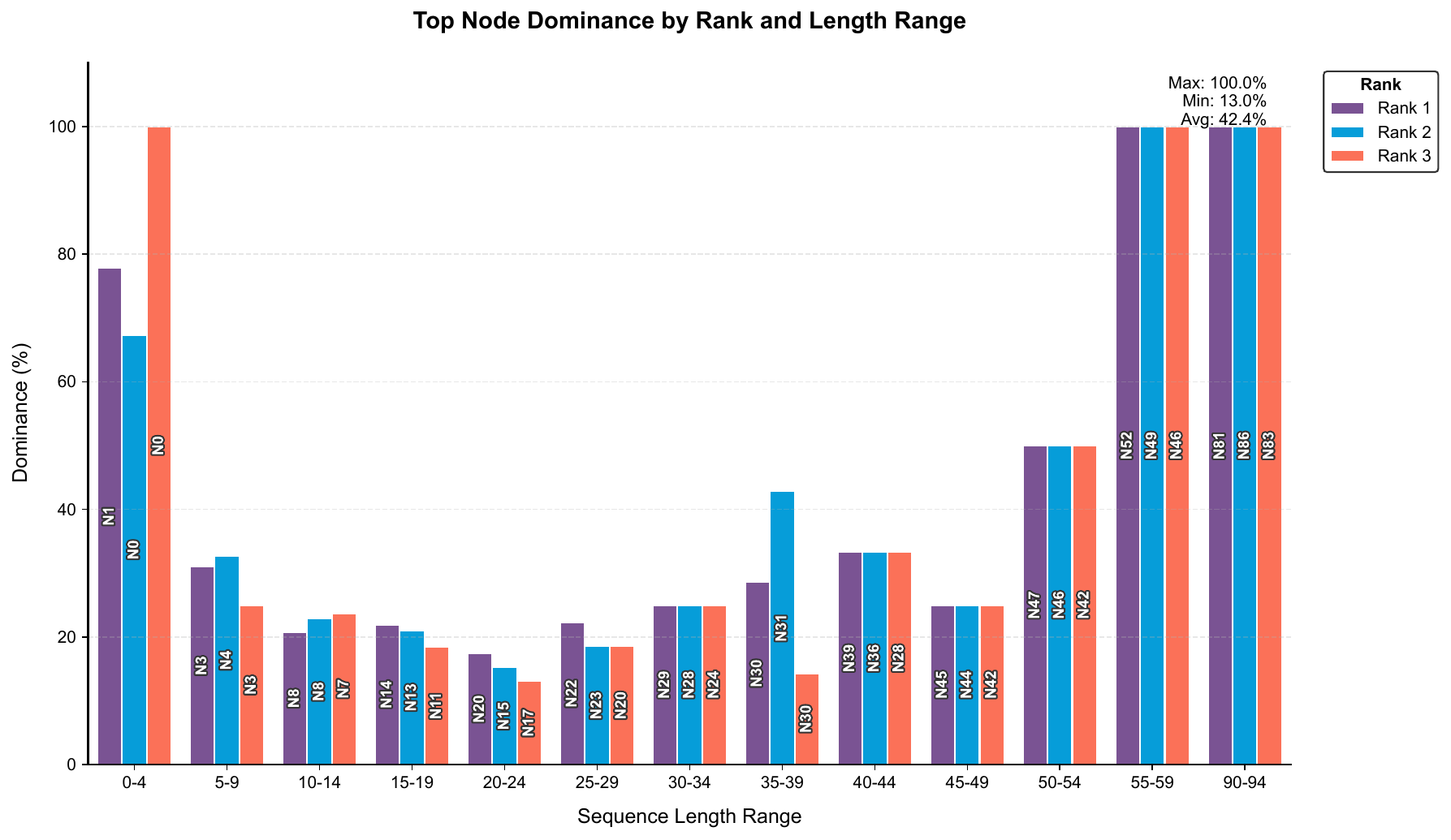}
    \caption{Dataset: BPI13i; Model GAT-TDTE}
    \label{app_subfig:rk13it}
\end{subfigure}

\vspace{0.5em}
\begin{subfigure}{0.48\textwidth}
    \includegraphics[width=\textwidth, trim={0 0 0 1cm}, clip]{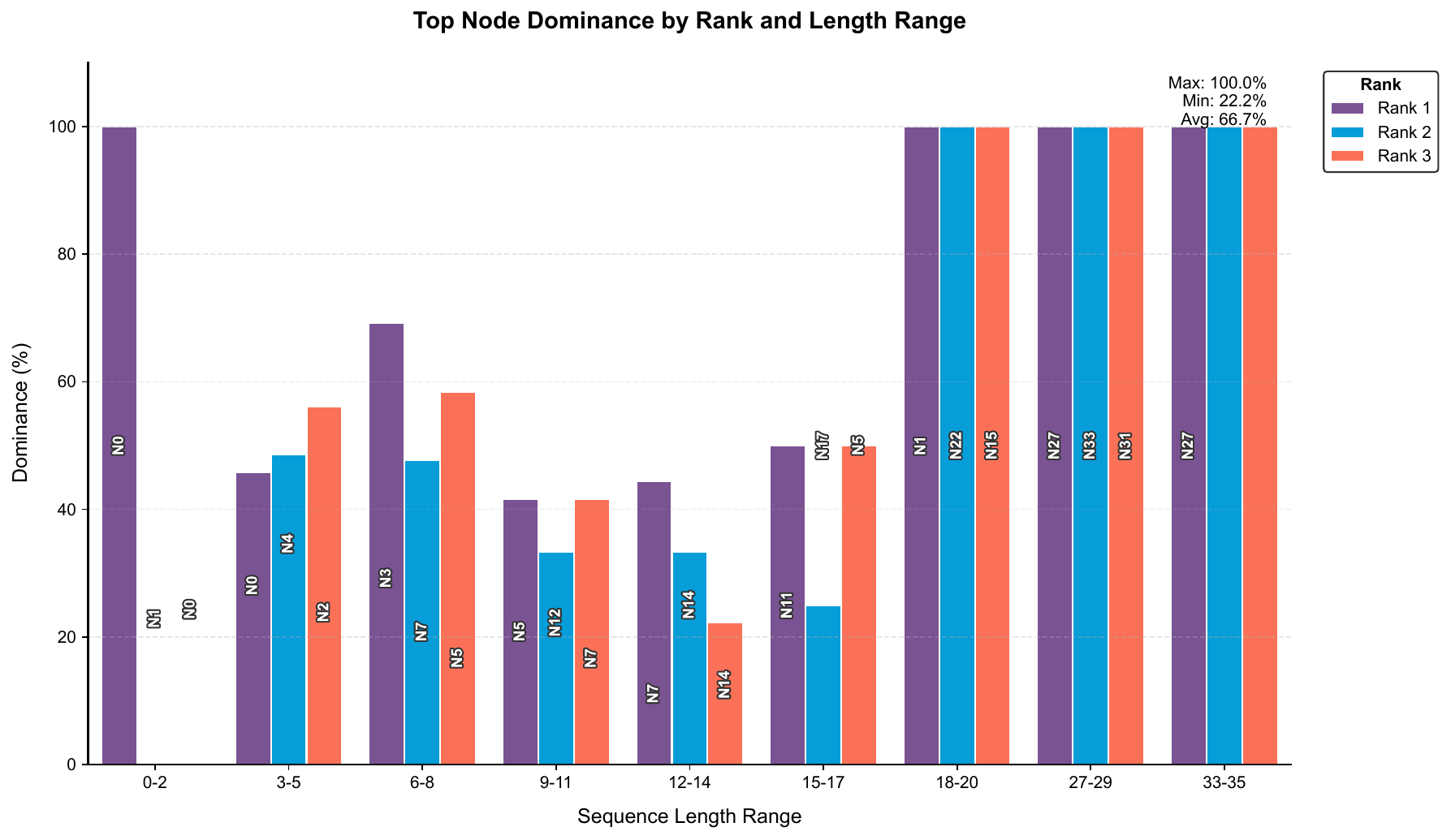}
    \caption{Dataset: BPI13c; Model: GAT-TD}
    \label{app_subfig:rk13c}
\end{subfigure}
\hfill
\begin{subfigure}{0.48\textwidth}
    \includegraphics[width=\textwidth, trim={0 0 0 1cm}, clip]{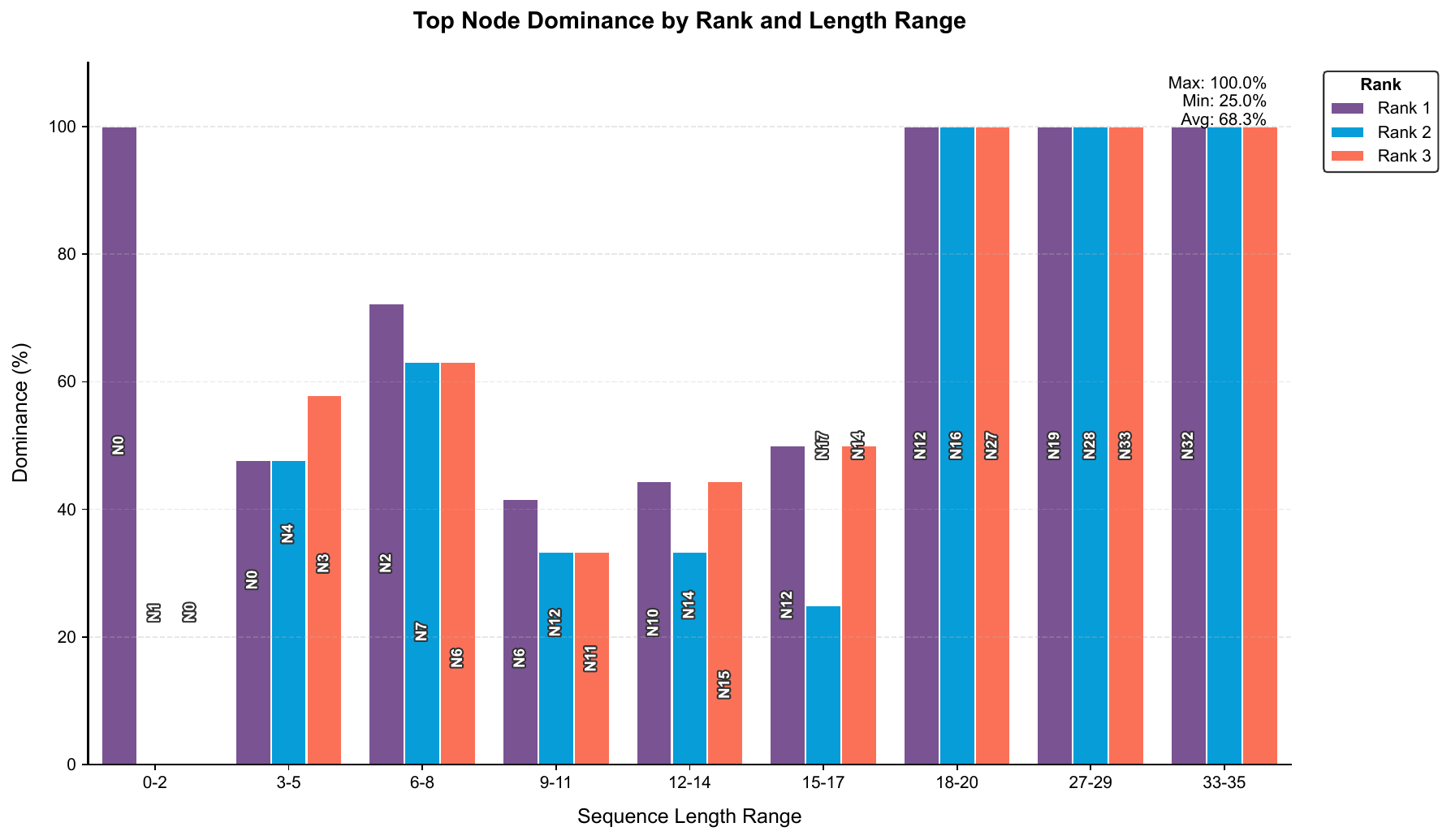}
    \caption{Dataset: BPI13c; Model GAT-TDTE}
    \label{app_subfig:rk13ct}
\end{subfigure}

\caption{Top Node Dominance by Rank and length Range (part 1)}
\label{fig:rk5x2grid_p1}
\end{figure}

\begin{figure}[htbp!]
\centering
\begin{subfigure}{0.48\textwidth}
    \includegraphics[width=\textwidth, trim={0 0 0 1cm}, clip]{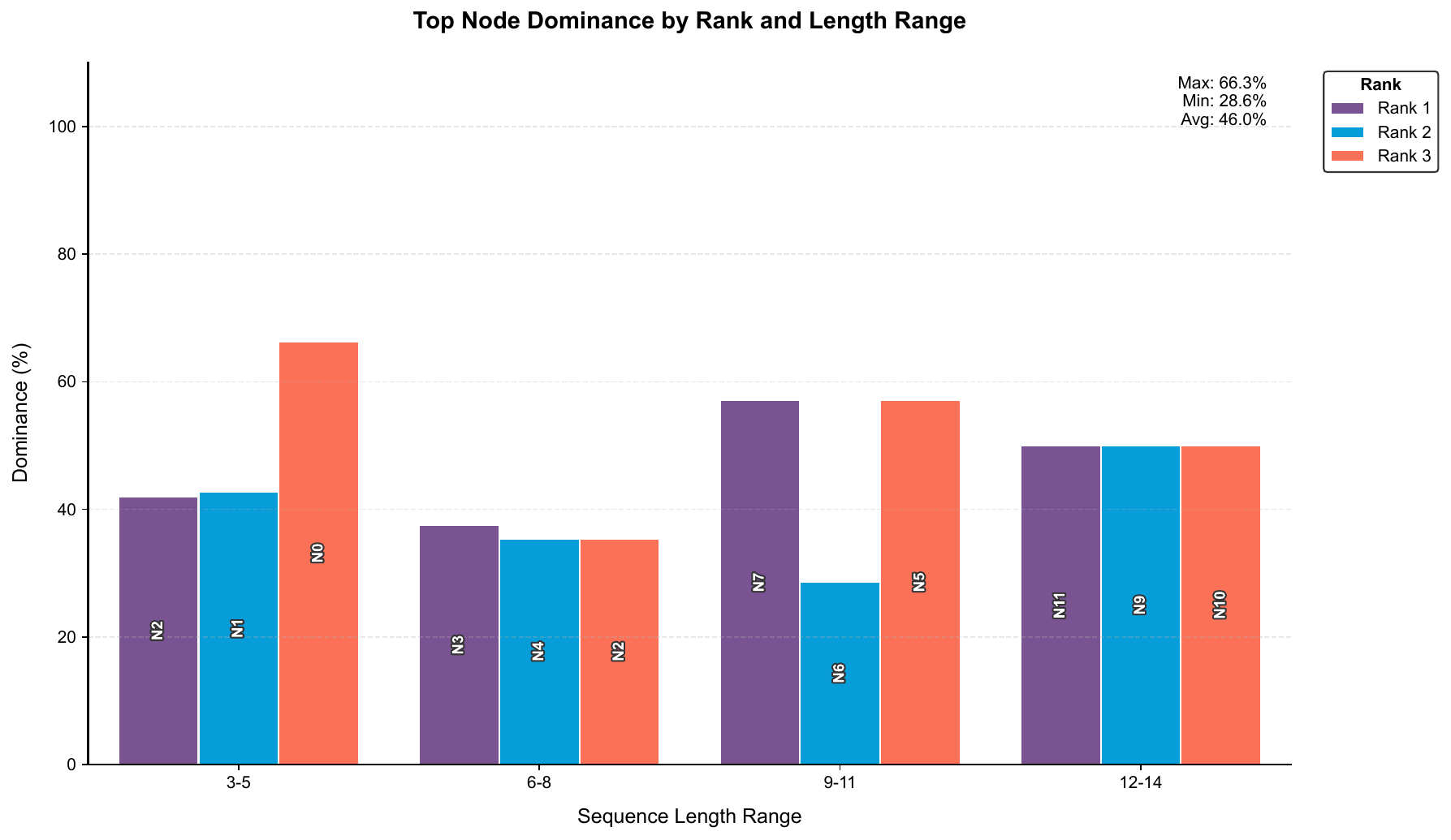}
    \caption{Dataset: Helpdeask; Model: GAT-TD}
    \label{app_subfig:rk0}
\end{subfigure}
\hfill
\begin{subfigure}{0.48\textwidth}
    \includegraphics[width=\textwidth, trim={0 0 0 1cm}, clip]{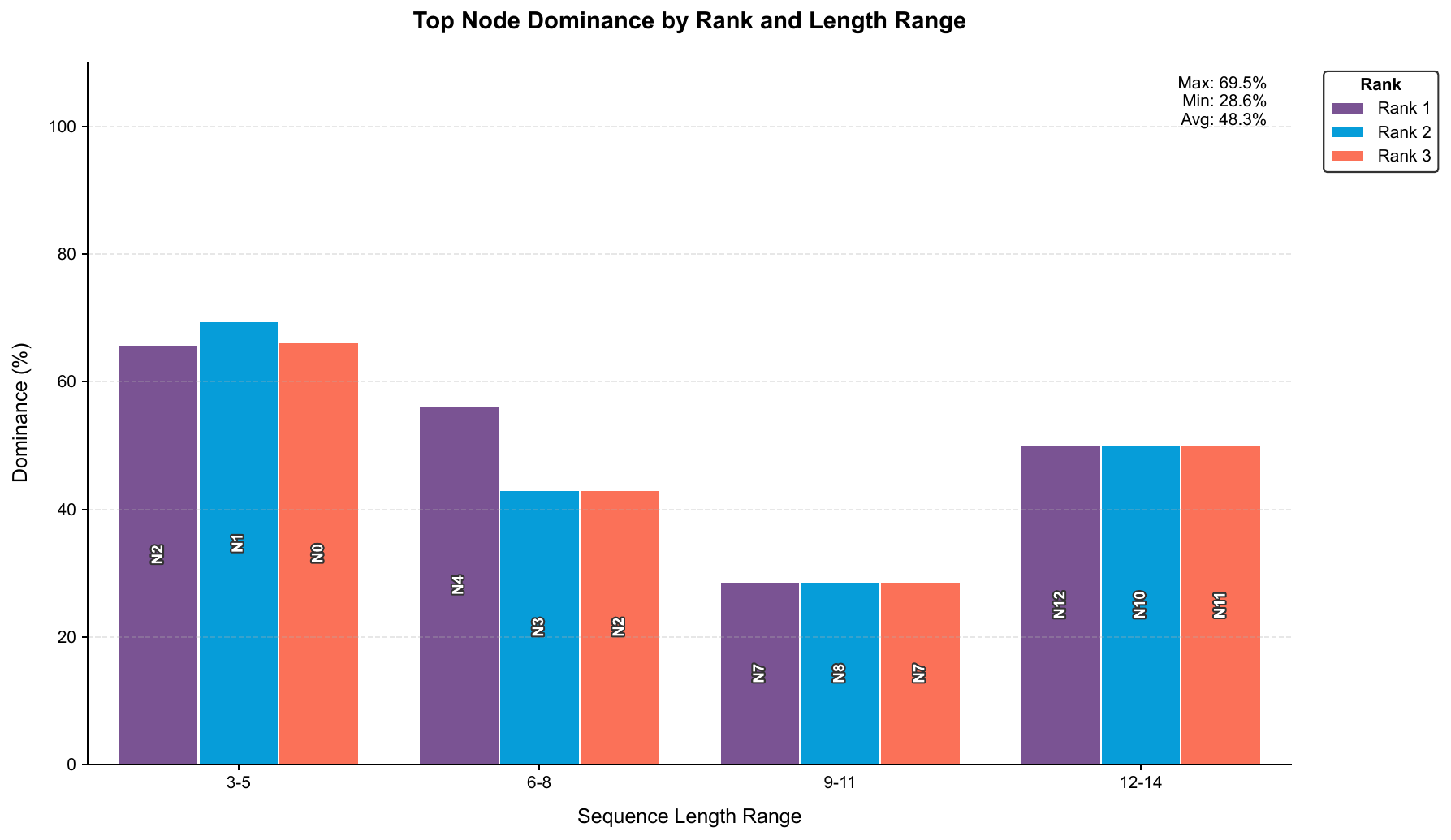}
    \caption{Dataset: Helpdesk; Model GAT-TDTE}
    \label{app_subfig:rk0t}
\end{subfigure}

\vspace{0.5em}
\begin{subfigure}{0.48\textwidth}
    \includegraphics[width=\textwidth, trim={0 0 0 1cm}, clip]{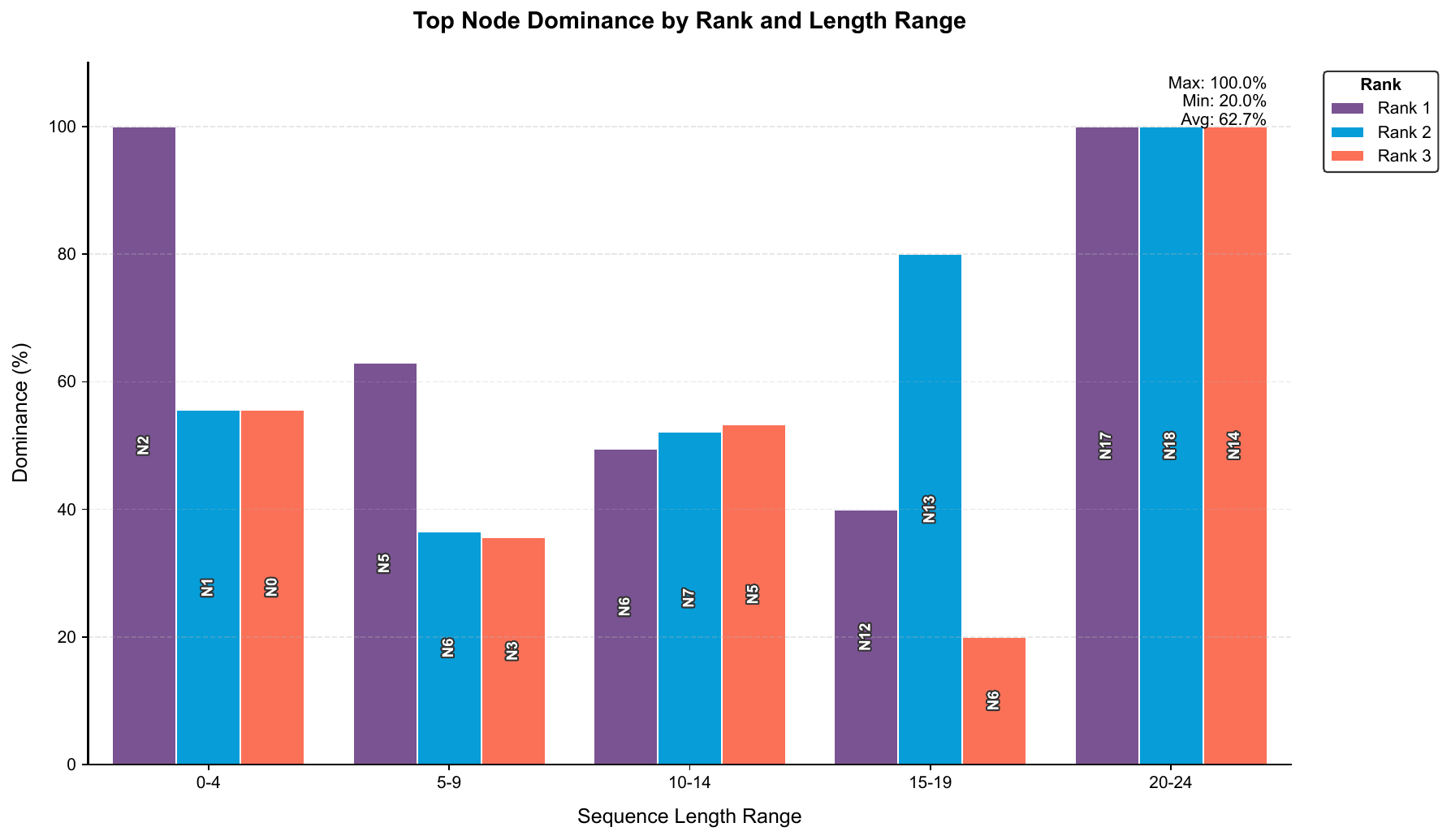}
    \caption{Dataset: BPI20; Model: GAT-TD}
    \label{app_subfig:rk20}
\end{subfigure}
\hfill
\begin{subfigure}{0.48\textwidth}
    \includegraphics[width=\textwidth, trim={0 0 0 1cm}, clip]{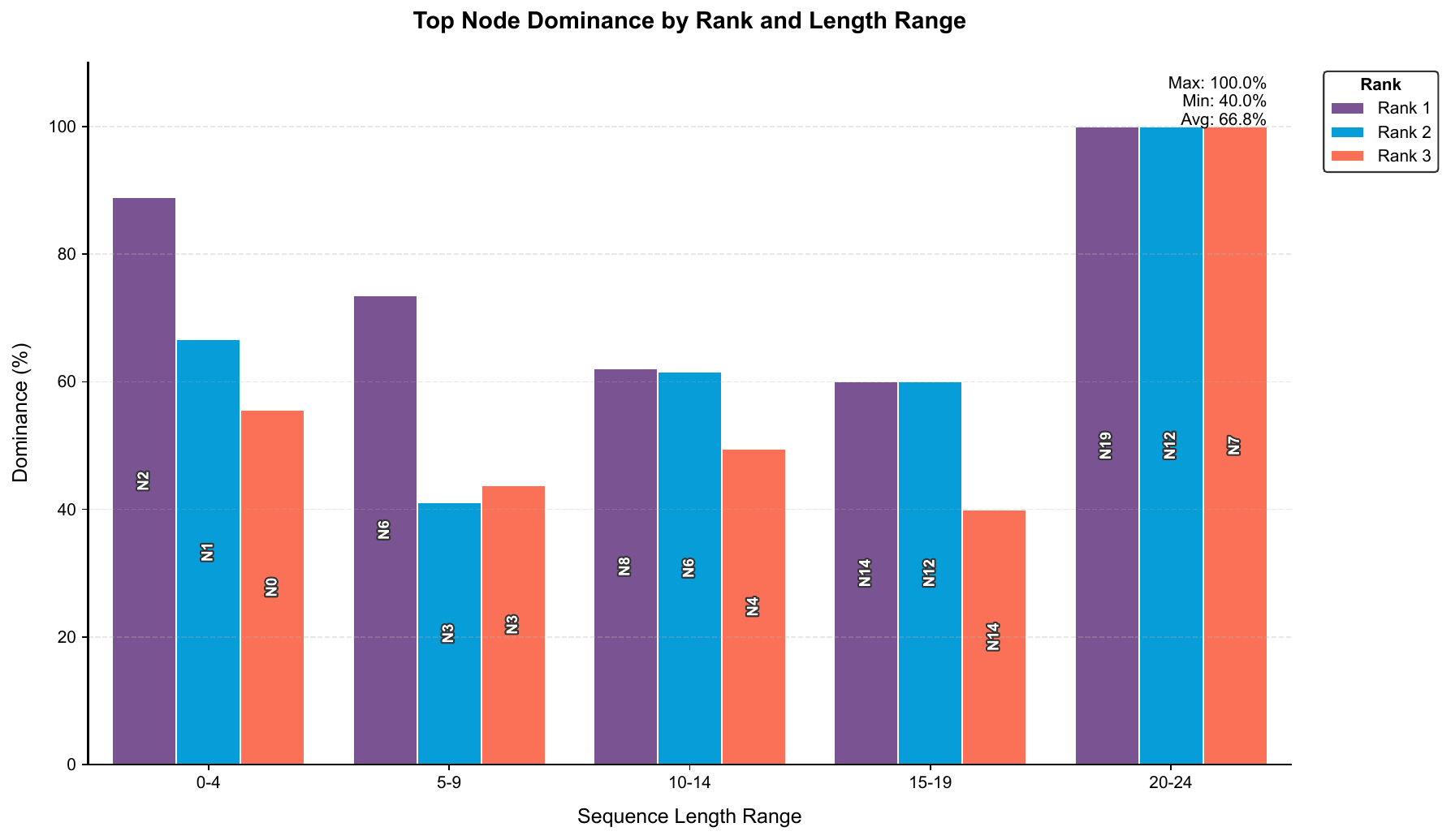}
    \caption{Dataset: BPI20; Model GAT-TDTE}
    \label{app_subfig:rk20t}
\end{subfigure}

\caption{Top Node Dominance by Rank and length Range (part 2)}
\label{fig:rk5x2grid_p2}
\end{figure}

\begin{figure}[htbp!]
\centering

\begin{subfigure}{0.48\textwidth}
    \includegraphics[width=\textwidth, trim={0 0 0 1.25cm}, clip]{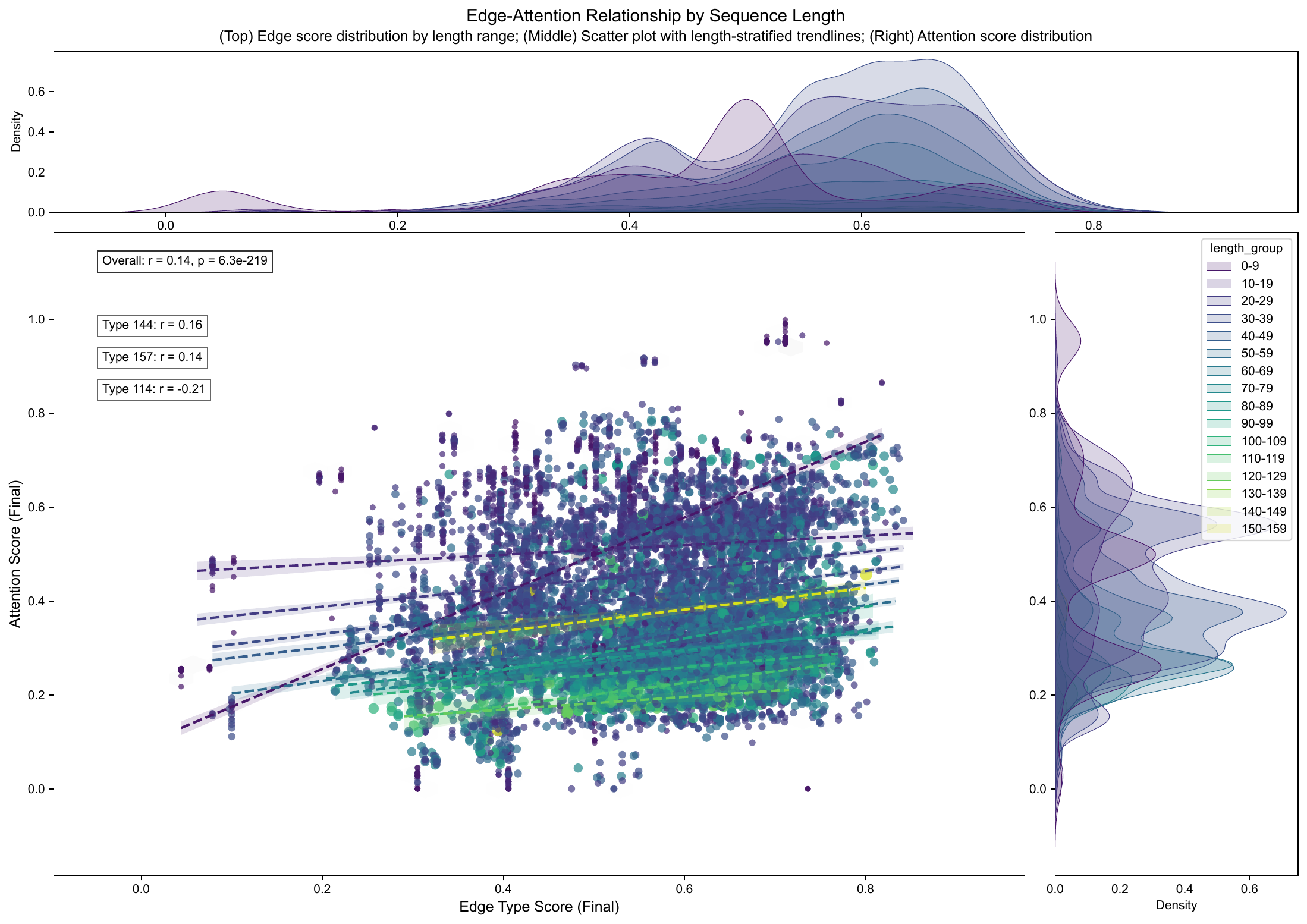}
    \caption{Dataset: BPI12; Model GAT-TDTE;}
    \label{app_subfig:cr12}
\end{subfigure}
\hfill
\begin{subfigure}{0.48\textwidth}
    \includegraphics[width=\textwidth, trim={0 0 0 1.25cm}, clip]{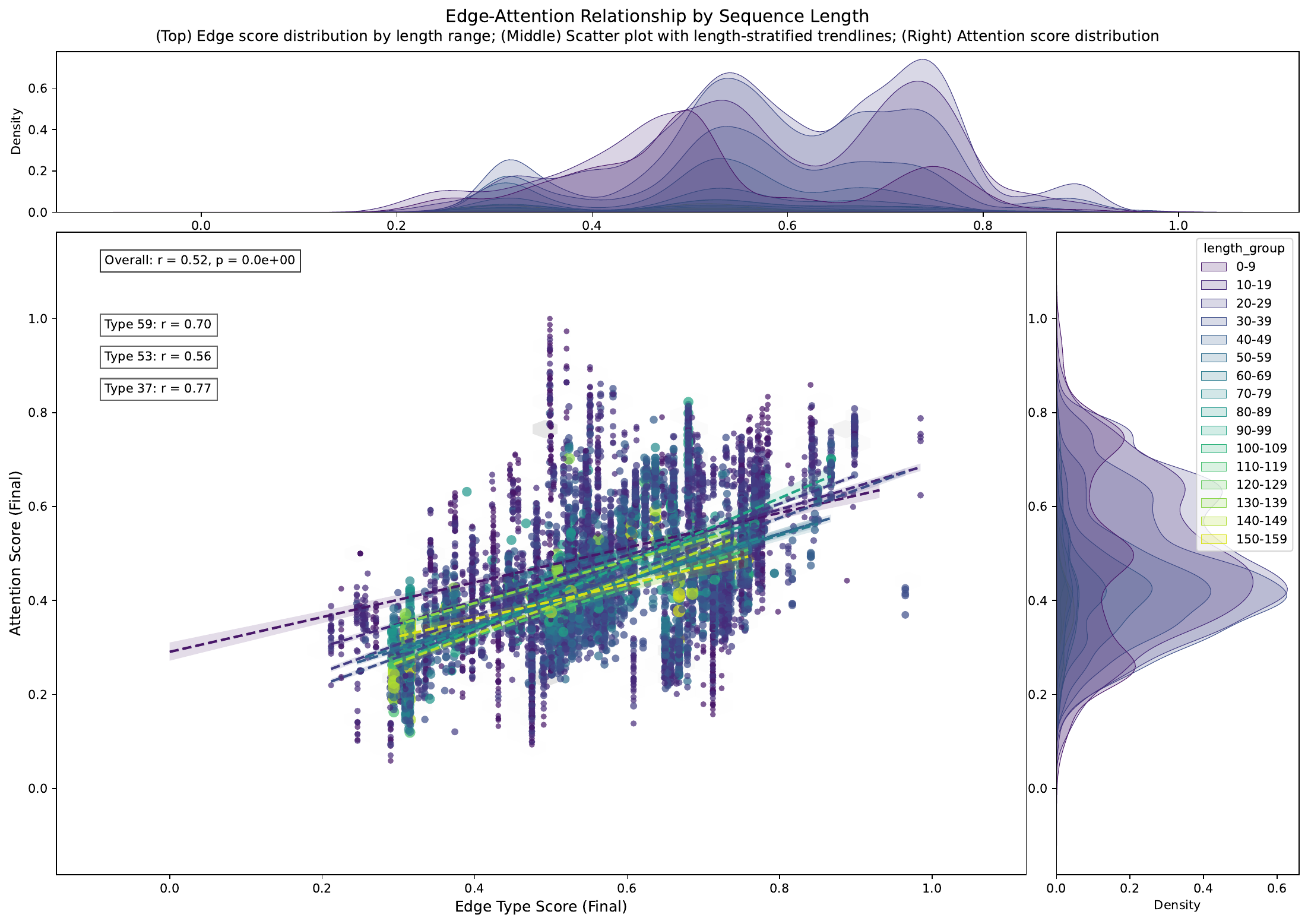}
    \caption{Dataset: BPI12W; Model: GAT-TDTE}
    \label{app_subfig:cr12w}
\end{subfigure}

\vspace{0.5em}
\begin{subfigure}{0.48\textwidth}
    \includegraphics[width=\textwidth, trim={0 0 0 1.25cm}, clip]{BPI13i_transedgedecay_edgeattention.pdf}
    \caption{Dataset: BPI13i; Model GAT-TDTE}
    \label{app_subfig:cr13i}
\end{subfigure}
\hfill
\begin{subfigure}{0.48\textwidth}
    \includegraphics[width=\textwidth, trim={0 0 0 1.25cm}, clip]{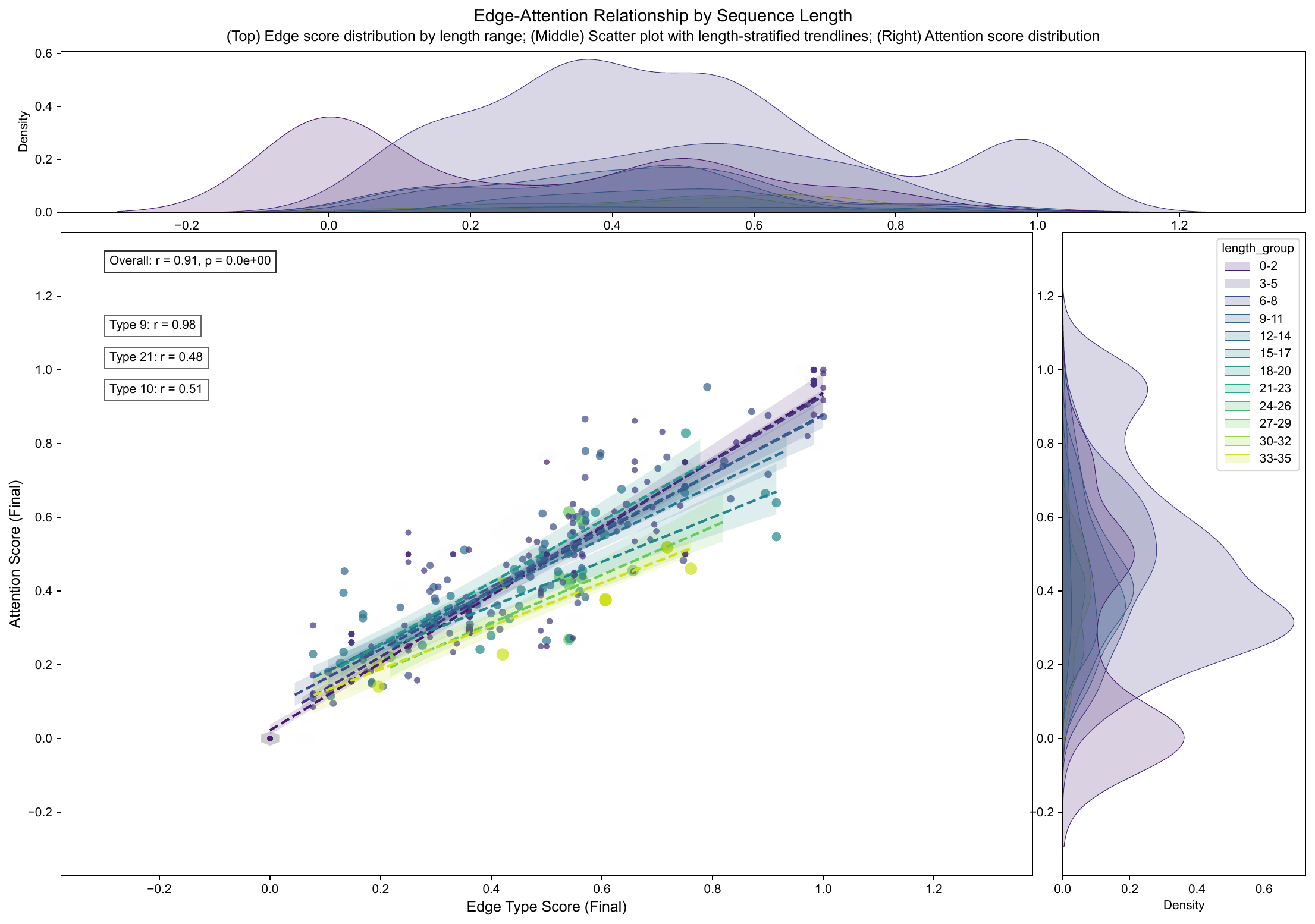}
    \caption{Dataset: BPI13c; Model: GAT-TDTE}
    \label{app_subfig:cr13c}
\end{subfigure}

\vspace{0.5em}
\begin{subfigure}{0.48\textwidth}
    \includegraphics[width=\textwidth, trim={0 0 0 1.25cm}, clip]{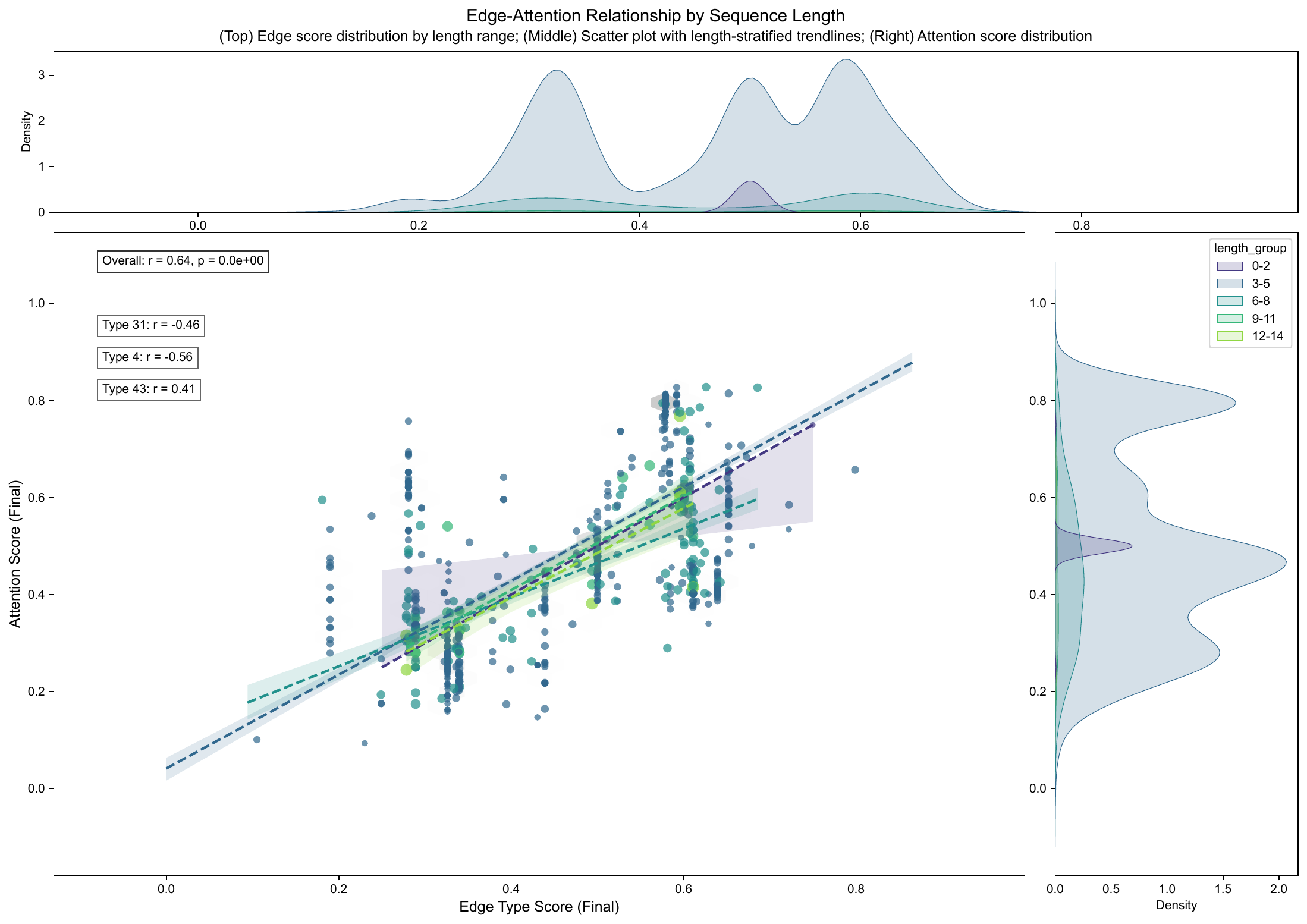}
    \caption{Dataset: Helpdesk; Model GAT-TDTE}
    \label{app_subfig:cr0}
\end{subfigure}
\hfill
\begin{subfigure}{0.48\textwidth}
    \includegraphics[width=\textwidth, trim={0 0 0 1.25cm}, clip]{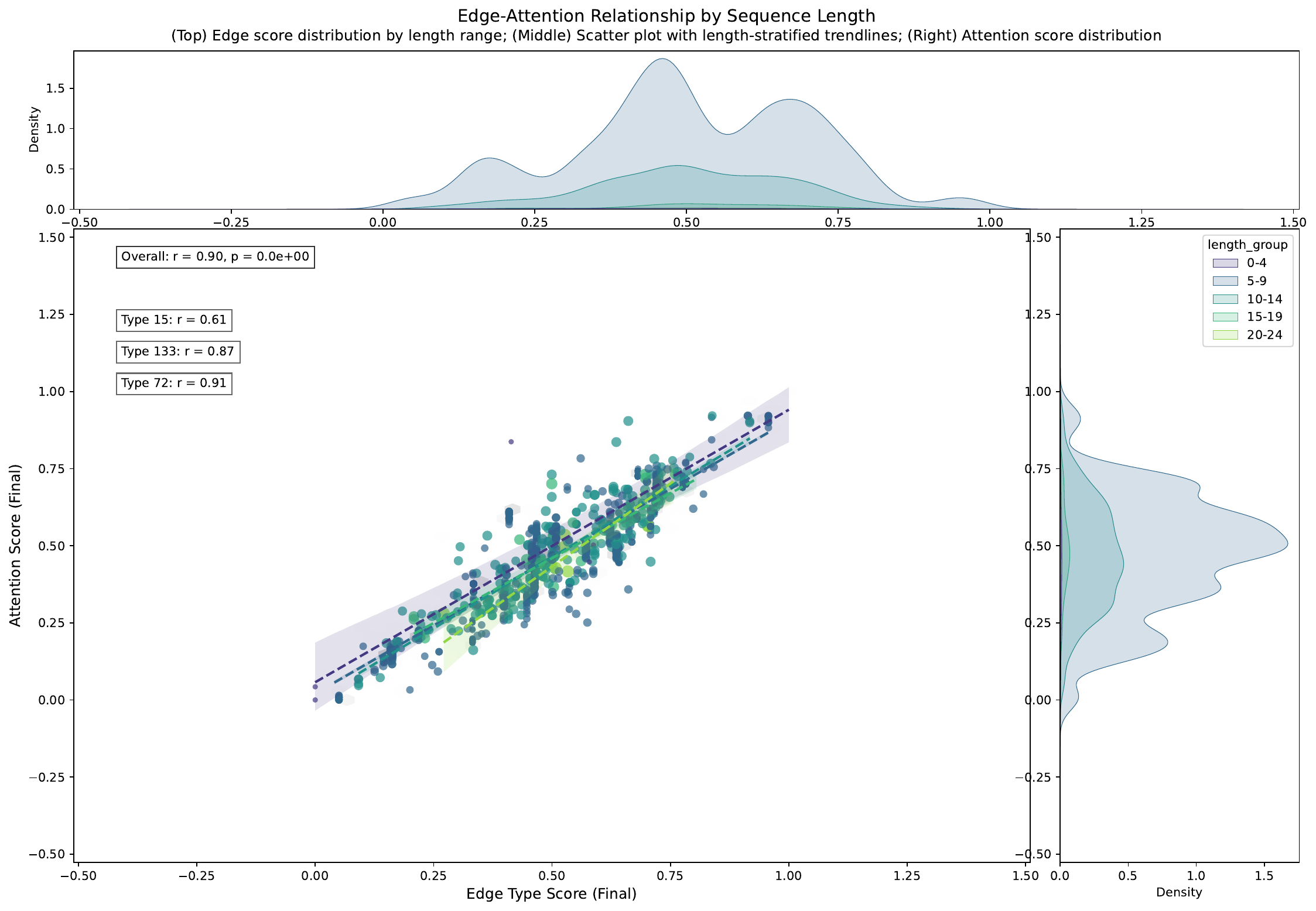}
    \caption{Dataset: BPI20; Model GAT-TDTE}
    \label{app_subfig:cr20}
\end{subfigure}
\caption{Correlation between Edge type Scores and Attention Weights in the GAT-TDTE model}
\label{fig:cor3x2grid}
\end{figure}

\begin{figure}[htbp!]
\centering
\begin{subfigure}{\textwidth}
    \includegraphics[width=\textwidth,  trim={0 0 0 0.75cm}, clip]{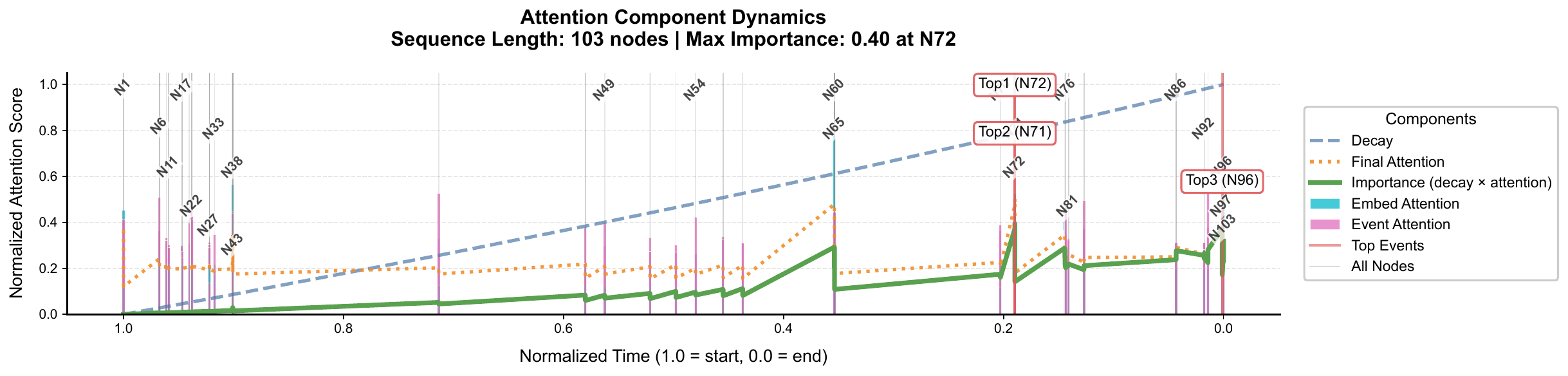}
    \caption{Dynamic Temporal Windows with Attention Decomposition}
     \label{subfig:sample12}
\end{subfigure}

\begin{subfigure}{\textwidth}
    \includegraphics[width=\textwidth,  trim={0 0 0 0.75cm}, clip]{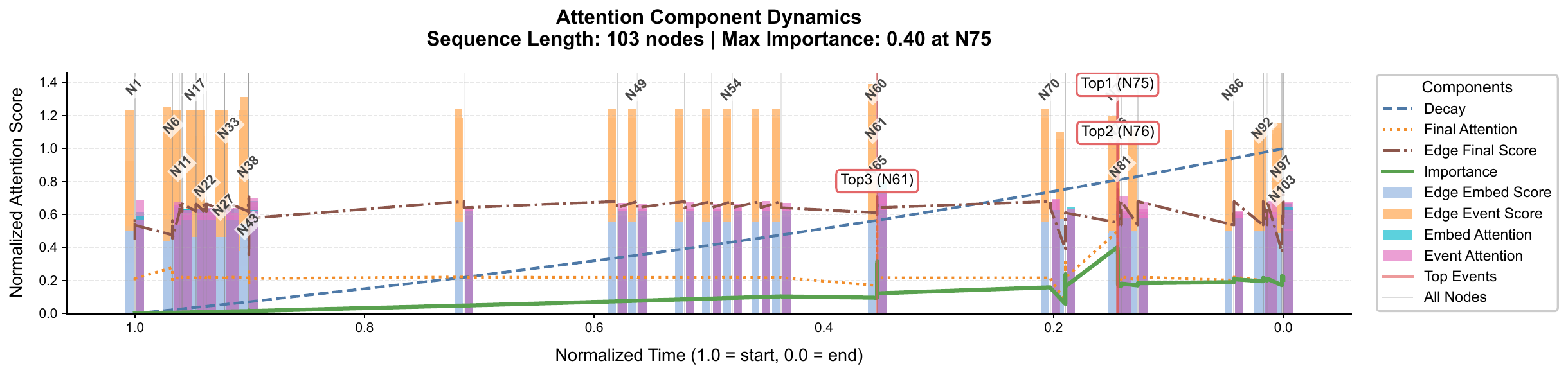}
    \caption{Dynamic Temporal Windows with Attention Decomposition Interacted with Edge Type Embedding}
    \label{subfig:transample12}
\end{subfigure}

\caption{Temporal Attention Decomposition for a Representative Sequence (103 events) from BPI12 Using GAT-TD and GAT-TDTE Model}
\label{fig:edgecasebpi12}
\end{figure}

\begin{figure}[htbp!]
\centering
\begin{subfigure}{\textwidth}
    \includegraphics[width=\textwidth,  trim={0 0 0 0.75cm}, clip]{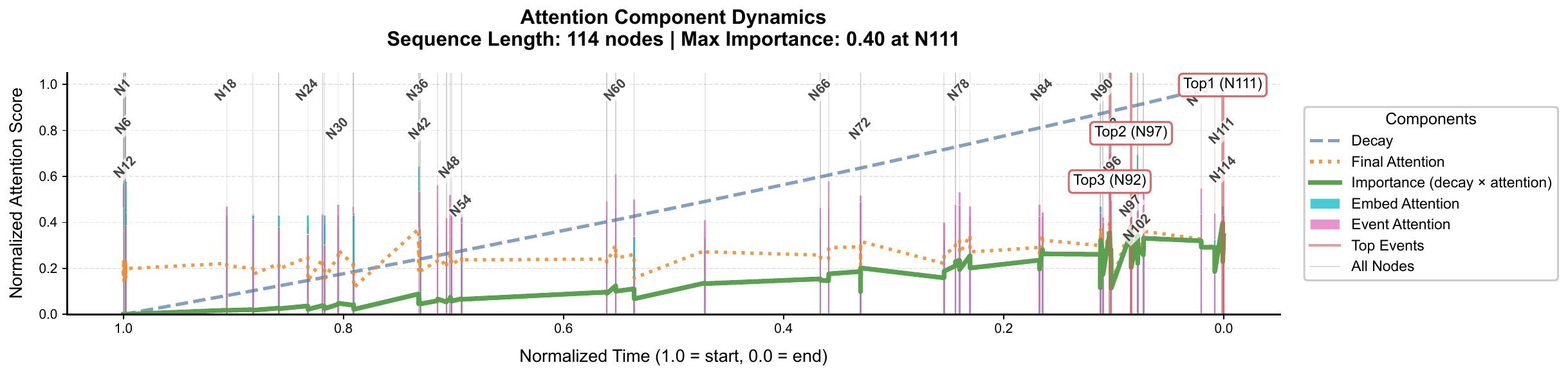}
    \caption{Dynamic Temporal Windows with Attention Decomposition}
     \label{subfig:sample12w}
\end{subfigure}

\begin{subfigure}{\textwidth}
    \includegraphics[width=\textwidth,  trim={0 0 0 0.75cm}, clip]{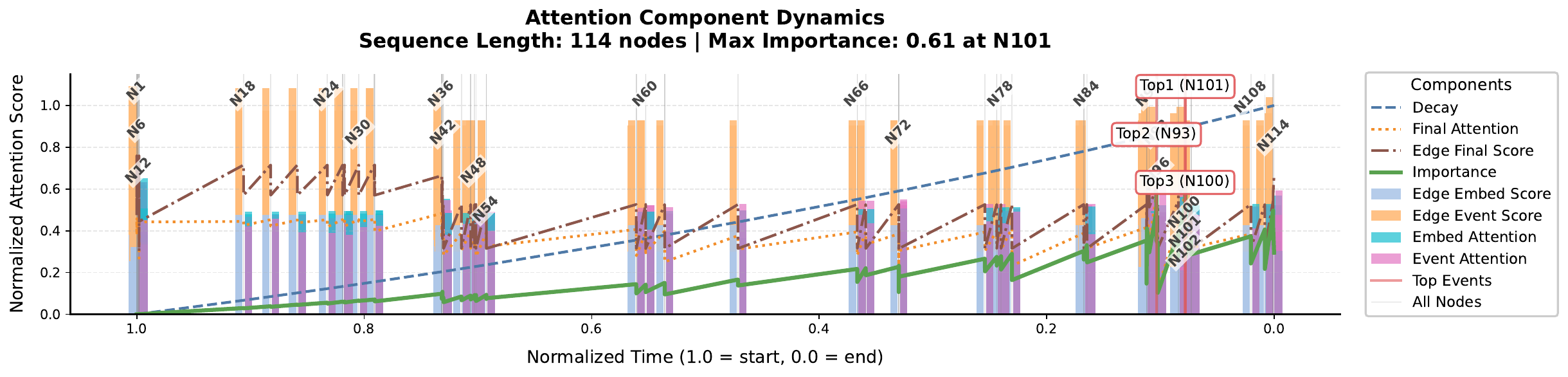}
    \caption{Dynamic Temporal Windows with Attention Decomposition Interacted with Edge Type Embedding}
    \label{subfig:transample12w}
\end{subfigure}

\caption{Temporal Attention Decomposition for a Representative Sequence (114 events) from BPI12w Using GAT-TD and GAT-TDTE Model}
\label{fig:edgecasebpi12w}
\end{figure}

\begin{figure}[htbp!]
\centering
\begin{subfigure}{\textwidth}
    \includegraphics[width=\textwidth,  trim={0 0 0 0.75cm}, clip]{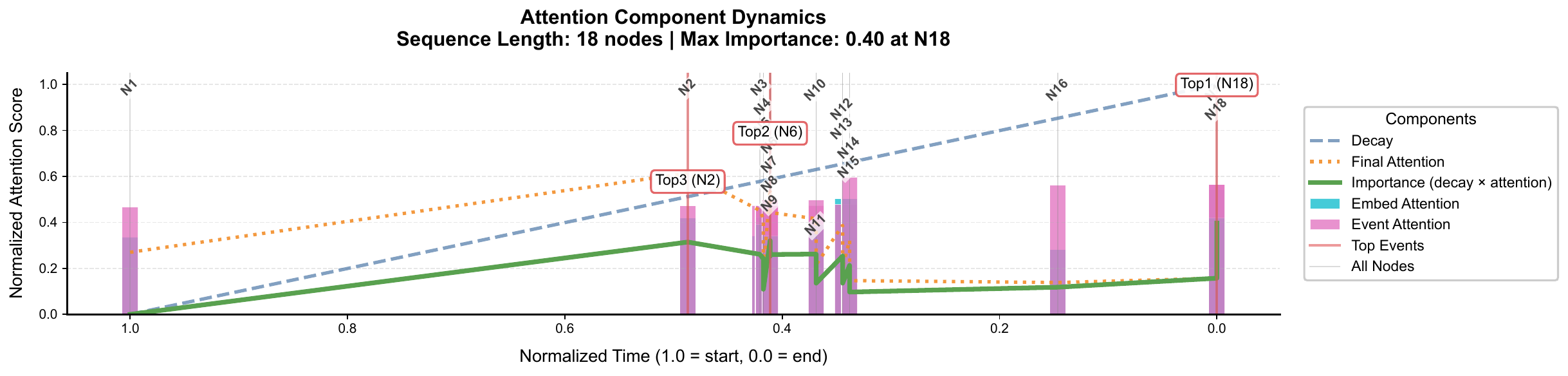}
    \caption{Dynamic Temporal Windows with Attention Decomposition}
     \label{subfig:sample13c}
\end{subfigure}

\begin{subfigure}{\textwidth}
    \includegraphics[width=\textwidth,  trim={0 0 0 0.75cm}, clip]{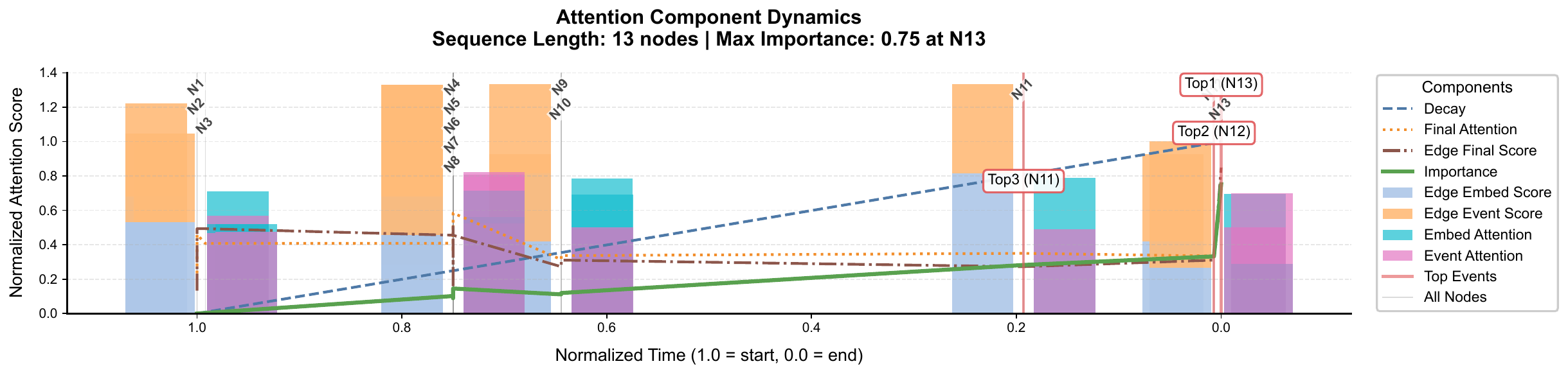}
    \caption{Dynamic Temporal Windows with Attention Decomposition Interacted with Edge Type Embedding}
    \label{subfig:transample13c}
\end{subfigure}

\caption{Temporal Attention Decomposition for a Representative Sequence (18 events) from BPI13c Using GAT-TD and GAT-TDTE Model}
\label{fig:edgecasebpi13c}
\end{figure}

\begin{figure}[htbp!]
\centering
\begin{subfigure}{\textwidth}
    \includegraphics[width=\textwidth,  trim={0 0 0 0.75cm}, clip]{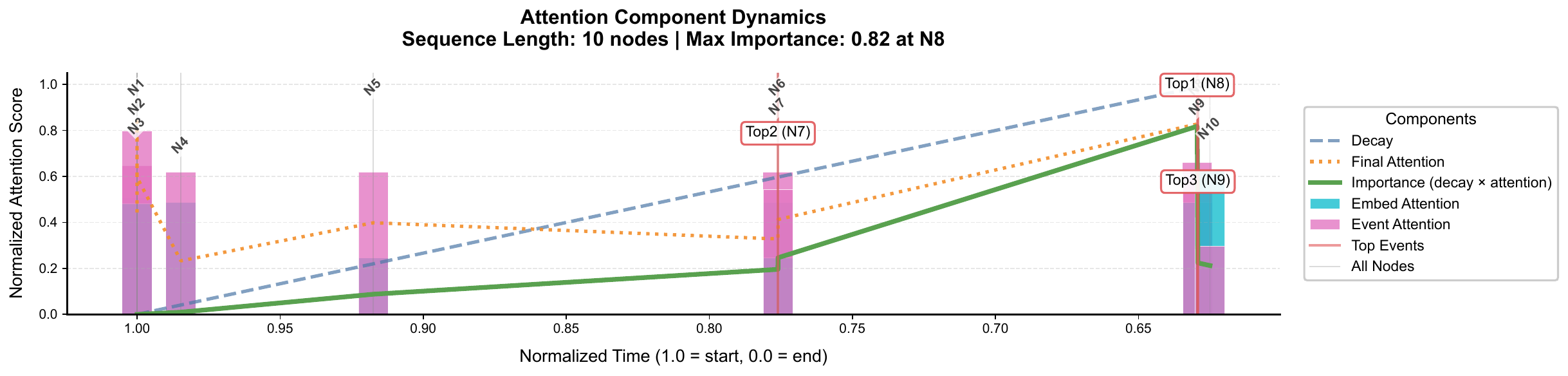}
    \caption{Dynamic Temporal Windows with Attention Decomposition}
     \label{subfig:sample0}
\end{subfigure}

\begin{subfigure}{\textwidth}
    \includegraphics[width=\textwidth,  trim={0 0 0 0.75cm}, clip]{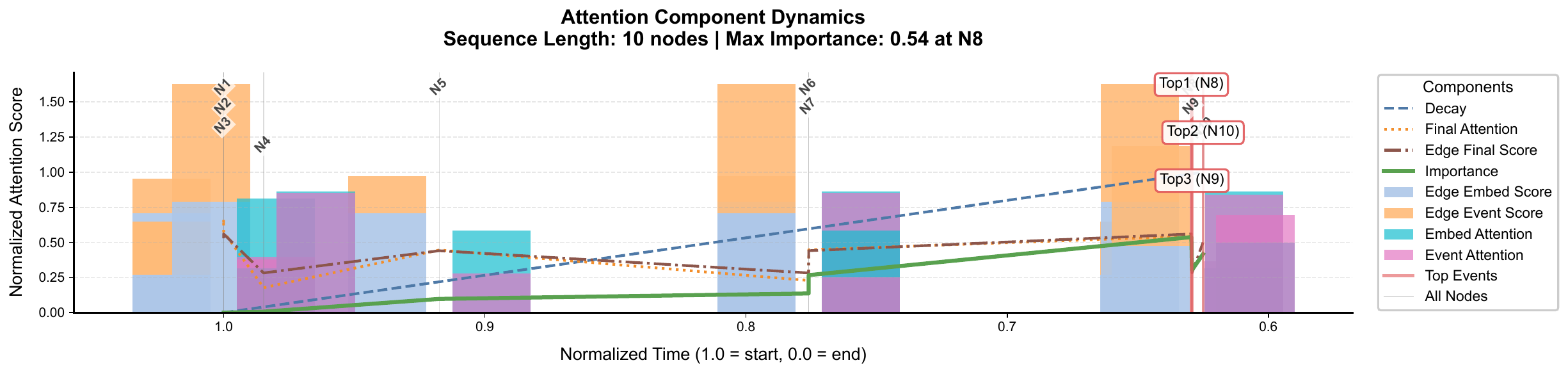}
    \caption{Dynamic Temporal Windows with Attention Decomposition Interacted with Edge Type Embedding}
    \label{subfig:transample0}
\end{subfigure}

\caption{Temporal Attention Decomposition for a Representative Sequence (10 events) from Helpdesk Using GAT-TD and GAT-TDTE Model}
\label{fig:edgecasebpi0}
\end{figure}

\begin{figure}[htbp!]
\centering
\begin{subfigure}{\textwidth}
    \includegraphics[width=\textwidth,  trim={0 0 0 0.75cm}, clip]{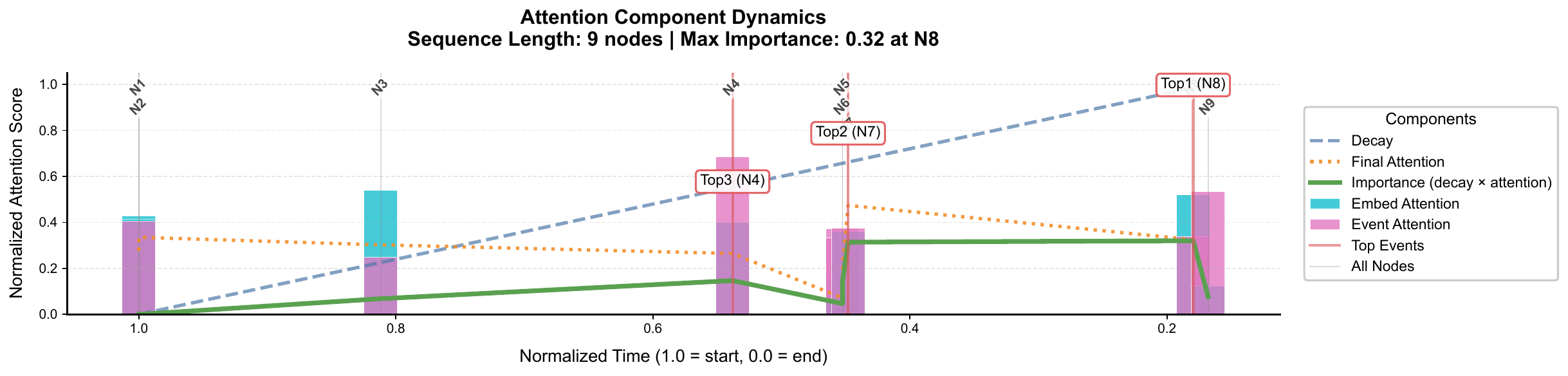}
    \caption{Dynamic Temporal Windows with Attention Decomposition}
     \label{subfig:sample20}
\end{subfigure}

\begin{subfigure}{\textwidth}
    \includegraphics[width=\textwidth,  trim={0 0 0 0.75cm}, clip]{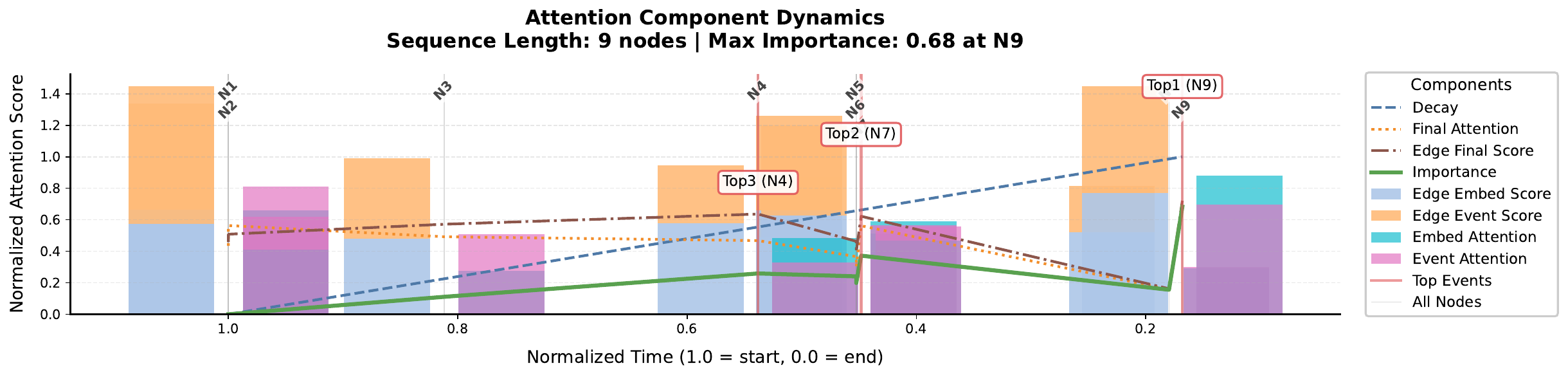}
    \caption{Dynamic Temporal Windows with Attention Decomposition Interacted with Edge Type Embedding}
    \label{subfig:transample20}
\end{subfigure}

\caption{Temporal Attention Decomposition for a Representative Sequence (9 events) from BPI20 Using GAT-TD and GAT-TDTE Model}
\label{fig:edgecasebpi20}
\end{figure}

\end{document}